%% file: paper.tex
\documentclass[]{bytedance_seed}

\usepackage[toc,page,header]{appendix}

\usepackage{minitoc}
\usepackage{makecell}
\usepackage{diagbox}
\usepackage{xcolor}
\usepackage{amsfonts}
\usepackage{amsmath}
\usepackage{capt-of}
\usepackage{pifont}
\usepackage{fontawesome5}

\usepackage{fancyvrb}

\usepackage[most]{tcolorbox}
\newtcolorbox[list inside=prompt,auto counter,number within=section]{prompt}[1][]{
    colbacktitle=black!60,
    coltitle=white,
    fontupper=\footnotesize,
    boxsep=5pt,
    left=0pt,
    right=0pt,
    top=0pt,
    bottom=0pt,
    boxrule=1pt,
    #1,
}

\usepackage{CJKutf8}
\usepackage{newunicodechar}
\usepackage[export]{adjustbox}
\usepackage{booktabs}
\definecolor{mygreen}{HTML}{76B900}

\newcommand{\cmark}{\ding{51}}  % ✓
\newcommand{\xmark}{\ding{55}}  % ✗
\newcommand{\OnDemand}{\textcolor{green!55!black}{\faSmile[regular]}}
\newcommand{\Always}{\textcolor{gray!65!black}{\faTired[regular]}} % fallback: \faMeh[regular]
\title{Lumine: An Open Recipe for Building Generalist Agents in 3D Open Worlds}

\author{ByteDance Seed}
\affiliation{See \hyperref[sec:contribution]{Contributions} section for a full author list.}

\abstract{
We introduce Lumine, the first open recipe for developing generalist agents capable of completing hours-long complex missions in real time within challenging 3D open-world environments. Lumine adopts a human-like interaction paradigm that unifies perception, reasoning, and action in an end-to-end manner, powered by a vision-language model. It processes raw pixels at 5 Hz to produce precise 30 Hz keyboard–mouse actions and adaptively invokes reasoning only when necessary. Trained in Genshin Impact, Lumine successfully completes the entire five-hour Mondstadt main storyline on par with human-level efficiency and follows natural language instructions to perform a broad spectrum of tasks in both 3D open-world exploration and 2D GUI manipulation across collection, combat, puzzle-solving, and NPC interaction. In addition to its in-domain performance, Lumine demonstrates strong zero-shot cross-game generalization. Without any  fine-tuning, it accomplishes 100-minute missions in Wuthering Waves and the full five-hour first chapter of Honkai: Star Rail. These promising results highlight Lumine’s effectiveness across distinct worlds and interaction dynamics, marking a concrete step toward generalist agents in open-ended environments. 
}

\date{\today}
\correspondence{\email{weihao001@ntu.edu.sg}, \email{shiguang.sg@bytedance.com}}

\checkdata[Project Page]{\url{https://www.lumine-ai.org/}}

\begin{document}
\maketitle
\begin{figure}[h]
    \centering
    \vspace{-20pt}
    \includegraphics[width=\linewidth]{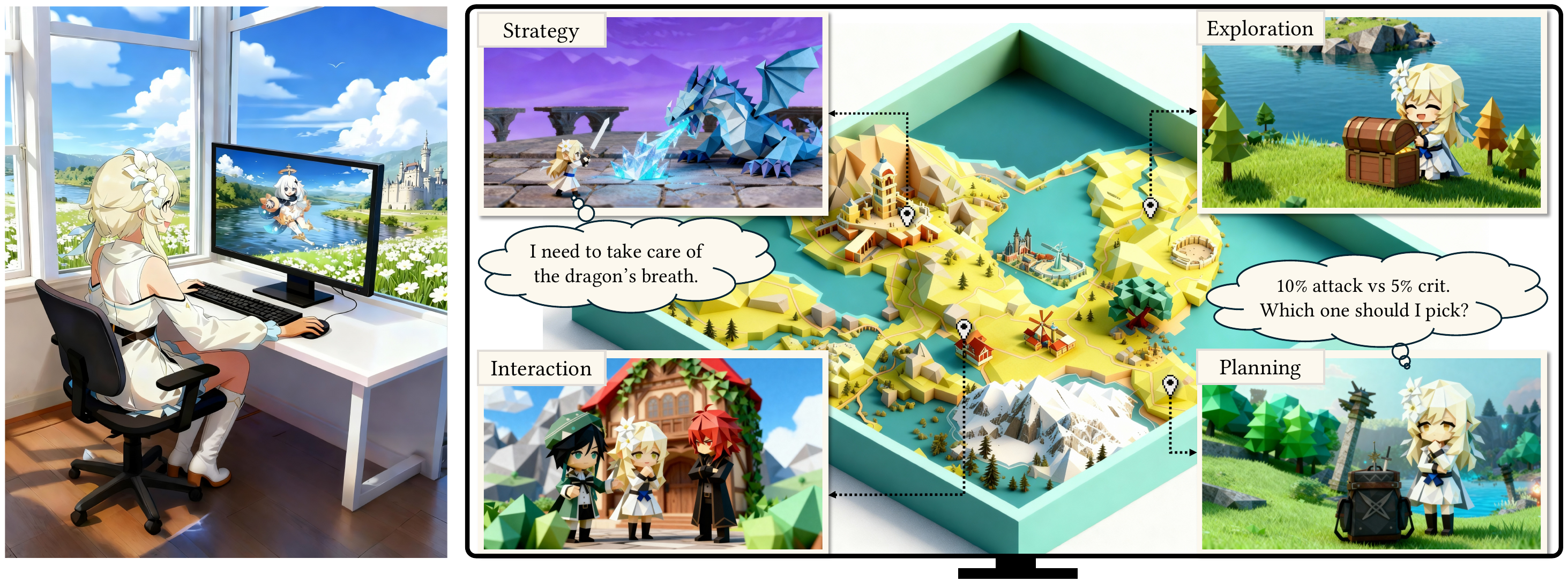}
    \caption{Lumine, the first AI agent to complete hours-long missions in real time within expansive 3D open worlds. }
\end{figure}

\newpage
\tableofcontents
\newpage

\input{sections/1_introduction}
\input{sections/2_relatedwork}

\input{sections/3_preliminary}
\input{sections/4_methods}
\input{sections/5_results}

\input{sections/6_conclusion}

\bibliographystyle{plainnat}
\bibliography{main}

\clearpage

\beginappendix

\input{sections/appendix}

\end{document}

%% file: sections/1_introduction.tex
% \textsc{Lumine}
\section{Introduction}
Building generalist autonomous agents that can \textbf{perceive}, \textbf{reason}, and \textbf{act} at human-level within \textbf{open worlds} has long been the north star for artificial general intelligence research~\citep{laird1987soar,mountcastle1979organizing,AgreChapman1989,russell1995modern,Fuster2004,schmidhuber2018one,fung2025embodied}. Over the past decades, remarkable progress has been achieved in constrained domains: agents can match or surpass humans in games such as Atari~\citep{mnih2015human}, Go~\citep{silver2017mastering}, Dota II~\citep{berner2019dota},  StarCraft II~\citep{vinyals2019grandmaster} and Gran Turismo~\citep{wurman2022outracing}. Despite these achievements, such agents remain confined to closed environments, optimizing a single, well-defined objective with explicitly shaped rewards via large-scale reinforcement learning in a perceive-then-act paradigm. This specialization yields mastery but brittle intelligence with limited abstraction, weak transfer, and poor adaptability to the ambiguity and diversity inherent in open-world scenarios.

A promising path to broader generalization is to ground agents in natural language, leveraging it as a universal medium for knowledge and capabilities across modalities and domains. The recent success of large language foundation models~\cite{achiam2023gpt,team2023gemini,hurst2024gpt,guo2025deepseek} reinforces this view through their superior common-sense understanding and reasoning. These models have begun to extend beyond static text processing into interactive environments such as robotics~\citep{ahn2022can,driess2023palm,zitkovich2023rt,intelligence2025pi} and video games~\citep{wang2023describe,wang2023voyager,tan2024cradle,hershey2025claude}. While these agents have yet to achieve human-level performance in accuracy and efficiency, these developments signal a promising direction for building versatile, language-grounded agents capable of perception, reasoning and acting in complex worlds.

In this work, we take a step toward this goal by exploring the recipe for building generalist agents in 3D open-world environments. We systematically summarize the six core challenges in achieving this goal:

\begin{itemize}
    \item \textbf{Scalable Environments.} The selection and design of environments that are both rich and diverse, providing compositional dynamics that challenge agents to interact, learn, and generalize, while remaining standardized, scalable, and reproducible~\citep{bellemare2013arcade,brockman2016openai,beattie2016deepmind,tan2024cradle, raad2024scaling}.
    
    \item \textbf{Multimodal Perception.} The ability to fuse and interpret heterogeneous sensory streams, including embodied 3D vision, 2D graphical user interfaces (GUIs), textual information and other modalities, to construct an actionable understanding of both the external world and the agent’s state~\citep{alayrac2022flamingo,driess2023palm,reed2022generalist}.

    \item \textbf{High-Level Planning.} The ability to generate self-motivated, long-horizon plans that adapt as environmental dynamics evolve, incorporating reflection and self-revision to refine strategies while balancing competing objectives and environmental feedback~\citep{kulkarni2016hierarchical,yao2023react,ahn2022can,wang2023voyager, tan2024cradle}.

    \item \textbf{Low-Level Control.} The skill to ground abstract intentions into precise and executable actions that enable coherent behavior across diverse embodiments~\citep{zitkovich2023rt,black2024pi0}.
    
    \item \textbf{Memory.} The capacity to maintain and leverage experience over various timescales, providing context for consistent decision-making to enable coherent exploration and test-time adaptation~\citep{hochreiter1997long,chen2021decision,ha2018world,wang2023voyager, park2023generative}.

    \item \textbf{Real-Time Inference.} The ability to operate under strict latency constraints, balancing computational deliberation with timely responses and managing asynchronous interactions to avoid missing critical opportunities~\citep{kober2013reinforcement, black2025real}.
\end{itemize}

We introduce Lumine, a comprehensive and scalable recipe together with its prototype model for addressing these challenges as a concrete step toward general-purpose agents. 
We select Genshin Impact, a globally popular 3D open-world game, as our primary testbed. To the best of our knowledge, Lumine is the first agent capable of completing hours-long missions in real time within such an extremely challenging environment. Lumine adopts a human-like interaction paradigm that unifies perception, reasoning and action. Built upon the Qwen2-VL-7B-Base model \citep{wang2024qwen2}, Lumine perceives the game world directly from raw pixels at 5 Hz, and autoregressively generates textual keyboard and mouse actions at 30 Hz using action chunking \citep{zhao2023learning}. The standardized human-like interface and sufficient interaction frequency enable Lumine to operate seamlessly across a wide range of video game environments. Additionally, Lumine adopts a hybrid thinking strategy, allowing it to adaptively enter a thinking mode to produce inner-monologue reasoning before generating executable actions when necessary, thereby avoiding redundant computation and latency without compromising decision quality. 
We further design a three-stage training curriculum to empower the base model with these capabilities:
i) 1731 hours of human gameplay for pre-training to master action primitives;
ii) 200 hours of instruction following data to ground control in language; and
iii) 15 hours of reasoning data to enable hybrid thinking.
To ensure consistency across long horizon, Lumine dynamically maintains up to 20 recent steps in context as short-term memory and preserves reasoning steps as long-term memory. Finally, an end-to-end optimization yields a 25.3× overall latency reduction, enabling real-time inference and smooth task execution.

During pretraining, we observe a distinct progression of emergent capabilities. Lumine first masters object interaction, then develops basic combat and GUI manipulation, and finally acquires an understanding of game-specific mechanisms and navigation skills, all essential for effective exploration in open-world environments. These emergent behaviors indicate that structured visuomotor competence can naturally arise from large-scale imitation of human gameplay, even without explicit supervision. 

Building upon these foundational abilities, instruction-following fine-tuning enables Lumine to demonstrate robust short-horizon control, successfully completing a wide range of tasks lasting from 10 seconds to several minutes, with a success rate exceeding 80\% and generalizing effectively to unseen objectives and scenarios. 

Following the final reasoning fine-tuning stage, Lumine achieves expert human-level efficiency in completing Act I of Mondstadt’s main storyline, a mission lasting about one hour. To evaluate its reasoning generalization, we further assess performance on the remaining main storyline,  Acts II and III, which are excluded from the reasoning dataset but included in the pretraining dataset. Lumine continues to demonstrate comparable performance on these missions, which together typically require about four hours for human players. Beyond the training domain, Lumine exhibits strong zero-shot generalization. It manages to navigate to the new region, Liyue, complete the initial one-hour mission, and reach the Adeptus hidden within the mountains, despite no exposure to such content during training. Lumine’s capabilities further extend across entirely different games, demonstrating cross-game generalization without any additional fine-tuning. It successfully completes a 100-minute mission in Wuthering Waves and the five-hour first chapter of Honkai: Star Rail, showcasing its ability to transfer visuomotor and reasoning competence to unseen environments.

These results establish Lumine as the first agent capable of real-time long-horizon task completion and cross-environment transfer in open-world settings, demonstrating the effectiveness of the Lumine recipe in developing generalist agents for complex 3D worlds.

%% file: sections/2_relatedwork.tex
\section{Related Work}
Lumine aims to advance the development of general-purpose agents capable of solving diverse, long-horizon tasks. Although we primarily validate this concept within video game environments, the underlying design naturally shares common principles with GUI agents and robotic vision-language-action (VLA) models. We discuss these connections across the following dimensions and compare Lumine with other representative game agents in Table~\ref{tab:related_work}.

\input{tables/methods_comparison}

\textbf{Agent}. Traditional agents~\citep{mnih2015human, pathak2017curiosity, jaderberg2019human, berner2019dota, vinyals2019grandmaster, wurman2022outracing, baker2022videopretrainingvptlearning} optimize policy networks from scratch via supervised learning or reinforcement learning in a system 1 style~\citep{kahneman2011thinking}. Such agents often generalize poorly to unseen scenarios and struggle to incorporate prior knowledge for temporal adaptation. Moreover, their limited language grounding fundamentally constrains their application in modern interactive environments saturated with textual and symbolic information. 
The recent rise of large language models (LLMs) and vision–language models (VLMs)~\citep{achiam2023gpt, anthropic2025claude3.7, google2025gemini2.5} has demonstrated strong capabilities in language understanding and commonsense reasoning. Even without additional training, prompt-based agents built upon LLMs and VLMs have shown that such models can serve as powerful foundation models for general-purpose agents, achieving impressive results in complex, long-horizon tasks across diverse domains, ranging from web navigation~\citep{zheng2023synapse, zhang2024ufo, wu2024copilot, wang2024mobile} and robotics~\cite{ahn2022can, huang2022inner, hu2023look, driess2023palm} to video games~\citep{wang2023describe, wang2023voyager, ma2024large, tan2024cradle}. While effective at high-level reasoning and planning, these agents struggle with domain-specific yet essential challenges such as generating precise low-level actions and recognizing fine-grained visual patterns. Moreover, their inefficient inference leads to high latency, making it difficult to meet the real-time requirements of interactive environments.
Beyond prompt-based methods, data-driven training approaches have also been explored. Continue training with large-scale robotic data, VLMs can be converted into VLA models~\citep{zitkovich2023rt, kim2024openvla, black2024pi0, shukor2025smolvla, intelligence2025pi}, which are capable of following instructions and performing a wide range of robotic tasks, demonstrating strong generalization capabilities. A similar paradigm is also applied in GUI agents~\citep{cheng2024seeclick, anthropic2024b, wu2024atlas, xu2024aguvis, qin2025ui, openai2025operator}. More recently, this idea has been extended to game environments~\citep{li2025jarvis, chen2025combatvla}. These works typically rely on pretraining with high-quality instruction-following or reasoning datasets, annotated by human labelers, but this leaves open the risk for continued scaling. In this work, Lumine also applies a similar VLA setting but aims to provide a more efficient and scalable training recipe for general-purpose agents.

\textbf{Environment}. Video games have long served as popular environments for developing AI agents, primarily due to their efficient and low-cost interactions while providing rich and diverse dynamics. Traditional game environments are typically built on games that expose APIs for accessing internal states and actions~\citep{bellemare2013arcade, dosovitskiy2016learning, paquette2016supermario, beattie2016deepmind, leibo2021scalable, johnson2016malmo, fan2022minedojo, berner2019dota, vinyals2019grandmaster, samvelyan2019starcraft, ellis2023smacv2}. However, most environments are limited to fixed maps with constrained dynamics and minimal textual content. The limited volume restricts their applicability for developing broader aspects of intelligence. Moreover, each environment adopts its own conventions for encapsulating observation and action spaces, making it difficult to develop general-purpose agents transferable across environments. It further limits scalability to more games, particularly commercial games, which constitute the vast majority of the market but typically do not provide API access. Commercial games, especially AAA games, however, often feature more realistic physics engines, richer content, and more diverse gameplay, making them especially valuable yet challenging testbeds for general-purpose agents. 
Recent advances~\cite{tan2024cradle,raad2024scaling} demonstrate the feasibility of interacting with arbitrary PC games via human-like interfaces (monitor, mouse and keyboard), substantially extending the reach of AI agents. Building upon this direction, Lumine aims to provide a general solution to develop agents in these challenging video games.

\textbf{Task}. Commercial video games are usually built around carefully designed missions that mirror human learning curricula, progressing from simple to complex challenges and comprehensively evaluating agents across multiple levels of competence, from fundamental skills such as navigation, interaction, and combat to higher-level abilities such as adapting to dynamic environments, leveraging newly acquired knowledge, and composing learned skills to solve novel problems. These missions frequently include long-horizon objectives that may span hours or even days to complete, providing natural and richly structured testbeds for investigating long-term planning and compositional intelligence. Traditional RL agents~\citep{mnih2015human, jaderberg2019human, berner2019dota, vinyals2019grandmaster, wurman2022outracing} typically optimize a single objective within closed environments, exhibiting limited multitask capability and poor open-ended exploration ability. In contrast, VLA models~\citep{zitkovich2023rt, kim2024openvla, black2024pi0, raad2024scaling} can follow instructions to perform diverse tasks, yet remain limited to short horizons of only a few seconds to minutes. Recently, prompt-based reasoning agents have achieved remarkable progress, successfully completing one-hour main storyline missions in Red Dead Redemption 2~\citep{tan2024cradle} and full playthroughs of Pokémon Red~\citep{hershey2025claude, zhang2025gemini}. These systems exhibit task composition, reflection, and contextual reasoning over extended periods of time. Building upon these advances, Lumine aims to combine the instruction-following versatility of VLA models with the long-horizon autonomy demonstrated by prompt-based agents, pursuing a unified framework capable of accomplishing diverse, extended missions through reasoning-driven planning and adaptation.

\textbf{Reasoning}. Some robotics VLAs adopt a hierarchical architecture, where one model performs high-level reasoning at a low frequency, and another model generates low-level actions based on that reasoning~\citep{shi2025hi, bjorck2025gr00t, figure2025helix}. Although this structure provides temporal abstraction, it is challenging to optimize both stages jointly and stably due to non-stationarity~\cite{hutsebaut2022hierarchical}. In contrast, other approaches~\citep{xu2024aguvis,qin2025ui,hershey2025claude, zhang2025gemini,intelligence2025pi} follow the ReAct paradigm~\citep{yao2023react}, where the agent performs explicit reasoning and outputs an action at every step. While straightforward, this design can be computationally inefficient and prone to hallucinations in continuous, high-frequency control settings.
To combine the strengths of both paradigms, we draw inspiration from hybrid thinking~\citep{hershey2025claude, qwen2025qwen3}, where LLMs can flexibly decide whether to perform explicit reasoning or directly output an action based on the context. Lumine is trained in an end-to-end manner and is capable of generating reasoning only when necessary, seamlessly coordinating reasoning and control.

\textbf{Memory}. Recent VLA models~\citep{zitkovich2023rt, kim2024openvla, black2024pi0, figure2025helix} and data-driven game agents~\citep{baker2022videopretrainingvptlearning, lifshitz2023steve, raad2024scaling} typically operate in a purely reactive, single-step manner that consumes only the current observation. While this design simplifies both training and inference, it inherently suffers from partial observability, limiting temporal coherence and hindering performance on long-horizon tasks. In contrast, prompt-based agents~\cite{wang2023describe, wang2023voyager, tan2024cradle, zhang2025gemini} exploit extended context windows to retain historical information and periodically summarize trajectories into natural language, enabling stronger long-horizon competence.  We posit that this paradigm can benefit data-driven approaches as well. Lumine makes an initial step in this direction: it maintains recent observations as short-term memory and leverages reasoning to summarize the past and plan future goals as long-term memory. This "context as memory" design allows us to study how an extended context window influences temporal coherence and action consistency without introducing specialized memory modules.

\textbf{Interface}. To interact with arbitrary video games and software applications, a human-style interface, receiving pixel inputs from the screen and using mouse and keyboard for control, serves as the most unified and standardized approach~\cite{tan2024cradle}. Typical GUI agents~\citep{xie2025osworld, xu2024aguvis, qin2025ui} adopt this paradigm, but often oversimplify input modeling. For instance, most agents ignore mouse movement traces: they couple movement with clicks using absolute positioning, effectively teleporting the cursor to the target location before issuing a click. While sufficient for typical websites or desktop applications, this abstraction fails in video games, where the trajectory and dynamics of motion are critical. In 3D first- or third-person games, for example, the mouse directly controls the camera, making relative movement indispensable. Similarly, mouse trajectories may themselves carry meaning, such as drawing gestures or simulating physical interactions, that absolute teleportation cannot capture.
A comparable limitation also exists in keyboard input. GUI agents usually support only coarse-grained operations like press or type, without modeling finer-grained events such as key down, key up, or hold. Yet in video games, these distinctions are essential: holding a key versus tapping it can trigger entirely different actions, while combinations of key states underlie complex mechanics such as sprinting, crouching, charging, or chaining skill sequences. Without this level of expressivity, agents cannot faithfully reproduce the rich and continuous interaction patterns demanded by gaming environments.
While some recent game agents~\cite{tan2024cradle, li2025jarvis, chen2025combatvla} attempt to address the above issues, their actions are still represented in code-like formats, resulting in inefficiencies. Lumine aims to provide an efficient and accurate solution for modeling keyboard and mouse operations while covering the full spectrum of functionalities.

\textbf{Inference}. Real-time inference poses a great challenge for VLM-based agents, especially in fast-paced video games requiring high-frequency interactions. Efficient real-time inference for VLM-based GUI or game agents with keyboard and mouse control remains largely unexplored. It usually takes GUI agents several seconds to produce a single executable action in an autoregressive manner~\citep{anthropic2024b, wu2024atlas, xu2024aguvis, qin2025ui, openai2025operator}. In robotics, VLAs employ techniques such as action chunking~\citep{zhao2023learning}, flow matching~\citep{black2024pi0}, and action tokenization~\citep{pertsch2025fast} to accelerate policy learning and action generation. While these methods are promising to be adapted to game agents,  the semantic nature of actions must be carefully considered. Unlike robotics, where low-level actions are usually with limited semantic meaning and trained from scratch, mouse movements and key presses carry clear semantic intent that is naturally interpretable by VLMs. Lumine provides a practical solution by combining these techniques with traditional LLM inference optimization strategies and efficient action modeling, enabling autoregressive models to achieve real-time inference in gaming environments.

%% file: tables/methods_comparison.tex
{
\renewcommand{\arraystretch}{1.2}
\begin{table}[t]
\caption{Comparison between Lumine and other representative game agents in terms of environment open-endedness, the longest task duration achievable (measured by the average time required for a human player to complete), multimodal understanding (vision and text), ability to follow instructions to accomplish diverse tasks, reasoning capability, real-time inference, and the interface used for interaction (K\&M denotes keyboard and mouse).}
\label{tab:related_work}
\centering
\small
\setlength{\tabcolsep}{3pt}
\begin{tabular}{c|ccccccc}
\hline
\textbf{Method} & \textbf{Open-World}   & \textbf{Task Horizon} & \textbf{\begin{tabular}[c]{@{}c@{}}Multimodal\\ Understanding\end{tabular}} & \textbf{\begin{tabular}[c]{@{}c@{}}Instruction \\ Following\end{tabular}} & \textbf{Reasoning}                 & \textbf{Real-Time}    & \textbf{Interface} \\ \hline
DQN~\citep{mnih2015human}             & \xmark & 5 mins                & \xmark                                                        & \xmark                                                     & \xmark              & \cmark & APIs               \\
AlphaStar~\citep{vinyals2019grandmaster}       & \xmark & 15 mins               & \xmark                                                        & \xmark                                                     & \xmark              & \cmark & APIs               \\
OpenAI Five~\citep{berner2019dota}     & \xmark & 45 mins               & \xmark                                                        & \xmark                                                     & \xmark              & \cmark & APIs               \\
VPT~\citep{baker2022videopretrainingvptlearning}             & \cmark & 20 mins               & \xmark                                                        & \xmark                                                     & \xmark              & \cmark & K\&M               \\
Voyager~\citep{wang2023voyager}         & \cmark & 20 mins               & \xmark                                                        & \cmark                                                     & Stepwise \Always    & \xmark & APIs               \\
Cradle~\citep{tan2024cradle}          & \cmark & 1 hr                  & \cmark                                                        & \cmark                                                     & Stepwise \Always    & \xmark & K\&M               \\
SIMA~\citep{raad2024scaling}            & \cmark & 10 secs               & \cmark                                                        & \cmark                                                     & \xmark              & \cmark & K\&M               \\ 
CombatVLA~\citep{chen2025combatvla}       & \xmark & 1 min                 & \cmark                                                        & \xmark                                                     & Stepwise \Always    & \xmark & K\&M               \\
JAVIS-VLA~\citep{li2025jarvis}       & \cmark & 10 secs               & \cmark                                                        & \cmark                                                     & \xmark              & \xmark & K\&M               \\ \hline
Lumine (Ours)         & \cmark & 5 hrs              & \cmark                                                                     & \cmark                                                     & Adaptive \OnDemand & \cmark & K\&M               \\ \hline
\end{tabular}
\end{table}
}

%% file: sections/3_preliminary.tex
\section{Environment}
\begin{figure}
    \centering
    \includegraphics[width=\linewidth]{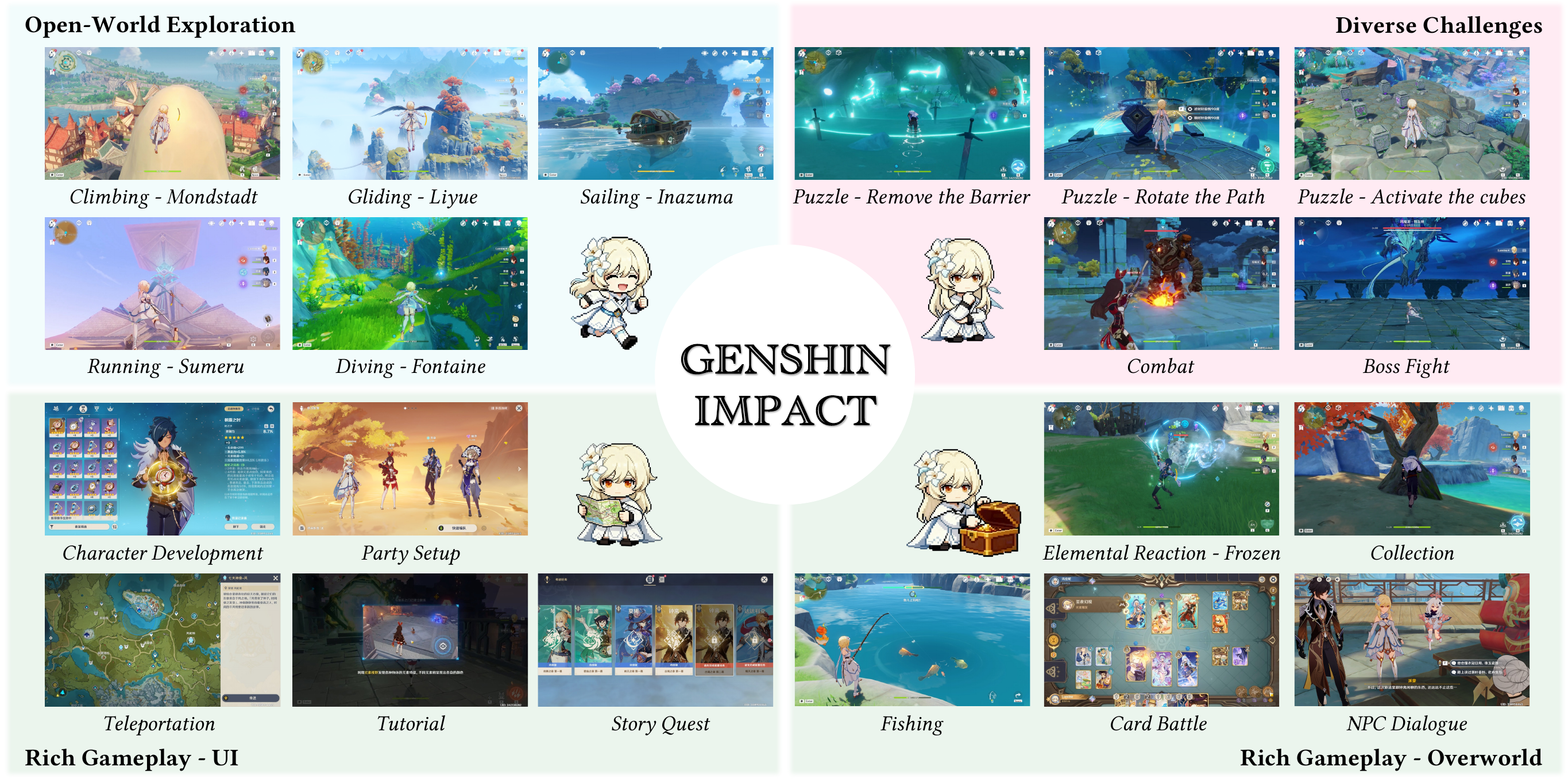}
    \caption{
Overview of the gameplay environment in \textit{Genshin Impact}. 
The game combines large-scale open-world exploration and multi-level reasoning challenges within a richly interactive 3D environment. Players can freely traverse diverse regions, glide, swim, dive, and interact with characters while engaging in quests, puzzles, and combat.
}
    \label{fig:environment}
\end{figure}

Agents must learn from interaction with diverse dynamics to achieve general-purpose capabilities. Video games have emerged as popular platforms for developing such agents due to their efficient interaction and rich dynamics. Among these, commercial video games stand out since they typically feature high-quality graphics, offer expansive open worlds and complex systems that require nuanced and timely decision-making. While these games provide ideal environments for developing general-purpose agents, they remain largely unexplored due to their closed-source nature. To bridge this gap, we select Genshin Impact, a globally popular commercial video game, as a representative case, aiming to present a general and scalable solution for developing agents in such challenging, content-rich environments, shown in Figure~\ref{fig:environment}.  

As an online, third-person, action role-playing open-world game with hundreds of hours of playthrough content, Genshin Impact has incorporated most of the common challenges found in video games, making it both an ideal and demanding testbed for our vision:

\textbf{Open-World Exploration.}  
Genshin Impact offers players a vast and highly interactive open world that blends high-fidelity natural beauty, rich cultural landscapes, and immersive gameplay. Powered by high-quality graphics, realistic physics, and richly detailed environments, the world draws inspiration from real-world geography and features diverse landforms, such as towering mountains, tropical forests, expansive deserts, frozen tundras, and vibrant coastlines, combined with dynamic weather and lighting conditions, all crafted with fine-grained environmental modeling. Players are granted extensive freedom of movement, including running, jumping, climbing, gliding, swimming, sailing and diving, requiring sophisticated spatial understanding and 3D navigation. Fully traversing and understanding the environment typically requires weeks or even months of continuous exploration, reflecting the game’s exceptional scale and structural complexity. Beyond its natural terrains, Genshin Impact features multiple nations and urban regions, each characterized by distinct cultural aesthetics, architectural styles, and environmental motifs. This diversity enriches the overall world structure and contributes to the game’s function as a comprehensive testbed for open-world exploration and embodied intelligence research.

\textbf{Long-Horizon Progression}. In the game, players assume the role of a traveler undertaking an extensive journey across seven nations in search of their lost sibling. This adventure spans hundreds of hours through main story missions and countless side quests, many of which demand sustained effort and long-term engagement. Character progression and team development evolve gradually over weeks or even months, requiring continuous planning, strategic resource allocation, and deliberate decision-making. These long-horizon tasks stand in contrast to the short-term objectives typically emphasized in traditional research benchmarks~\citep{bellemare2013arcade, beattie2016deepmind, kolve2017ai2, puig2018virtualhome, samvelyan2019starcraft}, positioning Genshin Impact as an ideal environment for examining extended strategic planning and complex decision-making.

\textbf{Diverse Gameplay}. As a role-playing game, Genshin Impact encompasses the full spectrum of core gameplay elements characteristic of the genre: expansive 3D open-world exploration, real-time combat, map navigation, dungeon challenges, interactive NPC dialogue, quests completion, team composition, character progression, equipment collection, resource management, and crafting systems. Beyond these foundation components, the game continuously broadens its scope through a wide range of mini-games and special events, ranging from housing systems, card battles and auto-chess to rhythm games, tower defense, and hide-and-seek, effectively covering most common forms of both 3D and 2D interaction found in modern video games. This extensive gameplay diversity presents an exceptionally rich environment for generalization research, allowing for the study of how agents can acquire, transfer, and adapt skills across heterogeneous tasks within a single, coherent virtual world. Consequently, Genshin Impact serves as a promising platform for exploring the development of general-purpose intelligence that extends beyond domain-specific competence.

\textbf{Rich Puzzles}. A distinct characteristic of Genshin Impact lies in the richness and variety of its puzzles distributed throughout the game’s expansive world. 
These puzzles extend far beyond conventional logic-based challenges, integrating elements of exploration, observation, environmental interaction, and strategic reasoning. 
Players must interpret terrain and environmental clues, manipulate mechanisms, and master elemental mechanics to activate or combine triggers.
Many puzzles emphasize temporal and spatial coordination, requiring players to activate devices or eliminate enemies within limited time windows, glide or climb to reach distant targets. Others encourage close observation and memory, as players must recall previously encountered symbols, device states, NPC hints or environmental patterns that serve as clues for subsequent actions. Larger multi-stage puzzles, which are often embedded within ancient ruins or region-specific questlines, demand reasoning across extended spatial and causal structures, such as understanding dependencies among multiple devices or predicting the effects of sequential activations. This multi-layered, multidimensional design constitutes a comprehensive test of perception, logic, and execution, providing a challenging yet structured environment for studying embodied reasoning, spatial cognition, long-term memory and adaptive problem-solving.

\textbf{Comprehensive Guidance}. As a globally popular video game, Genshin Impact offers a well-structured onboarding experience with detailed tutorials and beginner-friendly guidance. Whenever players encounter a new gameplay mechanic or system for the first time, the game provides clear and accessible instructions directly on screen. These tutorials are also archived within the game for easy reference, allowing players to revisit key information as needed. This built-in system not only supports players but also greatly benefits agents designed to learn from or interact with the game. Furthermore, the difficulty curve is carefully calibrated: challenges progress gradually, and gameplay features are unlocked step by step, making the overall experience highly approachable for beginners starting from scratch and providing a natural curriculum for agents to acquire skills in a staged and progressive manner.

\textbf{Large Population Base}. As a globally popular game, Genshin Impact has cultivated a massive and diverse global player base, facilitating large-scale participant recruitment and offering abundant opportunities for gameplay data collection and content annotation. The game further benefits from a vibrant online community that actively produces a wide array of guides, walkthroughs, tutorials and analytical discussions. These resources not only support human players but also constitute valuable auxiliary data sources for agent learning.

While the in-game content already presents diverse content and sufficient challenges, developing autonomous agents within such commercial video games introduces additional systematic difficulties~\citep{raad2024scaling}:
\begin{itemize}
    \item Commercial video games usually do not expose APIs for interaction. Agents must rely on the standard interface, i.e., receiving raw pixel images from the screen and issuing keyboard and mouse inputs for control.
    \item Unlike traditional research environments~\citep{bellemare2013arcade, beattie2016deepmind, kolve2017ai2, puig2018virtualhome, samvelyan2019starcraft, fan2022minedojo}, where environments are frozen during action generation,  agents must act under strict latency constraints and execute actions asynchronously. 
    \item Internal states and reward signals are usually inaccessible. Moreover, reliance on GPU rendering limits scalability to thousands of parallel rollouts. These challenges make reinforcement learning particularly difficult and highlight the need for more efficient approaches.
\end{itemize}

%% file: sections/4_methods.tex
\section{The Lumine Model}
\label{sec:model}

\begin{figure}[h]
    \centering
    \includegraphics[width=1\linewidth]{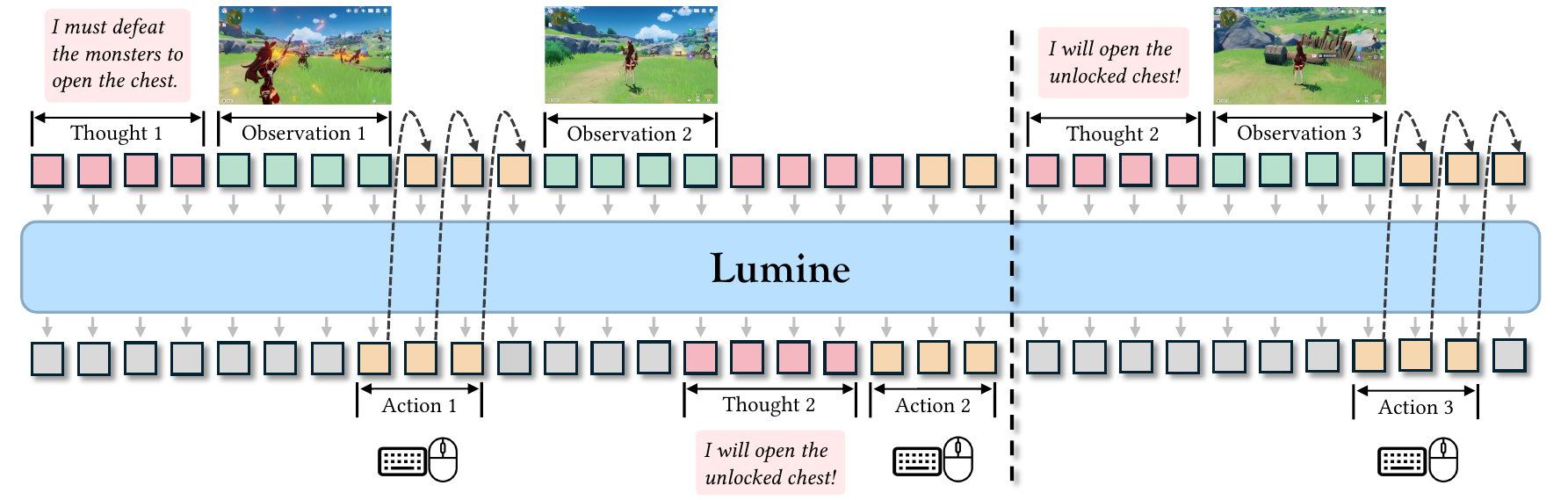}
    \caption{Overview of the Lumine model. Built upon a VLM, Lumine receives pixel inputs along with historical context, such as previous actions and reasoning, and outputs textual keyboard and mouse actions. It employs a hybrid reasoning strategy, generating new reasoning steps only when necessary; otherwise, it directly produces actions for efficient real-time control. }
    \label{fig:env}
\end{figure}

As illustrated in Figure~\ref{fig:env}, we introduce Lumine, a 7B-parameter model designed to process sequences of raw pixel images as inputs and generate executable keyboard and mouse operations, accompanied by interpretable intermediate reasoning. Lumine builds upon Qwen2-VL-7B-Base~\citep{wang2024qwen2}, inheriting its strong multimodal understanding and generation capabilities gained from large-scale pretraining on diverse web data. By augmenting this foundation with an explicit reasoning and action prediction mechanism, Lumine is capable of closed-loop visual decision-making within interactive environments.

At each timestep $t$, the model first determines whether to enter a thinking mode to generate reasoning $r_t$, conditioned on its historical visual observations $o_{\leq t}$ as well as prior reasoning traces $r_{<t}$ and actions $a_{<t}$. If the model decides not to reason explicitly, $r_t$ is set to null. Subsequently, it predicts the next executable action $a_t$. Formally, the model $\pi_\theta$ captures the joint distribution over reasoning and actions as:
\begin{equation}
    \pi_\theta(a_t, r_t \mid o_{\leq t}, r_{<t}, a_{<t})
    = \pi_\theta(a_t \mid o_{\leq t}, r_{\leq t}, a_{<t}) \,
      \pi_\theta(r_t \mid o_{\leq t}, r_{<t}, a_{<t}),
    \label{eq:reasoning_then_action}
\end{equation}
This factorization reflects the model's perceive-reason-action paradigm, 
where intermediate reasoning provides an explicit latent structure that guides the subsequent action generation. 
Next, we provide a concrete definition of Lumine's observation and action space.

\begin{table}[t]
\small
\centering
\caption{Examples of Lumine switching between thinking and non-thinking modes by generating \texttt{<|thought\_start|>} or \texttt{<|action\_start|>} as the first token at each step. Historical information is not shown for simplicity.}
\label{tab:thinking_modes}
\begin{tabular}{@{}p{6cm}|p{6cm}@{}}
\toprule
\textbf{Thinking Mode} & \textbf{Non-Thinking Mode} \\ \midrule
\texttt{<|im\_start|>user} \newline
\texttt{\{image\}} \texttt{<|im\_end|>} \newline
\texttt{<|im\_start|>assistant} \newline
\texttt{<|thought\_start|>} \newline
\textcolor{red}{\texttt{\{reasoning\_content\}}}  \newline
\texttt{<|thought\_end|>} \newline
\texttt{<|action\_start|>} \newline
\textcolor{blue}{\texttt{\{action\_content\}}}  \newline
\texttt{<|action\_end|>}\texttt{<|im\_end|>} &
\texttt{<|im\_start|>user} \newline
\texttt{\{image\}} \texttt{<|im\_end|>} \newline
\texttt{<|im\_start|>assistant} \newline
\texttt{<|action\_start|>} \newline
\textcolor{blue}{\texttt{\{action\_content\}}}  \newline
\texttt{<|action\_end|>}\texttt{<|im\_end|>} \\
\bottomrule
\end{tabular}
\end{table}

\textbf{Observation Space.} Lumine perceives continuous visual input from the game environments. Each frame is resized to 1280$\times$720 (720p), a resolution chosen to balance UI text legibility and computational efficiency. To align with human visual reaction time of roughly 200–250 ms \citep{woods2015factors} and to avoid missing critical timing events, Lumine processes one observation frame every 200 ms (5 Hz). In addition to visual inputs, Lumine maintains a history of past reasoning and actions as contextual information, preserved in the form of multi-turn dialogues interleaved across model outputs with the visual inputs.

\textbf{Hybrid Thinking.} 
As shown in Table~\ref{tab:thinking_modes}, at every step, Lumine first optionally produces intermediate reasoning as inner monologue and then generates the executable keyboard and mouse actions.  Lumine adopts a hybrid thinking mode, engaging in explicit reasoning only when necessary, while continuously predicting actions that align with its ongoing thought process. Each reasoning phrase is enclosed by the special tokens \texttt{<|thought\_start|>} and \texttt{<|thought\_end|>}, explicitly marking the boundaries between the model’s inner monologue and actions. The reasoning acts both as a reflection on previous behavior and as a plan for subsequent steps. Reasoning typically emerges at critical transitions, such as when sudden environmental changes cause prior plans invalid and adjustments are required, or when a task has been completed and new goals need to be proposed.

\textbf{Keyboard and Mouse Modelling.} While prior works often introduce additional action heads~\citep{black2024pi0, intelligence2025pi} or redefine the vocabulary~\cite{zitkovich2023rt, li2025jarvis} to represent actions, they fail to exploit the inherent semantics of keyboard and mouse operations, which are well captured by LLMs. Other approaches~\cite{tan2024cradle, chen2025combatvla, qin2025ui, li2025jarvis} model such interactions in code formats, which tend to be verbose and inefficient for high-frequency interactions. Motivated by these limitations, Lumine introduces a concise and efficient action representation that allows models to autoregressively generate both high-level reasoning and low-level control signals without modifying the model architecture or vocabulary, enabling seamless integration across different LLM-based agents. We unify all keyboard and mouse actions within the language space, covering the full spectrum of functionalities, which we formally define as follows:

\newcommand{\narrowDelta}{\scalebox{0.6}[1.0]{$\Delta$}}
    \[
    \underbrace{\narrowDelta X~\narrowDelta Y~\narrowDelta Z}_{\text{Mouse movements}}
    ~;~
    \underbrace{K_1~;~K_2~;~K_3~;~K_4~;~K_5~;~K_6}_{\text{Key presses}}
    \]
\begin{itemize}
    \item \textbf{Format.}
    Each action is enclosed by special tokens \texttt{<|action\_start|>} and \texttt{<|action\_end|>}. Internally, the action consists of a semicolon-separated sequence of components that specify mouse movements ($~\narrowDelta X~\narrowDelta Y~\narrowDelta Z$), and a series of key presses ($~K_1~\texttt{;}~K_2~\texttt{;}~K_3~\texttt{;}~K_4~\texttt{;}~K_5~\texttt{;}~K_6$). 
    \item \textbf{Mouse Movement.}
    We discretize the mouse action space. Lumine predicts the \textit{relative displacement} $(\narrowDelta X, \narrowDelta Y)$ as integer values within the range $(-1000, 1000)$, along with an associated scroll value $\narrowDelta Z$ as an integer in $[-5, 5]$ representing the number of scroll steps. The full predicted movement is then executed smoothly over a 200 ms interval to ensure execution efficiency. 
    \item \textbf{Key Presses.} To capture fine-grained dynamics while maintaining computational efficiency, Lumine adopts an action chunking strategy~\citep{zhao2023learning}. At each step, the model predicts six consecutive action chunks over a 200~ms window, with each chunk lasting 33~ms, resulting in a 30 Hz interaction frequency. An action chunk $K_t$ specifies zero to four keys, including both keyboard inputs and mouse buttons, that are pressed during this interval. Any keys not listed in the chunk are automatically released. Keys that appear in consecutive chunks retain their key-down state and are not pressed again. The choice of a 33~ms granularity is supported by empirical evidence: we observe that the minimal keyup–keydown interval is around 40~ms, which is consistent with the reported lower bound of approximately 60~ms for inter-key intervals (keydown+keyup)~\citep{dhakal2018observations}. This action modeling allows Lumine to faithfully mimic player behavior and accurately execute a wide range of complex interactions, such as long and short key presses, key combinations, rapid clicking, drag-and-drop gestures, or simultaneous multi-key actions typical in fast-paced scenarios such as combat. For efficient inference, each key is represented by a single token, shown in Appendix Table~\ref{app_tab:key_mapping}.
\end{itemize}

A possible instantiation of such an action could be: "92 0 0 ; Shift W ; Shift W ; Shift W ; F W ; F W ; F". This action depicts the agent dashing (\texttt{Shift} \& \texttt{W}) toward a treasure chest on the right (mouse turn right \texttt{92} units), then stopping upon reaching it and attempting to open the chest (\texttt{F}).

\begin{figure}[h]
    \centering
    \includegraphics[width=0.9\linewidth]{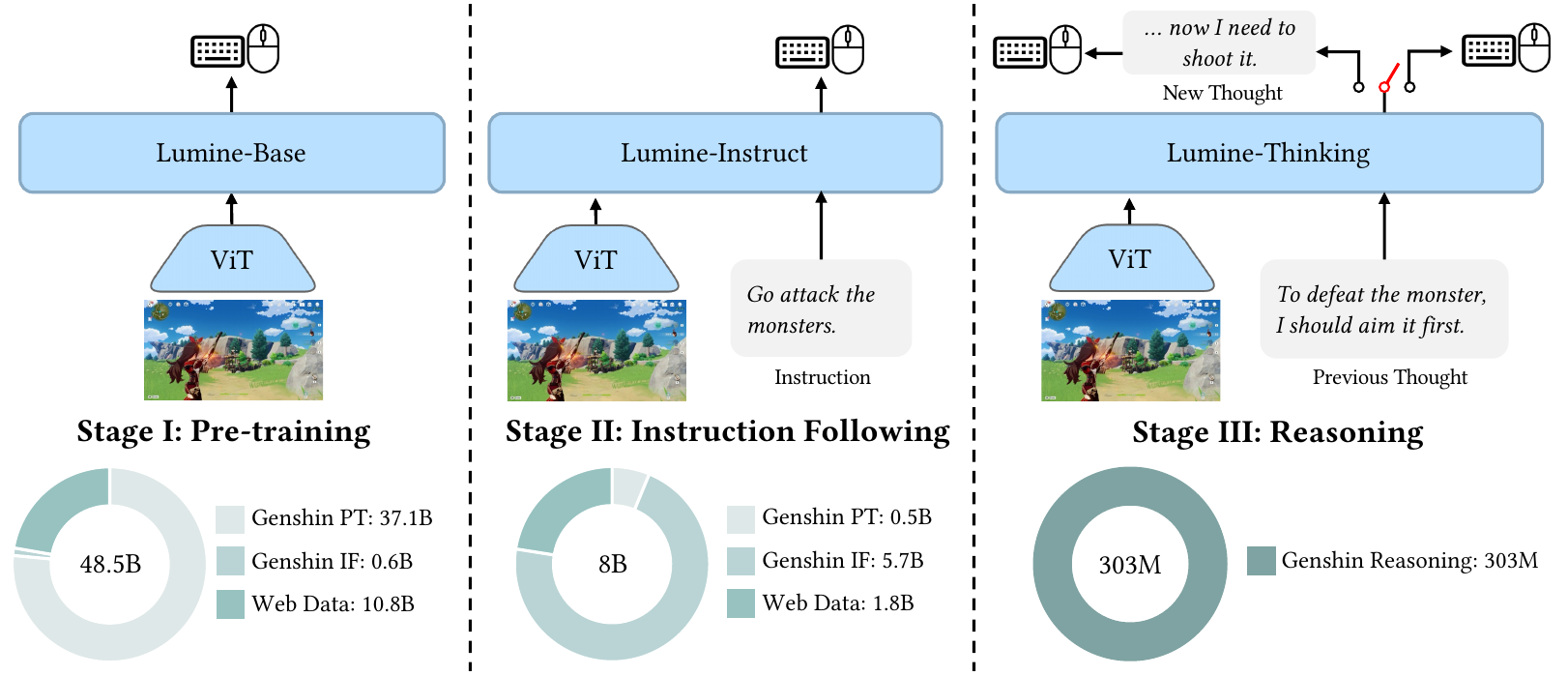}
    \caption{Overview of Lumine’s three-stage training recipe. In the first pre-training stage, Qwen2-VL-Base is trained on large-scale image–action data to learn fundamental action primitives, resulting in the Lumine-Base model. In the second instruction-following stage, Lumine-Base is further trained on instruction–image–action triplets for language grounding, producing the Lumine-Instruct model. In the final reasoning stage, the instruction input is replaced with a thought, and an optional new thought is prepended before the action sequence, yielding the Lumine-Thinking model.}
    \label{fig:method_overview}
\end{figure}

\begin{figure}[!htbp]
    \centering
    \includegraphics[width=0.9\linewidth]{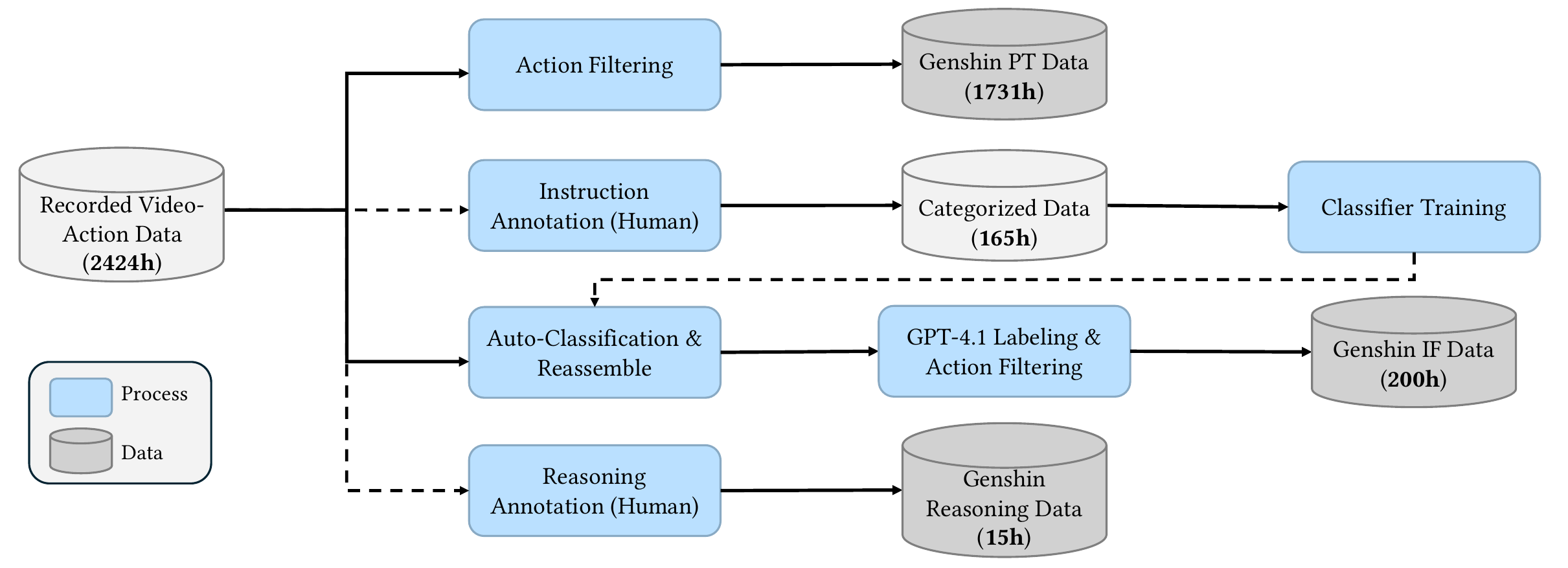}
    \caption{Overview of the data processing pipeline from raw gameplay recordings to curated datasets for pre-training, instruction following, and reasoning. i) Starting from 2424 hours of synchronized video-action data, we first apply rule-based filtering to produce a 1731-hour dataset for pre-training. ii) A subset of 165 hours is human-annotated for instruction-level activities, used to train a classifier that auto-labels all the raw data, further refined into 200 hours of high-quality instruction following data via GPT-4.1 captioning and action filtering. iii) Meanwhile, 15 hours of manually annotated reasoning data support the training of Lumine’s hybrid thinking. Together, this multi-stage curation pipeline enables scalable, structured curriculum learning from human demonstrations.}
    \label{fig:data_overview}
\end{figure}

\section{Data Curation and Training Recipe}
In this section, we present an efficient and scalable data curation pipeline and multi-stage training recipe, designed to enable Lumine to perceive, reason, and act as a generalist agent in the challenging 3D open-world environment of Genshin Impact. 

% \textit{Genshin Impact}. 

\textbf{Training Stages}. As illustrated in Figure~\ref{fig:method_overview}, we employ a multi-stage training procedure for our model. In the first pre-training stage, the model learns to act from raw observations across diverse scenarios, enabling robust and reactive control behaviors. In the second instruction-following stage, the agent’s actions are grounded in natural language, allowing it to follow textual instructions to complete short-horizon tasks. Finally, in the third reasoning stage, the model learns to perform explicit reasoning that guides subsequent action generation, supporting complex, long-horizon decision making. 

\textbf{Raw Data}. Figure~\ref{fig:data_overview} demonstrates that we start from a collected dataset comprising 2424 hours of human gameplay, consisting of synchronized video streams paired with recorded keyboard and mouse operations. Contractors were tasked to start with a brand-new account and progress through the entirety of Mondstadt (the first nation in Genshin Impact), completing its main storyline and achieving over 80\% map exploration using only system-provided characters. On average, this process requires approximately 30 hours of playtime for a human player. Comprehensive details on data collection and curation are provided in Appendix ~\ref{app:data_collection}. 

\subsection{Pre-Training}
Unlike prior works~\citep{kim2024openvla,black2024pi0,raad2024scaling,li2025jarvis} that rely on large-scale instruction-following data for pre-training, our approach focuses on exposing models to diverse in-game dynamics by primarily training on image–action pairs without additional labels. This design choice is motivated by two key considerations.
i) Human annotation is both costly and difficult to scale, while automatic labeling using VLMs remains unreliable due to their limited long-horizon understanding and insufficient domain knowledge. In contrast, raw video–action data are naturally abundant and straightforward to collect.
ii) In open-world exploration settings, it is inherently difficult to assign precise labels to every gameplay segment, as players may wander aimlessly or perform suboptimal and sometimes even confusing actions. These seemingly irregular behaviors, however, capture valuable corner cases that significantly enhance the model’s robustness and generalization. 

Based on these considerations, we apply rule-based filtering to remove 95\% of idle actions and clips dominated by camera jitter, resulting in 1731 hours of high-quality gameplay data. The remaining data are then fully utilized for Lumine’s pre-training. To preserve the broad perceptual and reasoning capabilities of the base model during this action-centric training, which are essential for downstream instruction following and reasoning generation, we incorporate an approximately 20\% mixture of multimodal web data to retain general knowledge. Additionally, we include a small amount of instruction-following data, which is reserved for evaluation purposes. This stage yields the Lumine-Base model, which serves as the foundation for subsequent language grounded.

\subsection{Instruction Following}

Building on the action primitives learned during pre-training, we align action prediction with language through a modest amount of instruction-following data. To achieve this, we adopt a label-then-augment strategy that transfers the generalization capability of large vision-language models into the action domain.

We first collect 165 hours of human-annotated instruction data from raw gameplay. Annotators were asked to identify the start and end timestamps of 38 predefined activity categories within 20-second video clips. These annotations are used to fine-tune a classifier based on Qwen2-VL-2B, which enables scalable auto-labeling of the all the raw data.

Empirically, we observe that the Lumine-Base model can instinctively interact with nearby objects, NPCs, and enemies, yet struggles with situations that require breaking its current behavioral inertia, for example, navigating to specific target locations. This is precisely where natural language instructions play a critical role, guiding the model to overcome local behavioral priors and execute goal-directed behaviors. To address this, we identify transition points between adjacent gameplay segments that are assigned with different labels by the classifier, typically indicating a shift in task context or objective. Around each transition point, we extract a 20-frame (4s) snippet and prompt GPT-4.1~\citep{openai2025gpt4.1} to generate diverse, context-aware instructions based on the labeled categories.  While the provided category labels supplement GPT-4.1’s limited understanding of game mechanics and objectives, the model also acts as a verifier, detecting and discarding mislabeled samples when inconsistencies are found. 

After applying the same action filtering as in pre-training, we obtain 200 hours of high-quality instruction-following data. These are mixed with multimodal web data in the same ratio as used during pre-training, while retaining a minimal portion of action-only data to preserve behavioral diversity. The resulting model, denoted as Lumine-Instruct, can generate contextually grounded actions in response to textual instructions and serves as the foundation for the subsequent reasoning-stage training.

\subsection{Reasoning}
Building upon Lumine-Instruct’s ability to follow textual guidance, we further enhance the model with explicit reasoning skills for autonomous exploration and long-horizon decision-making. To achieve this, we curate a specialized dataset of human-annotated inner monologues. We select the first act of Genshin Impact’s main storyline in Mondstadt,  \textit{Prologue: Act I - The Outlander Who Caught the Wind}, as a testbed and sample 27 gameplay videos from the raw data in which human players are engaged in this mission. Annotators are provided with consecutive 10-second clips and instructed to identify key decision points at the frame level, then write first-person thoughts that articulate the underlying rationale behind actions in a concise and accurate tune. This process yields a high-quality dataset of 15 hours gameplay, containing 15K reasoning traces, with an average interval of 3.2 seconds between consecutive thoughts. Each reasoning sequence contains an average of $37.4 \pm 11.7$ tokens, achieving a balance between detail and brevity.

To better align with real inference scenarios, we do not apply any action filtering, enabling Lumine to learn to wait appropriately at critical decision points. We then fine-tune Lumine-Instruct on this dataset, resulting in Lumine-Thinking, an autonomous model to complete hours-long missions without human intervention.

\begin{table}[h]
\caption{Hyperparameters and computational resources used by Lumine during the three-stage training process under both non-history and history settings. We apply VeOmni~\cite{ma2025veomni} as our training framework, which dynamically packs batches to match the target sequence length for efficient training.}
\label{tab:hyperparameters}
\small
\setlength{\tabcolsep}{3pt}
\begin{tabular}{lcccccc}
\hline
\textbf{}                                     & \multicolumn{3}{c}{\textbf{Non-history}}                                                                                                   & \multicolumn{3}{c}{\textbf{History}}                                                                                  \\ \hline
\multicolumn{1}{l|}{\textbf{Hyperparameters}} & \textbf{Pre-training} & \textbf{\begin{tabular}[c]{@{}c@{}}Instruction\\ Following\end{tabular}} & \multicolumn{1}{c|}{\textbf{Reasoning}} & \textbf{Pre-training} & \textbf{\begin{tabular}[c]{@{}c@{}}Instruction\\ Following\end{tabular}} & \textbf{Reasoning} \\ \hline
\multicolumn{1}{l|}{LLM Learning Rate}        & 2e-5                  & 2e-5                                                                     & \multicolumn{1}{c|}{1.83e-5}            & 2e-5                  & 2e-5                                                                     & 1.64e-5            \\
\multicolumn{1}{l|}{ViT Learning Rate}        & 7e-6                  & 7e-6                                                                     & \multicolumn{1}{c|}{-}                  & 7e-6                  & 7e-6                                                                     & -                  \\
\multicolumn{1}{l|}{LR Scheduler}             & Constant              & Cosine                                                                   & \multicolumn{1}{c|}{Cosine}             & Constant              & Cosine                                                                   & Cosine             \\
\multicolumn{1}{l|}{Gradient Norm Clip}       & 1.0                   & 1.0                                                                      & \multicolumn{1}{c|}{1.0}                & 1.0                   & 1.0                                                                      & 1.0                \\
\multicolumn{1}{l|}{Optimizer}                & \multicolumn{3}{c|}{AdamW$(\beta_1 = 0.9,\, \beta_2 = 0.95,\, \epsilon = 1.0 \times 10^{-8})$}                                             & \multicolumn{3}{c}{AdamW$(\beta_1 = 0.9,\, \beta_2 = 0.95,\, \epsilon = 1.0 \times 10^{-8})$}                         \\
\multicolumn{1}{l|}{Warm-up Ratio}            & 0.05                  & 0.05                                                                     & \multicolumn{1}{c|}{0.05}               & 0.05                  & 0.05                                                                     & 0.05               \\
\multicolumn{1}{l|}{Batch Packing Length}     & 32768                 & 32768                                                                    & \multicolumn{1}{c|}{32768}              & 32768                 & 32768                                                                    & 32768              \\
\multicolumn{1}{l|}{Batch Size}               & 128                   & 128                                                                      & \multicolumn{1}{c|}{64}                 & 128                   & 128                                                                      & 64                 \\
\multicolumn{1}{l|}{Training Epochs}          & 1                     & 2                                                                        & \multicolumn{1}{c|}{3}                  & 3                     & 3                                                                        & 3                  \\
\multicolumn{1}{l|}{GPU Num (H100)}           & 64                    & 32                                                                       & \multicolumn{1}{c|}{64}                 & 64                    & 32                                                                       & 64                 \\
\multicolumn{1}{l|}{Training Time}            & 3.5 Days              & 1.3 Days                                                                & \multicolumn{1}{c|}{1 Hour}             & 12.4 Days             & 2.2 Days                                                                 & 1 Hour             \\
\multicolumn{1}{l|}{GPU Hours (H100)}         & 5376                  & 960                                                                      & \multicolumn{1}{c|}{64}                 & 19008                 & 1664                                                                     & 64                 \\ \hline
\end{tabular}
\end{table}

\subsection{Training Details}
We explore both non-history (single-frame input) and history (multi-turn input) settings. The data organization varies slightly across the three training stages.

\begin{itemize}
    \item \textbf{Pre-training.} In the non-history setting, each sample consists of a single image paired with its corresponding action. In the history setting, each sample consists of 20 interleaved image-action pairs, resembling the multi-turn conversational structure used in VLMs.
    \item \textbf{Instruction Following.} In the non-history setting, each trajectory is decomposed into single-step samples, where each step includes an instruction, an image, and the corresponding action. In the history setting, training is performed on the entire trajectory sequence.
    \item \textbf{Reasoning.} In the non-history setting, most samples follow the same format as instruction-following data, where the model receives the previous reasoning and current visual observation as input. Only a small subset of data contains segments that generate new reasoning. For history training, a trajectory starts from the first frame after generating a new reasoning and ends upon the next reasoning generation, or earlier if it exceeds 20 frames in length.
\end{itemize}

We adopt VeOmni~\cite{ma2025veomni} as our training framework, and Table~\ref{tab:hyperparameters} presents the hyperparameters and computational resources used across the three-stage training process under both the non-history and history settings. The ViT backbone is kept frozen during the final reasoning stage. Empirically, we observe that during the pre-training stage, multiple training epochs continue to yield improvements in the history setting, whereas the non-history setting tends to overfit after the second epoch. We report the results corresponding to the number of training epochs that achieved the best performance on our benchmark.

\section{Inference}
In this section, we first introduce our inference strategy for context management, followed by the optimizations applied to achieve real-time inference.

\begin{figure}[t]
    \centering
    \includegraphics[width=0.9\linewidth]{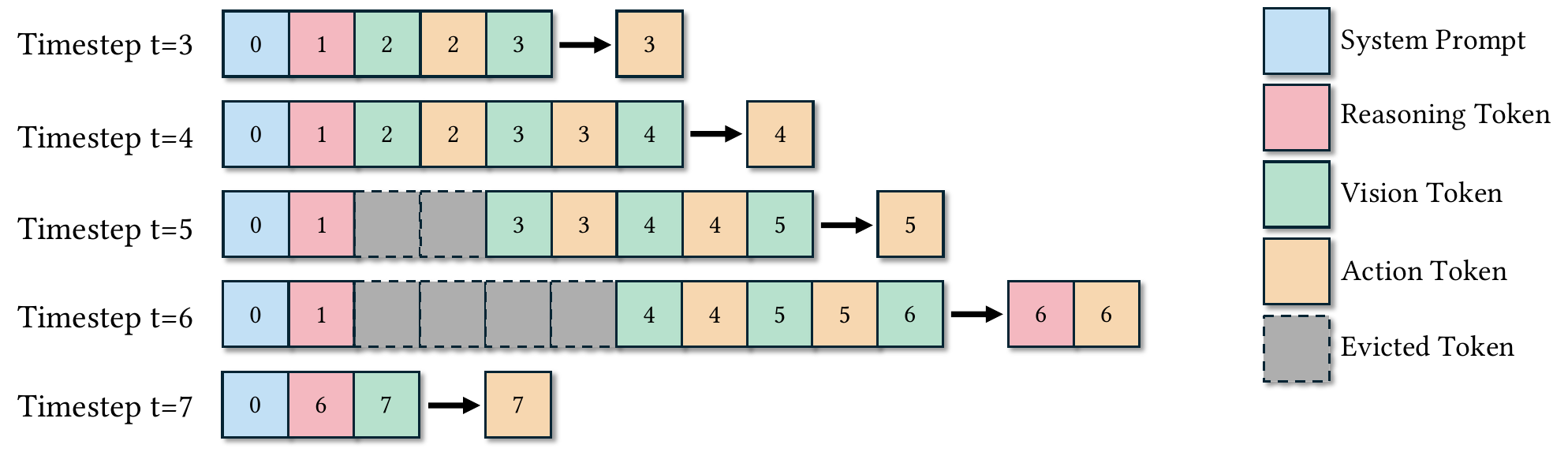}
    \caption{Visualization of the strategy Lumine uses for context management during inference. Lumine maintains a sliding window within the context to preserve image–action pairs across interaction steps, with a maximal window length of 2, as shown in the example. The context begins with the system prompt and previous reasoning, which guide subsequent action generation. When the number of image–action pairs exceeds the threshold, the oldest pair is discarded while retaining the system prompt and reasoning. Upon generating new reasoning, the context is flushed and re-accumulated from that point onward. }
    \label{fig:inference_context}
\end{figure}

\subsection{Context Management}
Figure~\ref{fig:inference_context} illustrates how Lumine manages context during inference using a sliding-window mechanism. In the history setting, the sliding window length is set to 20, whereas in the non-history setting, it is set to 1. At the beginning of the context are the system prompt and previous reasoning, which together guide subsequent action generation. These are followed by image–action pairs maintained in a multi-turn dialogue format. A first-in-first-out (FIFO) policy is applied: when the number of pairs exceeds the maximum threshold, the oldest image–action pair is discarded while retaining the system prompt and reasoning. When new reasoning is generated, the context is flushed and re-accumulated from that point onward. 
By applying this strategy, Lumine effectively uses previous reasoning steps as long-term memory and a 20-frame context as short-term memory, thereby maintaining the consistency and coherence of the model’s behavior during continuous interactions. In this work, we preserve only the most recent reasoning within the context, while the mechanism can be easily extended to maintain multiple reasoning segments if needed. We set the temperature to $1$ and $\texttt{top\_p}$ to $1$ for all inference settings.

\begin{figure}[t]
    \small
    \centering
    \includegraphics[width=1\linewidth]{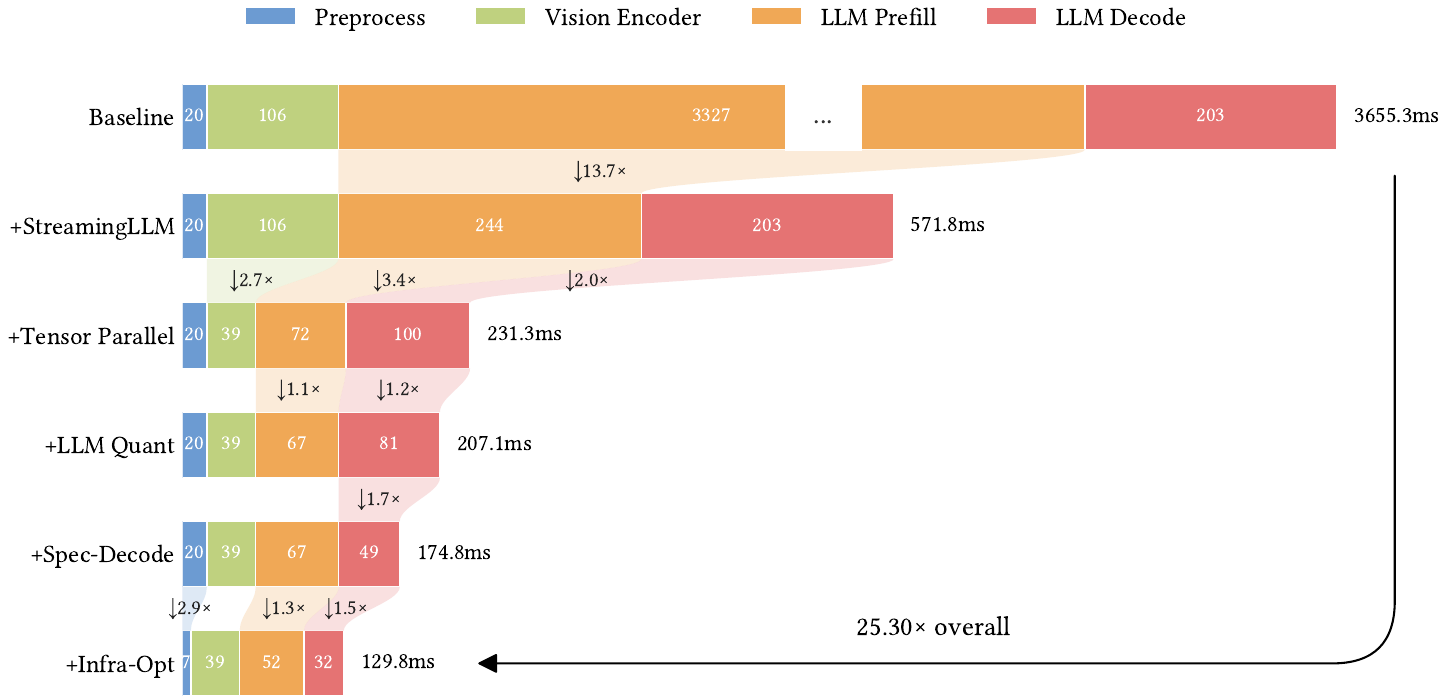}
    \caption{Latency breakdown by stage with corresponding ribbons and overall improvement. The figure shows time latency under different strategies for a typical 20-token action without reasoning generation and with the full context of 20 frames. Infra-Opt denotes the remaining infrastructure-level optimizations. The overall optimization yielding a 25.3× speedup compared with the baseline.}
    \label{fig:infra}
\end{figure}

\subsection{Real-Time Optimization}

Real-time operation is essential for continuous, closed-loop control within dynamic environments that require a high frequency of interaction. Achieving real-time interaction between Lumine and the game presents significant challenges for a 7B-parameter model.  A typical interaction involves two machines: a local Windows host running the game client and a remote server responsible for model inference. Communication between the two machines is conducted over the network. 

At an interaction rate of 5 Hz, the system must complete the following within 200 ms in each control cycle:
i) the host captures a screenshot of the game and sends an inference request to the model server;
ii) upon receiving the request, the server performs inference and returns an action string; and
iii) the host parses the action string into corresponding keyboard and mouse events and executes them. To achieve this level of responsiveness, we developed a series of optimizations across communication and model inference (prefill \& decoding). Figure~\ref{fig:infra} provides a detailed latency breakdown with different optimization strategies.

\textbf{Communication.}
We employ a streaming output mechanism for timely action execution. Lumine autoregressively generates an entire action sequence composed of six continuous chunks, each corresponding to a 33 ms executable segment of keyboard and mouse operations. Once a complete chunk is produced, signaled by a terminating semicolon, it can be executed immediately without waiting for the full sequence to finish. This substantially relaxes the strict timing constraint: as long as each action chunk is generated within 33 ms, the remaining 200 ms interval can be fully utilized for processing stages. Additionally, to reduce data payload, image observations are compressed into JPEG format, and then Base64-encoded before being transmitted. A persistent TCP connection is maintained to avoid repeated handshaking costs.

\textbf{Prefill.} In the history setting, we maintain and reuse the historical key–value (KV) cache from prior interactions for efficiency.  When the context window becomes saturated and the earliest turns need to be dropped, we adopt the StreamingLLM technique~\cite{xiao2023efficient} to maximize KV-cache reuse. An attention sink is anchored to the system prompt to prevent collapse, enabling a stable attention window for efficient inference. Note that we observed a performance degradation in long-horizon tasks with the use of StreamingLLM, which is effectively alleviated by our context management strategy that clears the context when a new reasoning appears.

\textbf{Decoding.} As discussed in Section~\ref{sec:model}, we minimize the number of decoding steps by designing a compact action space. Additionally, we implement a draft-model-less speculative decoding strategy~\citep{leviathan2023fast} to further reduce token decoding, leveraging fixed delimiters across generation stages. Specifically, we observe that \(\Delta X\) and \(\Delta Y\) end with a space, \(\Delta Z\) and \(K_1,\cdots K_5\) end with a semicolon, and \(K_6\) ends with the \texttt{<|action\_end|>} token. We use a simple state variable to track the generator’s stage to dynamically select the appropriate delimiter as a draft token. Standard reject sampling is then applied to ensure the final output matches the same distribution.

\textbf{Infrastructure.} We further optimize the low-level computational framework to maximize hardware utilization and throughput:
i) \textit{Tensor Parallelism:} Model weights are partitioned across four GPUs. Since Qwen2-VL-7B exposes only four KV heads, we deploy the server on four NVIDIA H20 GPUs with a tensor-parallel (TP) degree of 4, assigning one KV head per GPU. This contributes the a remarkable acceleration.
ii) \textit{Quantization:} We apply W8A8 (8-bit weights and activations) quantization using SmoothQuant~\citep{xiao2023smoothquant} to reduce computation and memory bandwidth demands during both the ViT and LLM prefill stages.
iii) \textit{Kernel and Graph Optimizations:} We perform search-based tuning of GEMM kernels for the ViT, prefill, and decode stages, and introduce a custom one-shot all-reduce kernel to enhance communication efficiency during decoding. With speculative decoding enabled, we further cut CPU latency by capturing a single CUDA graph that fuses the forward pass and rejection sampling process. Finally, image preprocessing is offloaded to the GPU to accelerate data handling and minimize CPU–GPU transfer overhead.

The combined effect of these optimizations significantly reduces end-to-end latency across preprocessing, vision encoding, prefill, and decoding stages, as illustrated in Table~\ref{tab:final_latency}. When reasoning generation is not invoked, the delay before producing the first action chunk is approximately 110 ms, well below the 200 ms threshold. Even in the worst case, when generating the longest action chunk (four keys and a semicolon), the latency remains only 12 ms, comfortably under the 33 ms threshold. This enables Lumine to interact with its environment seamlessly. However, the system is still affected by an asynchrony issue, as it perceives visual inputs that are roughly 200 ms old, which can introduce side effects in highly time-sensitive scenarios. Additionally, during steps where the model outputs reasoning, the latency may exceed 200 ms, resulting in a brief idle period. Nonetheless, due to the low frequency of reasoning events, we empirically observe no noticeable impact on the visual experience, maintaining smooth and stable gameplay. This further highlights the importance of hybrid thinking for balancing efficiency and responsiveness.

\begin{table}[h]
\caption{Inference time of Lumine at each stage with the combined optimizations. In addition to latency, we also report the statistical averages of token usage and forward steps. Due to speculative decoding, the number of forward steps is typically smaller than the number of generated tokens. }
\label{tab:final_latency}
% \small
\centering
\begin{tabular}{lccc}
\hline
\textbf{Stage} & \textbf{Time (ms)} & \textbf{Token} & \textbf{Forward Step} \\
\hline
Network latency & 6 & - & - \\
Preprocessing & 6.8 & - & - \\
Vision encoder & 39 & 1196 & 1 \\
LLM prefill & 52 & 1209 & 1 \\
\hline
Decode latency per token & 3.1 & 1 & - \\
First action chunk w/o reasoning & 113.9 & 8.4 & 4.7 \\
First action chunk w/ reasoning & 234.0 & 46.8 & 43.1 \\
Action chunk (average) & 3.1 & 1.8 & 1.02 \\
Action chunk (max) & 12.4 & 5 & 4 \\
\hline
\end{tabular}
\end{table}

%% file: sections/5_results.tex
\section{Benchmark}
\begin{figure}[t]
    \centering
    \includegraphics[width=\linewidth]{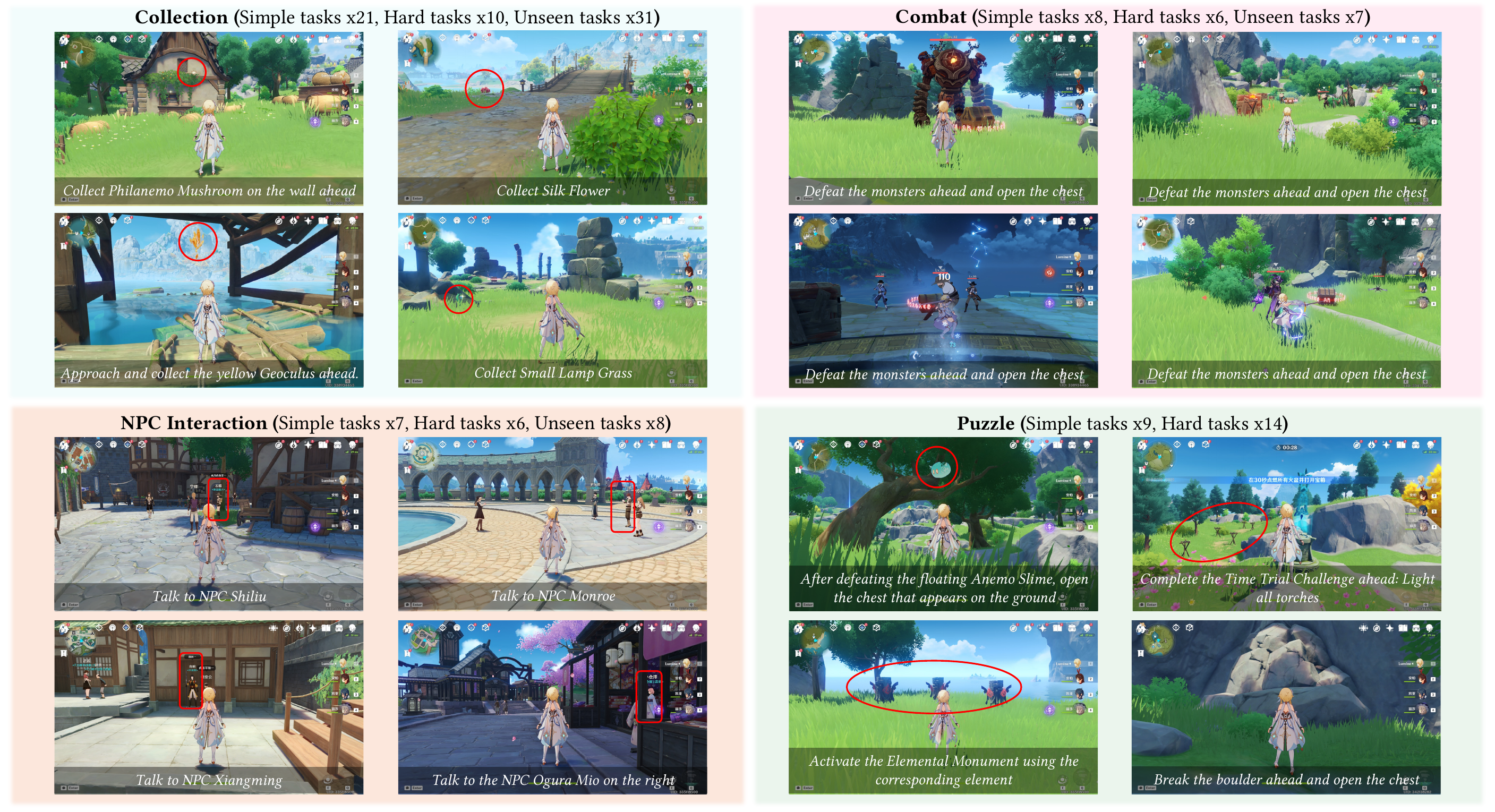}
    \caption{Overview of the benchmark comprising 141 tasks across four categories, Collection, Combat, NPC Interaction, and Puzzle. Each category includes simple, hard, and unseen tasks to comprehensively assess agents' various abilities in open-world gameplay.}
    \label{fig:benchmark}
\end{figure}

To systematically evaluate Lumine's performance in open-world environments, we constructed a comprehensive benchmark of 141 language-conditioned tasks. These tasks are grouped into four categories as shown in Figure~\ref{fig:benchmark}.

\begin{itemize}
    \item \textbf{Collection.} Agents collect items scattered across varied terrains, such as fruits on trees, plants by rivers, flowers on cliffs, treasure chests hidden in grass, and Oculi in the air. Success requires precise object recognition, strong 3D spatial reasoning, and robust navigation despite environmental distractions.

    \item \textbf{Combat.} Agents defeat varied enemy groups to unlock guarded treasure chests. They must adapt to enemy traits and terrain by strategically coordinating their team's skills and synergies. For example, using ranged attacks against elevated foes or exploiting elemental reactions like freezing water-based monsters with ice abilities. After combat, agents must also retrieve the unlocked treasure chest, which may be easy to overlook during the battle.

    \item \textbf{NPC Interaction.} Agents locate designated NPCs within crowds and engage in dialogues to complete quests or access services such as shops and crafting materials. This task combines OCR capacity for target detection and precise GUI manipulation.

    % NPC interactions are essential for quest completion as well as for accessing shops, crafting materials, and other services. Agents must locate designated NPCs within crowds and successfully engage in dialogues. This task requires not only effective navigation but also OCR capabilities for target detection and precise GUI manipulation.
    
    \item \textbf{Puzzle.} Agents solve diverse challenges, such as activating elemental mechanisms, stepping on pressure plates, lighting torches in sequence, completing time-limited trials, and uncovering hidden pathways. Unlike collection or combat tasks, puzzles require careful observation, logical reasoning, and strategic use of elemental interactions, while requiring precise spatial awareness and fine-grained control.
\end{itemize}

We categorize tasks into three difficulty levels: simple, hard, and unseen. \textbf{Simple tasks} are short-horizon challenges, typically completed within 10 seconds (except for combat, which may last up to 2 minutes). These tasks evaluate fundamental skills, involving continuously visible and ground-level objects that require basic navigation and interaction. \textbf{Hard tasks} demand more advanced capacities, such as a nuanced understanding of gameplay mechanics, sophisticated 3D spatial reasoning, and precise low-level control. They often involve distractors, temporally hidden targets, or objectives requiring vertical or aerial navigation and combat against elite enemies. Both simple and hard tasks are situated in Mondstadt, the first nation in Genshin Impact, allowing us to evaluate in-distribution performance. \textbf{Unseen tasks} are designed to measure out-of-distribution capacities. These tasks are set in new environments, like Liyue and Inazuma, and feature novel objects, items, NPCs, or enemies not encountered during training. We exclude puzzle tasks from this category, as they often depend on region-specific mechanics that the agent has not been exposed to.

\textbf{Baselines.} We benchmark Lumine against state-of-the-art VLMs: GPT-5~\cite{openai2025gpt5}, Gemini 2.5 Pro~\cite{google2025gemini2.5}, Grok4~\cite{xai2025grok4}, Doubao1.6-Vision~\cite{Volcengine2025} and Qwen3-VL-235B-A22B-Thinking~\cite{qwen2025qwen3vl}. All thinking models are with default thinking budgets. To adapt these general-purpose models for gameplay, we integrate them into the Cradle framework~\citep{tan2024cradle}. Interaction is enabled through a predefined set of skills in code format (e.g., \textit{turn(degree)}, \textit{move\_forward(duration)} and \textit{attack()}), that models can invoke via function calls. Models maintain at most five recent steps as historical information in the context as input, which we found empirically to yield the best performance. 
% More details are provided in Appendix~\ref{app:baselines}.

\textbf{Evaluation Setting.} In simple and hard tasks, all agents operate under identical early-game conditions, limited to the four starter characters, Traveler, Amber, Kaeya, and Lisa, at level 20 with default weapons in World Level 1. For unseen tasks in Liyue and Inazuma, the same four characters are used at level 40 and with appropriate equipment that aligns with the expected game progression. At the start of each task, the agent is placed at a predefined location and given textual instructions specifying the objective. We use human evaluation to assess task performance. Unless stated otherwise, each task is run five times. Since API responses can take as long as 30 seconds to return, we pause the game during inference following the setup in Cradle~\citep{tan2024cradle}.

\section{Experimental Results}
We design our experiments to systematically evaluate the effectiveness of our modeling approach and the contribution of each training stage, focusing on four key questions:
\begin{itemize}
    \item \textbf{Q1:} What does Lumine-Base learn during large-scale pre-training, and how do its core abilities emerge in this phase?
    \item \textbf{Q2:} How well does Lumine-Instruct follow natural language commands? Can it reliably complete both in-domain and unseen tasks given textual instructions, and does incorporating history introduce any benefits?
    \item \textbf{Q3:} After the full three-stage training, can Lumine-Thinking perform complex, long-horizon tasks and generalize to out-of-domain or even entirely new games?
\end{itemize}
We first evaluate Lumine under the non-history setting and then extend the evaluation to the history setting. By default, Lumine-Base, Lumine-Instruct, and Lumine-Thinking refer to models trained under the history setting, whereas their non-history counterparts are denoted as Lumine-Base-NonHis, Lumine-Instruct-NonHis, and Lumine-Thinking-NonHis.

\subsection{Scaling Results}
\begin{figure}[h]
    \centering
    % 第一张子图
    \begin{subfigure}[b]{0.35\linewidth}
        \centering
        \includegraphics[width=\linewidth]{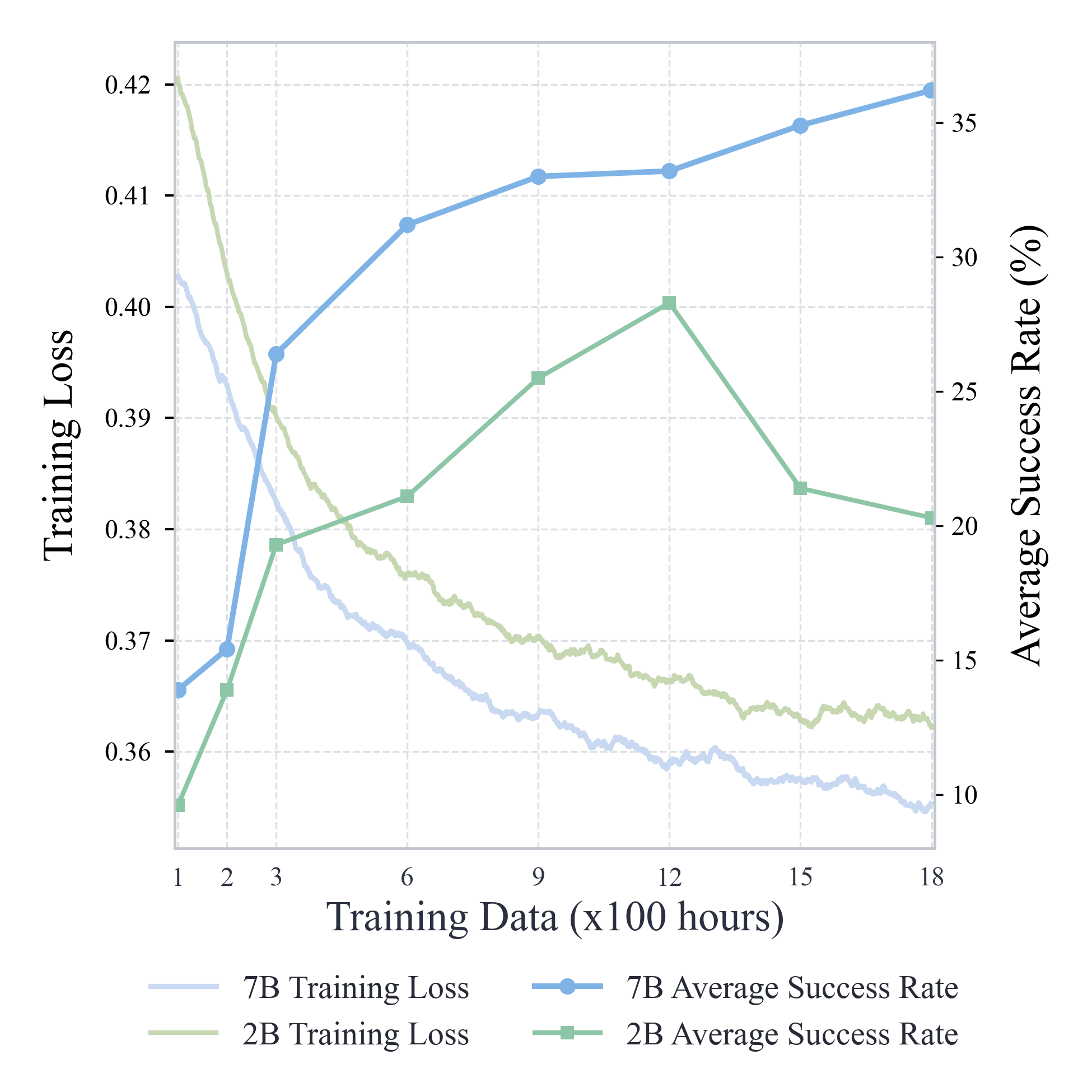}
        \caption{Training dynamics of base models.}
        \label{fig:scaling_loss_success_rate}
    \end{subfigure}
    % 第二张子图
    \begin{subfigure}[b]{0.62\linewidth}
        \centering
        \includegraphics[width=\linewidth]{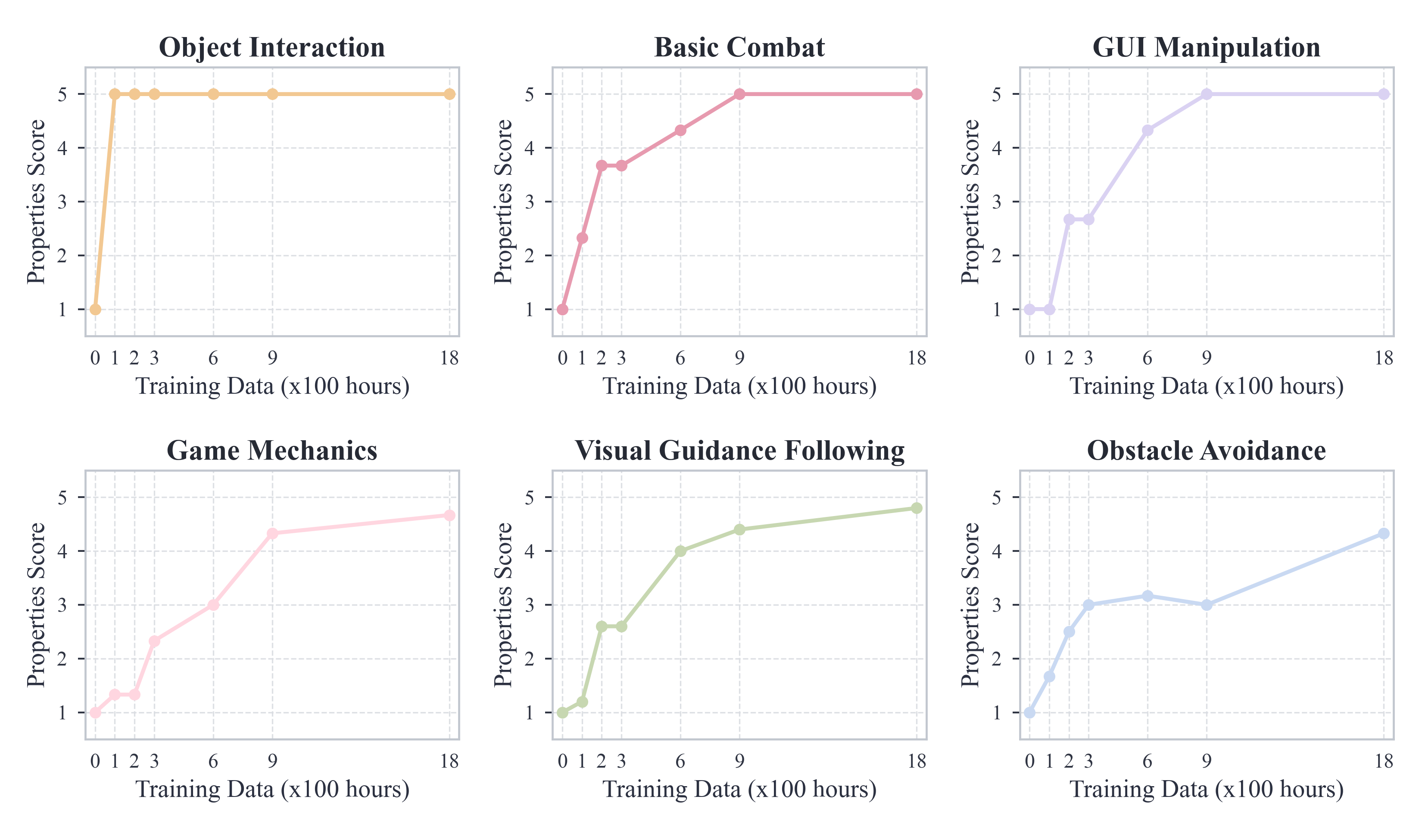}
        \caption{Capability emergence of 7B base model. }
        \label{fig:scaling_performance}
    \end{subfigure}
    \caption{Scaling analysis of Lumine-Base trained under non-history setting during pre-training stage. The left figure presents loss curves and corresponding success rates on the full set of benchmark tasks as the amount of Genshin training data increases in the first epoch of pre-training for both 2B and 7B base models, while the right figure illustrates the progressive improvement of key abilities for 7B base model, highlighting distinct scaling behaviors across different capability types.}
    \label{fig:capability_emergence}
\end{figure}

In this section, we investigate the scaling behavior of Lumine without history during the first epoch of pre-training, focusing on two model sizes: 2B and 7B. As shown in Figure~\ref{fig:scaling_loss_success_rate}, before the 1200 hours of data training, both models exhibit steadily decreasing training loss alongside consistent improvements in benchmark performance. Notably, the 7B model consistently achieves lower loss than the 2B model. However, beyond 1200 hours, a divergence emerges: while the 2B model’s loss continues to decline, its benchmark performance begins to degrade, revealing the limited volume of smaller models. In contrast, the 7B model maintains stable improvements across both loss and benchmark metrics. Based on these observations, we adopt the 7B model as our primary base. These findings highlight the effectiveness of our pre-training strategy and provide strong evidence that further scaling will yield additional performance gains.

\textbf{Atomic Ability Evaluation.} 
To better understand the sources of the improvements of the 7B model during the pre-training,  we conduct a study on the model's behaviors from an atomic perspective. The primary objective is to verify that the agent has learned context-appropriate reactions to various in-game scenarios. To this end, we designed a test suite of controlled scenarios to rigorously measure a set of core capabilities essential for effective gameplay. These atomic capacities include:

\begin{itemize}
    \item \textbf{Object Interaction.} Assesses the agent's capacity to interact with nearby objects within reach, such as picking flowers, opening treasure chests, or engaging with NPCs.
    \item \textbf{Basic Combat.} Measures the agent’s proficiency in combat scenarios, including executing basic attacks, switching characters to perform combos and identifying as well as engaging elevated enemies with ranged characters.
    \item \textbf{GUI Manipulation.} Tests the agent's ability to navigate and control graphical user interfaces meaningfully, especially for mouse movement, e.g., clicking options to continue dialogue, reviving characters, and navigating menu systems.
    \item \textbf{Game Mechanism.} Uses puzzles to examine agents’ understanding of core game mechanics, such as switching to the appropriate elemental character to activate an elemental monument, or using a Pyro character to light a torch.
    \item \textbf{Visual Guidance Following.} Evaluates the agent’s ability to follow visual directional cues, such as golden quest markers shown in the overworld. Specifically, it tests whether the agent can rotate the camera to center the marker in its view and move forward along the indicated direction.
    \item \textbf{Obstacle Avoidance.} Evaluates the agent's spatial awareness and locomotion skills across four aspects: keep following the path road without deviation, avoiding obstacles such as trees and walls, stopping safely at the edge of cliffs or rivers, and deploying a glider promptly during falls. 
\end{itemize}

The evaluation was conducted through human analysis of the models' gameplay videos, with performance rated on a 5-point scale:

\begin{itemize}
    \item \textbf{Score 1}: The agent exhibits no task-relevant behavior or intent, resulting in complete failure.
    \item \textbf{Score 2}: The agent shows initial intent but fails to make meaningful progress or complete the task.
    \item \textbf{Score 3}: The agent displays relevant behaviors but with significant hesitation and frequent errors, leading to a low success rate.
    \item \textbf{Score 4}: The agent demonstrates generally appropriate behavior with only occasional mistakes, achieving a high success rate.
    \item \textbf{Score 5}: The agent fully masters the capability, performing immediate and accurate actions to complete the task with proficiency.
\end{itemize}

Figure~\ref{fig:scaling_performance} shows that all core capabilities strengthen as scale increases, but at notably different rates, reflecting the relative difficulty of mastering each skill. 

\begin{itemize}
    \item \textbf{Immediate Interaction ($< 100$ hours).} Simple behaviors such as basic object interaction emerge quickly. At this stage, agents can consistently gather nearby resources, whether local specialties found in the wild or items dropped by enemies, and interact with NPCs in their vicinity by pressing key \texttt{F}.
    
    \item \textbf{Extended Interaction ($\sim 1{,}000$ hours).} More complex behaviors, such as basic combat and UI manipulation, become smooth and reliable. Agents can switch characters to chain skill combos, execute ranged attacks against elevated or distant targets. They also manage to handle common basic GUI events, from moving the cursor to select dialogue options, to clicking the close button in the top-right corner, or choosing Confirm or Cancel buttons when responding to interface prompts, such as during a character revival.

    \item \textbf{Game Mechanics ($>1{,}800$ hours).} Game mechanics pose remarkable challenges. Agents can recognize game-specific puzzle elements and exhibit reasonable reactions, but their sparse occurrence in raw gameplay, coupled with the wide variety present in Genshin Impact, makes these mechanics significantly harder for agents to master and complete the full task.

    \item \textbf{Navigation ($>1{,}800$ hours).} Navigation, the most essential component for open-world exploration and quest progression, requires significantly more data. By this point, agents exhibit robust road sense: they tend to follow the in-game roads to proceed, avoid obstacles such as trees and walls, halt at cliff edges, use the wind glider to prevent fall damage, and follow quest markers efficiently. This reliable navigation ability establishes a solid foundation for longer-horizon tasks. It sometimes exhibits hesitation, inefficient stamina management and mistimed actions, leaving room for further improvement.
\end{itemize}

\subsection{Instruction Following Performance}
After large-scale pretraining, Lumine-Base-NonHis has already developed the fundamental capabilities needed to accomplish a wide range of tasks. We next investigate whether Lumine-Instruct-NonHis can effectively follow human instructions and how its performance improves through alignment with language supervision. Our evaluation begins with the model under the non-history setting, followed by an extension to the history setting.

\subsubsection{Non-History Performance}
\begin{figure}[h]
    \centering
    \includegraphics[width=\linewidth]{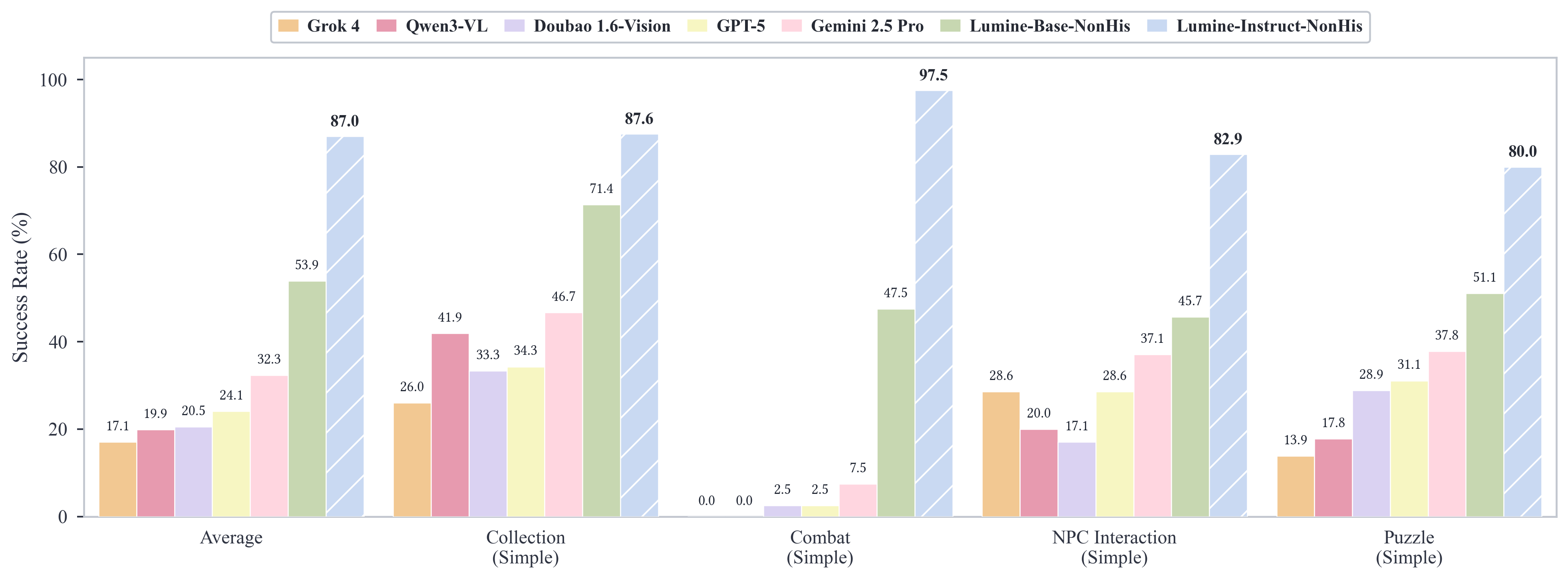}
    \caption{Average success rate of agents on the benchmark simple tasks by categories. Lumine-Instruct-NonHis achieves over 80\% success across all four categories, significantly outperforming its base model and all baseline methods.}
    \label{fig:baseline}
\end{figure}

\textbf{Baselines Comparasion.}
As shown in Figure~\ref{fig:baseline}, we evaluate both Lumine base and instruct models trained under non-history setting, against multiple baseline models on the benchmark’s simple tasks. While Lumine-Base-NonHis already significantly outperforms all baselines, the instruct model further achieves a 61\% performance gain, reaching over 80\% success across all four categories. The largest improvement occurs in \textit{combat} tasks, where the instruct model doubles the success rate of the base model. The instruct model demonstrates more consistent goal-conditioned behavior. For example, after battles, it actively adjusts the camera to locate and open the unlocked treasure chests, a step often neglected by the base model, which instead tends to just move forward. For \textit{NPC Interaction} and \textit{Puzzle} tasks, Lumine-Instruct-NonHis achieves more than a 50\% gain in performance, showing more consistent engagement with targets. Even in Collection tasks, where the base model already performs strongly, the instruct model still provides a 23\% improvement. These findings demonstrate Lumine-Instruct-NonHis’s robust ability to follow language instructions, maintain behavioral consistency, and reliably complete diverse common short-horizon tasks.

\begin{figure}[t]
    \centering
    \begin{minipage}[b]{0.68\textwidth}
        \centering
        \includegraphics[width=\linewidth]{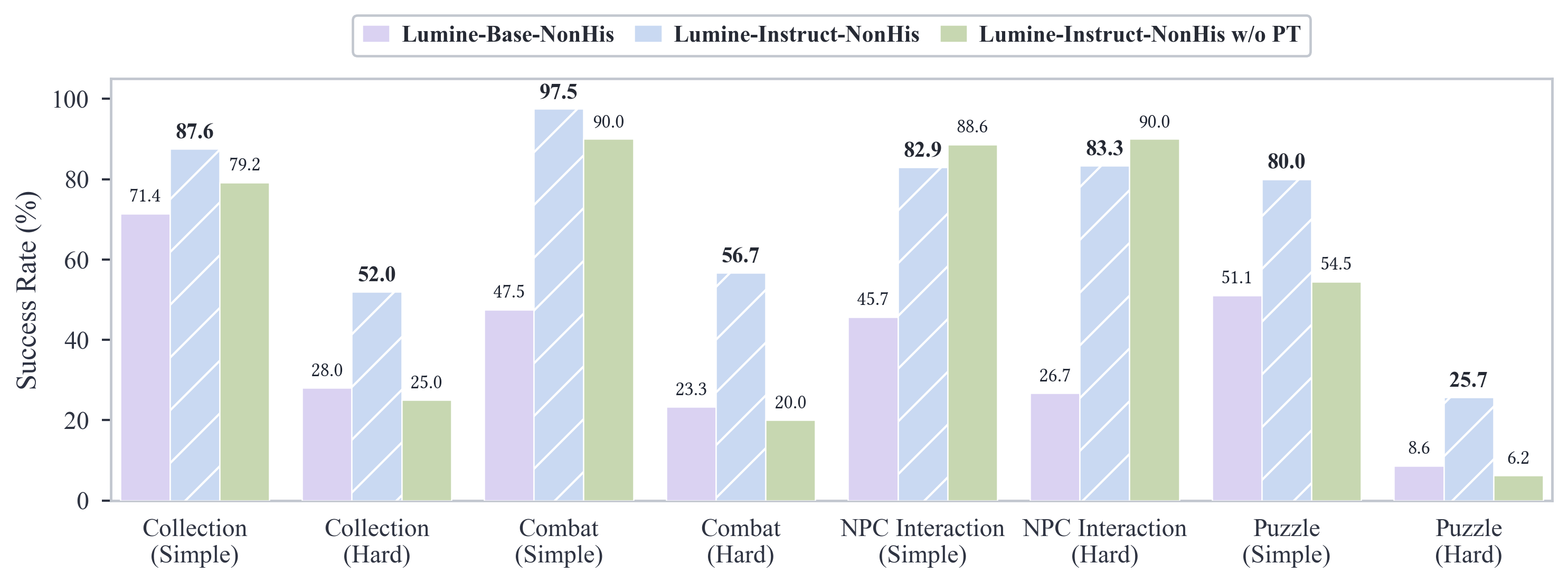}
        \caption{Performance of Lumine-Base, Lumine-Instruct and Lumine-Instruct without pre-training on simple and hard tasks under non-history setting.}
        \label{fig:performance_hard}
    \end{minipage}
    \hfill
    \begin{minipage}[b]{0.30\textwidth}
        \centering
        \includegraphics[width=\linewidth]{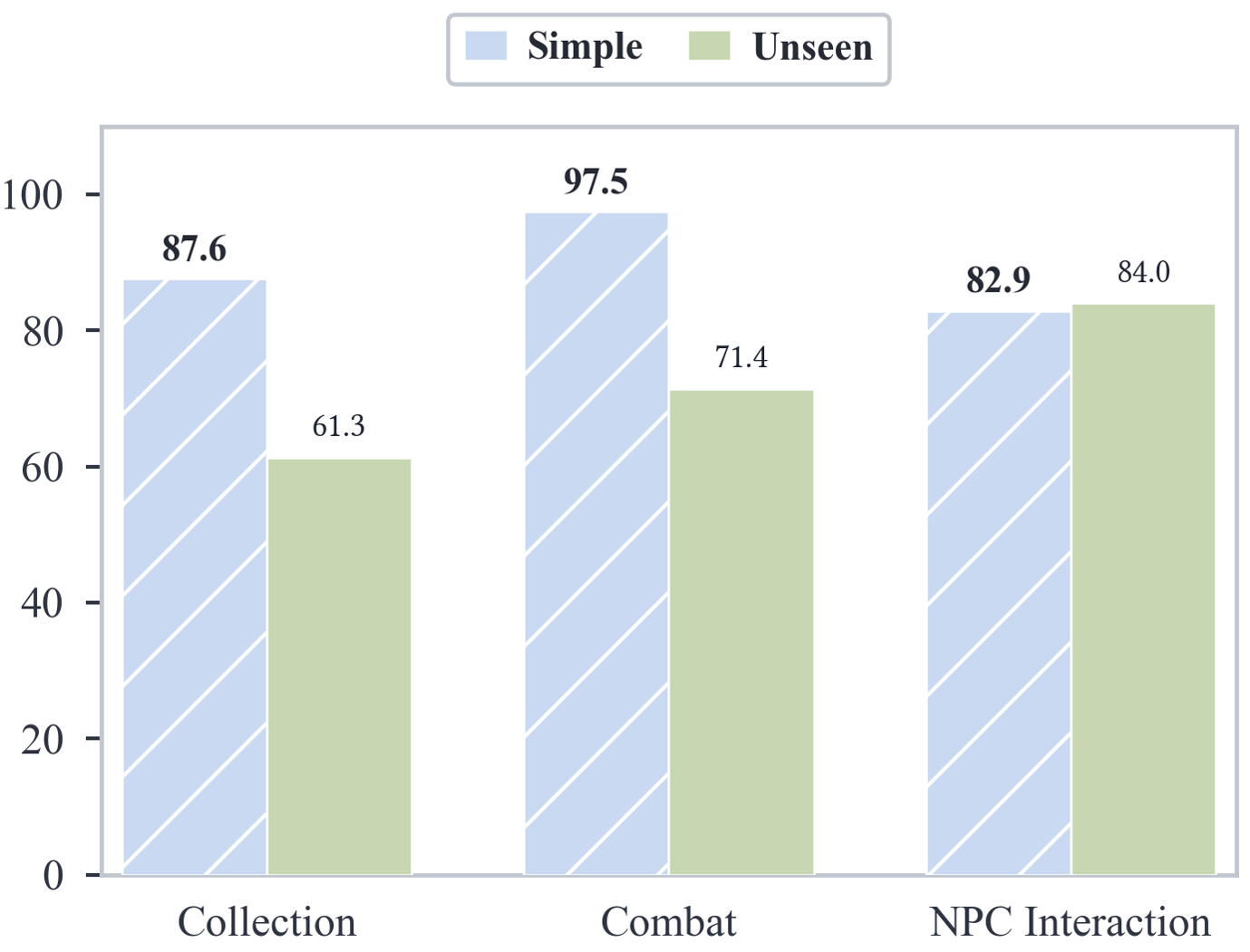}
        \caption{Comparison of Lumine-Instruct-NonHis performance in simple and unseen tasks.}
        \label{fig:performance_ood}
    \end{minipage}
\end{figure}

\textbf{Performance on Hard Tasks}. 
We further evaluate Lumine-Instruct-NonHis on hard tasks shown in Figure~\ref{fig:performance_hard}. The model exhibits consistent performance in \textit{Interaction} tasks, highlighting its robustness: even when surrounded by multiple non-target NPCs acting as distractors, it can still successfully engage with the designated NPC. More pronounced limitations appear in \textit{Combat} and \textit{Collect}. Although the model manages to defeat tough elite enemies such as \textit{Ruin Guard}, \textit{Debt Collector}, and \textit{Fatui Electro Cicin Mage}, it sometimes fails due to an insufficient understanding of game mechanics and poor dodging responses, often resulting in timeouts or even full-party eliminations. The model exhibits a zero success rate against the Eye of the Storm on elevated terrain, a flying enemy that lands only briefly. While the agent correctly switches to the ranged character Amber, its limited efficiency and precision in moving-target aiming, coupled with delayed evasive actions, often lead to significant damage or character death. The elevated terrain further increases difficulty by punishing movement errors with the risk of falling. These challenges demand precise timing, spatial awareness, and adaptive strategy, underscoring the need for further improvement in combat capabilities. The drop of \textit{Collection} tasks also expose deficiencies in spatial reasoning and fine-grained control, particularly when retrieving items that are not located on flat ground. Performance on \textit{Puzzle} tasks drops the most, reflecting their comprehensive requirement for the abilities above and emphasizing the importance of a deeper understanding of in-game mechanisms.

Figure~\ref{fig:performance_hard} also shows the impact of pre-training through an ablation study. As expected, models trained exclusively on instruction-following data exhibit overall lower performance, particularly on hard tasks. Interestingly, however, the model achieves slightly better performance in \textit{NPC Interaction} tasks. This suggests that the pre-training data introduces a notable bias, where many trajectories involve players merely passing by NPCs without engaging in interaction. The instruction-following data helps to mitigate this bias by reinforcing goal-directed behaviors and promoting more deliberate interactions with the environment.

\textbf{Performance on Unseen Tasks}. We further evaluate Lumine on unseen tasks. As shown in Figure~\ref{fig:performance_ood}, the model exhibits strong generalization capabilities in \textit{NPC Interaction}, achieving performance comparable to in-domain settings. This indicates that such interaction abilities can be effectively transferred to new scenarios and entities. In contrast, we observe a moderate but acceptable performance drop in \textit{Collection} tasks. Unlike \textit{NPC Interaction}, where unseen NPCs can still be identified by the names displayed above them, new collectibles must be recognized solely by their visual appearance, which the model has never encountered during training. Consequently, the agent must approach potential targets closely until their names are revealed to confirm correctness, significantly increasing task difficulty. Performance in \textit{Combat} tasks also declines, as Lumine-Instruct-NonHis is unfamiliar with previously unseen enemy attack patterns and combat mechanics, preventing it from reasoning about and reacting appropriately to new behaviors. Overall, while some degradation is observed, the results remain within an acceptable range, demonstrating Lumine's strong generalization to new scenarios and objects.

\begin{figure}[t]
    \centering
    \includegraphics[width=\textwidth]{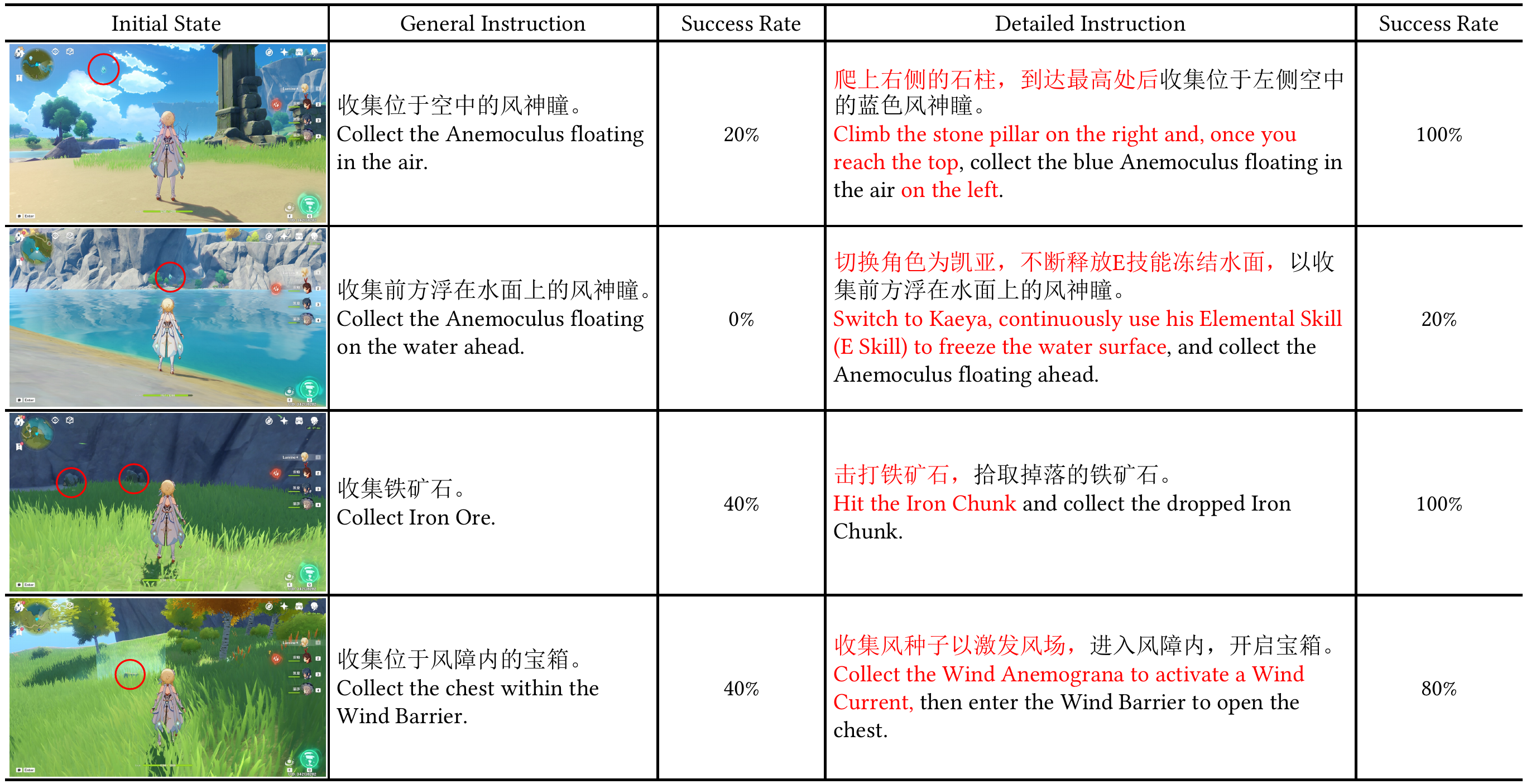}
    \caption{Case study of the in-context abilities of Lumine-Instruct-NonHis. When provided with additional contextual details that are relevant to the instruction, Lumine demonstrates improved performance and is able to complete previously low-performing tasks more effectively. The instructions given to Lumine were originally in Chinese; their English translations are provided here for reference.   }
    \label{fig:detailed_instruction}
\end{figure}

\textbf{Complex Instruction Following}. We observe that, beyond additional training, Lumine's performance can further be improved through in-context learning. As shown in Figure~\ref{fig:detailed_instruction}, providing more detailed instructions, incorporating prior knowledge or task decomposition that breaks a complex objective into a sequence of manageable steps, enables the model to successfully complete tasks that previously had low or even zero success rates. This demonstrates the strong generalization ability introduced by language grounding and establishes a solid foundation for subsequent reasoning training.

\begin{figure}[h]
    \centering
    \includegraphics[width=\textwidth]{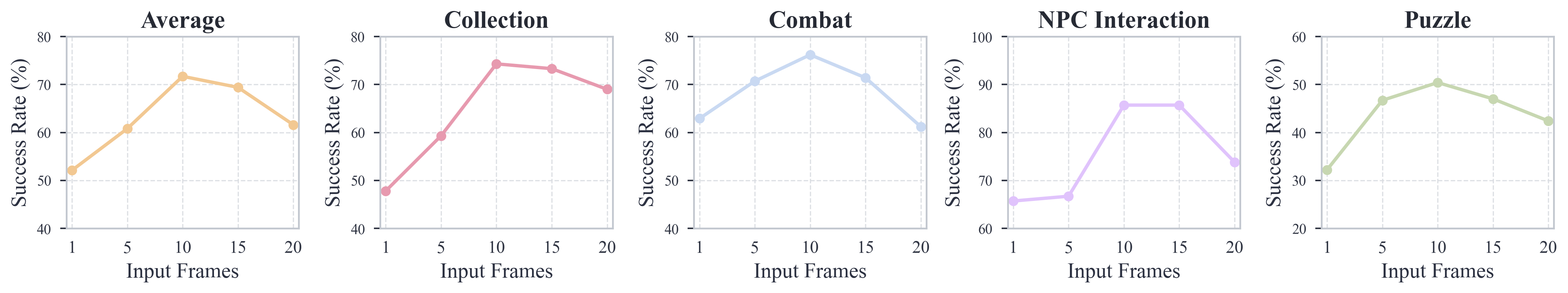}
    \caption{Performance of Lumine-Instruct preserving different lengths of frames in the context as historical information on the full set of benchmark.  }
    \label{fig:his_frames_ablation}
\end{figure}

\begin{figure}[h]
    \centering
    \includegraphics[width=\textwidth]{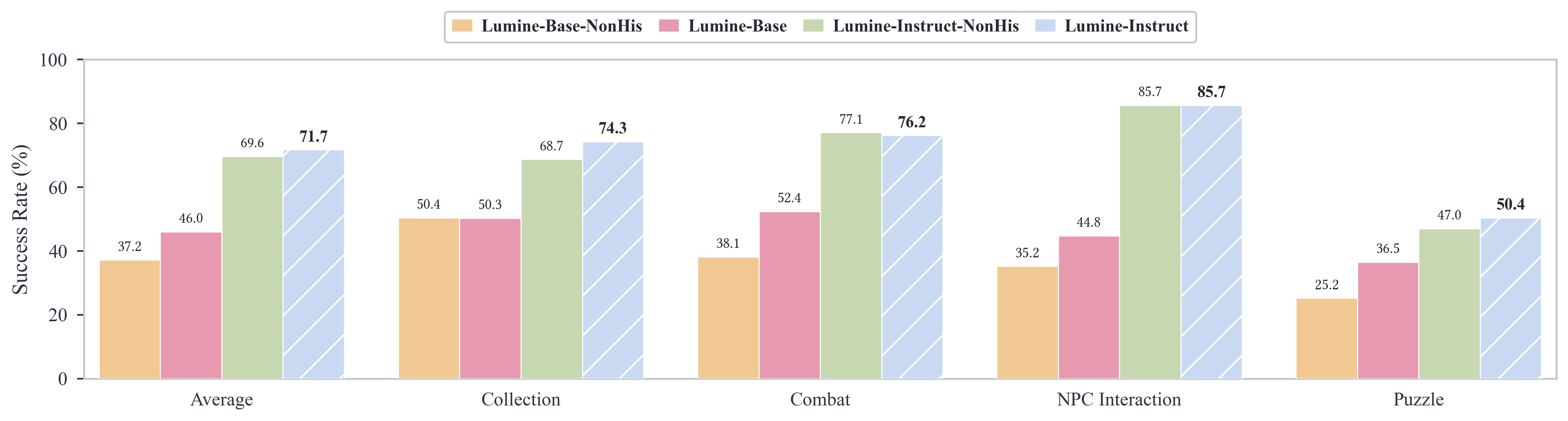}
    \caption{Comparison of Lumine trained under non-history and history settings on the full set of benchmark tasks.}
    \label{fig:his_vs_nonhis}
\end{figure}

\subsubsection{Benefits of History}
Since Lumine-Instruct-NonHis already demonstrates strong performance across a wide range of tasks, we next investigate whether incorporating historical information can further enhance model effectiveness.

As shown in Figure~\ref{fig:his_frames_ablation}, models that preserve multiple historical frames and actions in the context achieve substantially higher performance across all tasks compared to those limited to a single frame. Performance peaks when the model maintains 10 frames in context and starts to drop with more frames in the context. This decrease may be related to the data distribution: after the action filtering operation, which removes 95\% noop and trivial movements, the data segments typically less than 20 frames, making it harder for the model to learn long-term dependencies effectively at longer range. 

We then evaluate the best-performing configuration, the history model with 10 frames of context, on the full set of benchmark tasks, comparing it against the non-history baseline. Figure~\ref{fig:his_vs_nonhis} shows that history model exhibits a clear advantage in \textit{Collection} and \textit{Puzzle} tasks, achieving overall better performance.

\begin{figure}[h!]
    \centering
    \includegraphics[width=\linewidth]{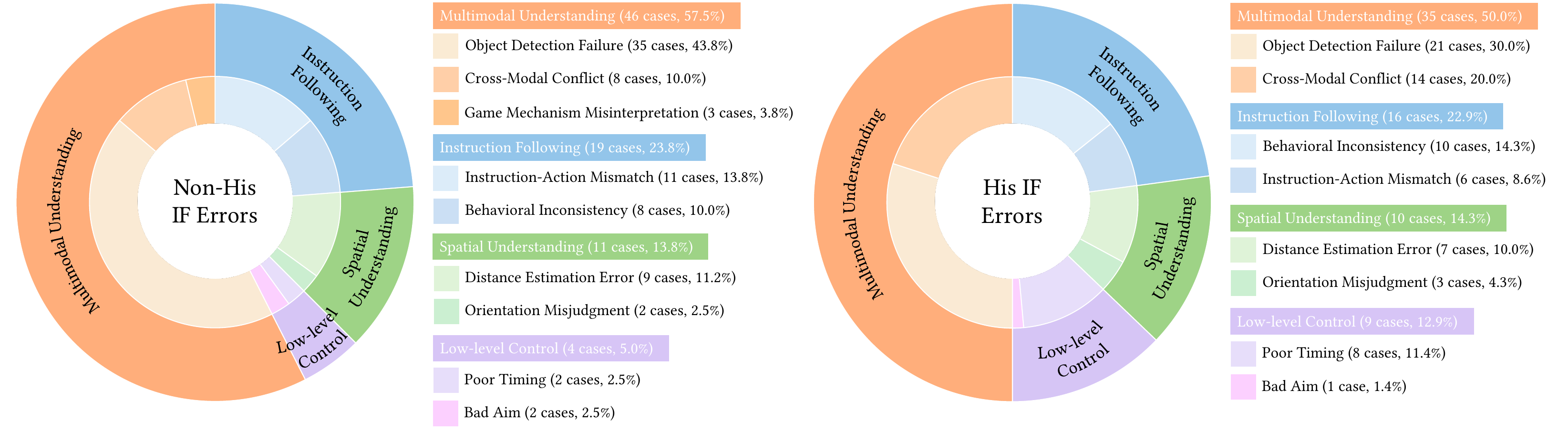}
    \caption{Error analysis of Lumine instruct models trained under non-history and history settings on the full set of benchmark tasks.
}
    \label{fig:IF_error_analysis}
\end{figure}

\textbf{Error Analysis}. As shown in Figure~\ref{fig:IF_error_analysis}, we conduct a comprehensive error analysis of the Lumine-Instruct models to better understand their behaviors across the full set of benchmark tasks. For tasks involving multiple mistakes, we report only the primary error that prevents further progress.

Our analysis reveals several key failure modes. The most significant is limited \textbf{Multimodal Understanding}, which accounts for nearly half of all errors. Agents frequently fail to detect target objects within the visual scene, particularly small, gatherable items that blend into the environment. Once the agent incorrectly infers that the target is not in view, it often rotates by a large angle to search, inadvertently causing already visible objects to move out of sight, making recovery difficult. History models show significantly fewer such errors, indicating that history improves dynamic tracking ability. Another common issue in multimodal understanding arises from conflicts between language and vision modalities. In these cases, the agent is overly influenced by vision, neglecting textual information embedded in images. For instance, when interacting with NPCs, agents may choose the wrong character despite the correct name being clearly visible. Interestingly, models without history make fewer of these modality-conflict errors, suggesting that historical context encourages a stronger focus on temporal consistency at the expense of single-frame understanding. The second major failure mode is \textbf{Instruction Following}. Both models occasionally demonstrate inconsistency when executing tasks. For example, after defeating two of three required \textit{Anemo Slimes}, an agent may abruptly abandon the objective and wander elsewhere. Agents also sometimes ignore fine-grained instructions, such as ``pick up the flower on the left'', instead turning away or moving forward in search of alternative targets. The third source of errors is \textbf{Spatial Understanding}, where agents misjudge their own position or fail to estimate distances accurately. This leads to poor navigation and timing errors, such as jumping too early or too late, thereby missing critical opportunities for progress. Finally, \textbf{Low-Level Control} contributes the smallest portion of failures. Both models exhibit unstable shooting performance and occasionally fail to interact with or collect objects, often due to delayed key presses. However, the history model makes noticeably more of these errors than the non-history model, which explains its lower performance on combat-related tasks. Overall, these findings highlight the diverse weaknesses across perception and control layers, leaving substantial room for future improvement.

\subsection{Reasoning Performance}
Lumine-Instruct models exhibit strong performance in language following and can successfully complete a wide range of short-horizon tasks across the world of Genshin Impact. Building on this foundation, we investigate whether Lumine-Thinking can address more challenging long-horizon tasks through its adaptive thinking capability. For evaluation, we primarily use the game’s main storyline as a testbed, an experience that typically takes human players hours to complete and encompasses most major events and gameplay mechanics. The storyline evolves rapidly, with both scenarios and objectives changing frequently, requiring the agent to react adaptively rather than relying on a fully pre-planned task decomposition. Such extremely long tasks serve as a comprehensive test of the agent’s capabilities and represent a critical benchmark for evaluating true autonomy.

\textbf{Setting}. Lumine trained under history setting is set with a maximum of 20 historical frames in the context. Thinking models will be forced to enter thinking mode when it does not generate a new reasoning for more than 100 steps, which is useful to help it recover from a stuck state. All the experiments are run in a real-time setting. Each model is provided with "Complete the main storyline mission" as an instruction in the prompt at the beginning of each task, which will be overridden by the following reasoning. Throughout the evaluation, agents are restricted to the four default system-provided characters, Traveler, Amber, Kaeya and Lisa, with their progression aligned to what a typical human player would reasonably have achieved at that stage.
       
\begin{figure}[h]
    \centering
    \includegraphics[width=\linewidth]{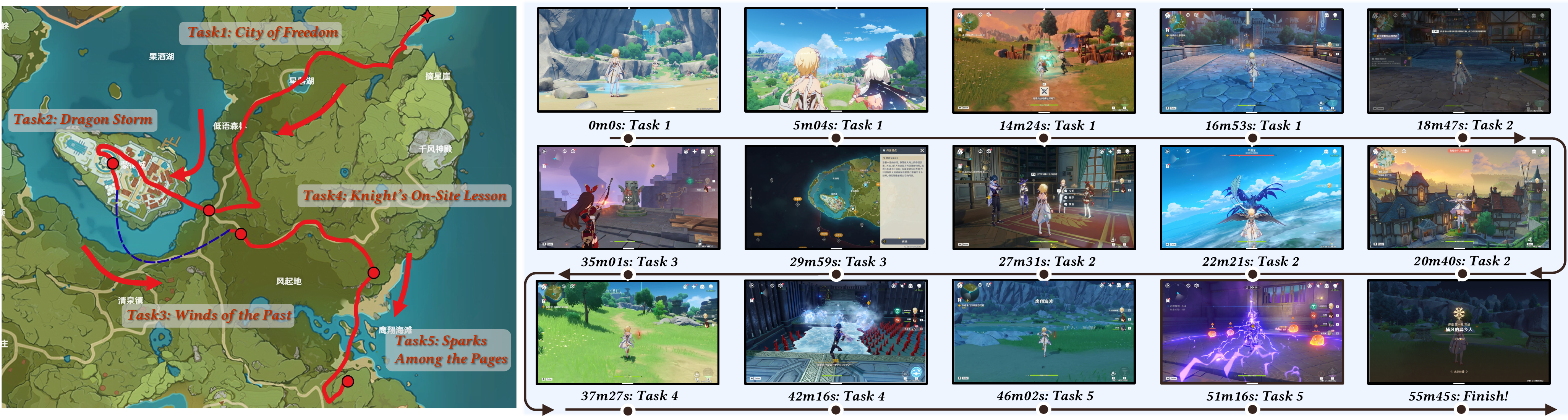}
    \caption{ Visualization of the in-domain evaluation mission, the main storyline of Mondstadt, \textit{Prologue: Act I - The Outlander Who Caught the Wind}, which is divided into five subtasks. The left figure illustrates the agent’s geographical trajectory during task completion. Red lines denote the character’s movement path, while blue lines indicate teleportation jumps between distant locations. The right figure presents the complete trajectory and corresponding timestamps for Lumine-Thinking, who completed the mission in \textbf{56} minutes, compared with fresh human players with an average of \textbf{78} minutes and expert human players with an average of \textbf{53} minutes.}
    \label{fig:reasoning_genshin_act1}
\end{figure}

\begin{table}[h]
    \centering
    \caption{Success rate of Lumine on the five subtasks of the in-domain mission. Each task is run three times.}
    \label{tab:navigationquest_reasoning}
    \begin{tabular}{l|cccccc}
        \toprule
        Model & Overall & Task 1 & Task 2 & Task 3 & Task 4 & Task 5 \\
        \midrule
        Lumine-Instruct-NonHis & 6.6\% & 0/3 & 0/3 & 0/3 & 1/3 & 0/3 \\
        Lumine-Thinking-NonHis & 53.4\% & 1/3 & 2/3 & 2/3 & 2/3 & 1/3 \\
        Lumine-Instruct & 66.8\% & 2/3 & \textbf{3/3} & 2/3 & 2/3 & 1/3 \\
        Lumine-Thinking & \textbf{93.4\%} & \textbf{3/3} & 2/3 & \textbf{3/3} & \textbf{3/3} & \textbf{3/3} \\
        \bottomrule
    \end{tabular}
\end{table}

\subsubsection{Performance in Genshin Impact}
\textbf{In-domain Performance}. We first evaluate Lumine-Thinking on missions that are covered in both the pretraining and reasoning datasets. As illustrated in Figure~\ref{fig:reasoning_genshin_act1}, we select the main storyline of Mondstadt \textit{Prologue:  Act I - The Outlander Who Caught the Wind}  as the primary testbed, where players are required to create a new account and progress entirely from scratch. This mission evaluates a broad spectrum of agent capabilities, including long-horizon navigation to quest locations, NPC interaction, instruction following, combat, boss fights, puzzle solving, domain exploration, and GUI operations such as character development, map teleportation, and dialog continuation. Completing this mission typically takes human players about one hour. To systematically evaluate agent performance, we divide the mission into five sequential subtasks:

\begin{itemize}
    \item \textbf{Task 1:} The Traveler follows Paimon across Starfell Lake and through the forest toward Mondstadt. Along the way, they meet Amber, join forces to defeat monsters, and ultimately reach the city.
    \item \textbf{Task 2:} The Traveler helps defend Mondstadt during Stormterror’s sudden attack, after which they are invited to the Knights of Favonius Headquarters to offer assistance against the escalating dragon threat.
    \item \textbf{Task 3:} The Traveler teams up with Amber to investigate and clear the \textit{Temple of the Falcon}.
    \item \textbf{Task 4:} The Traveler meets Kaeya and explores the \textit{Temple of the Wolf} under his guidance.
    \item \textbf{Task 5:} The Traveler joins Lisa in delving into the \textit{Temple of the Lion}.
\end{itemize}

As shown in Table~\ref{tab:navigationquest_reasoning}, thinking models achieve significantly better performance than instruct models, highlighting both the effectiveness and necessity of high-level reasoning. While instruct models can handle certain subtasks, they often fail to complete the full task chain. This failure typically arises from distractions encountered along the way, such as puzzles or enemies, which cause the model to lose track of the main objective. Once disoriented in the overworld, the model tends to wander aimlessly and struggles to recover the correct progression path. In contrast, thinking models are able to reflect and set appropriate goals for the current context and remain focused, avoiding such distractions. Lumine-Thinking with history demonstrates robust performance across all tasks, with only a single failure observed in Task 2. In that case, the model accidentally fell off a platform while traveling to the target quest location, Knights’ Headquarters and
crucially failed to generate intermediate reasoning steps to invoke the in-game quest guidance. Subsequently, it roamed the city in search of a way back, resulting in a timeout. Nevertheless, given additional time, the model would likely have recovered its route and completed the task successfully. Overall, Lumine-Thinking trained under history setting successfully completed the entire act in \textbf{56} minutes, outperforming fresh human players, whose average completion time was \textbf{78} minutes, and performing comparably to expert human players, who averaged \textbf{53} minutes. The expert group had substantial prior experience with the game and had played through the same story segment at least once within the preceding week.

On the other hand, models with history exhibit a clear advantage under real-time conditions, particularly in GUI manipulation and navigation. By contrast, non-history models are significantly affected by the 200ms delay introduced by asynchronous execution. Specifically, when a non-history model attempts to move the mouse to a designated position, the visual input it receives corresponds to the pre-execution state of the previous action. As a result, the model incorrectly believes that the cursor has not yet moved and generates another mouse movement action in the same direction. This leads to overshooting the target and makes precise clicking on GUI elements considerably more difficult. Since models with history have access to previous actions and trajectories in context, they are far less affected by latency and can perform operations much more smoothly and stably. The same phenomenon also occurs in navigation when turning directions.

\begin{figure}[htbp]
    \centering
    \includegraphics[width=\linewidth]{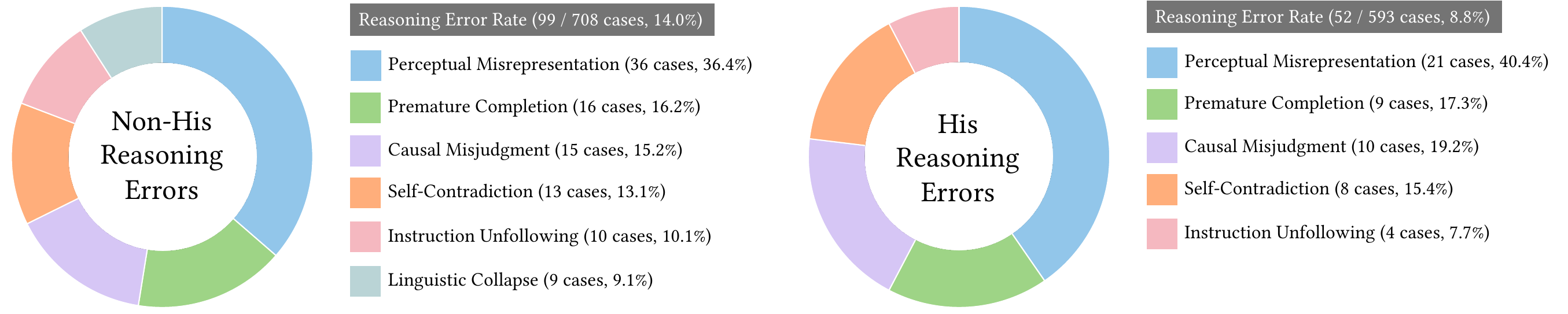}
    \caption{Error analysis of reasoning quality generated by Lumine during the completion of the entire in-domain mission. We investigate both non-history and history settings.}
    \label{fig:reasoning_error_analysis}
\end{figure}

\textbf{Error Analysis}. As shown in Figure~\ref{fig:reasoning_error_analysis}, we conducted an error analysis of the reasoning quality generated by Lumine under both non-history and history settings. During the completion of the entire mission, the history model generated 593 reasoning instances with an error rate of 8.8\%, which is significantly lower than that of the non-history model (708 instances, 14.0\% error rate) in both quantity and quality. These results demonstrate the advantage of the history model, which produces reasoning that is not only more efficient but also more accurate. It is worth noting that low-quality reasoning does not necessarily lead to task failure.

In terms of error composition, both settings show similar contributing factors and relative proportions. The most frequent error type in both settings is \textbf{Perceptual Misrepresentation}, which occurs when the model describes objects or scenarios that do not exist in the observation or misinterprets the character’s current status (e.g., health or stamina). This reflects the model’s limited situational understanding of the game environment. The second is \textbf{Premature Completion}, a type of hallucination in which the model incorrectly assumes that a proposed goal has already been achieved. Based on this false premise, it continues to generate reasoning that builds upon the nonexistent completion, thereby compounding the initial error. The third category, \textbf{Causal Misjudgment}, refers to instances where the model incorrectly believes that a given action or strategy will accomplish the intended outcome. These errors often arise from an incomplete understanding of the system’s causal mechanisms or underlying dynamics. \textbf{Self-Contradiction} describes reasoning that is internally inconsistent. For instance, the model sometimes claims, "I closed the task interface before… so I did not close the task interface," producing an inherently impossible statement that exposes the base model’s limited capacity for logical reasoning. We also observed several instances of \textbf{Instruction Unfollowing}, where the model’s generated actions failed to align with its own reasoning. Finally, we occasionally observed \textbf{Linguistic Collapse} in the non-history model where the model produced unreadable or incoherent outputs. This suggests that reasoning based solely on single-frame information imposes greater cognitive strain, making the model more prone to collapse. In contrast, the model trained under the history setting demonstrates greater robustness and stability.

\begin{figure}[t]
    \centering
    \includegraphics[width=\linewidth]{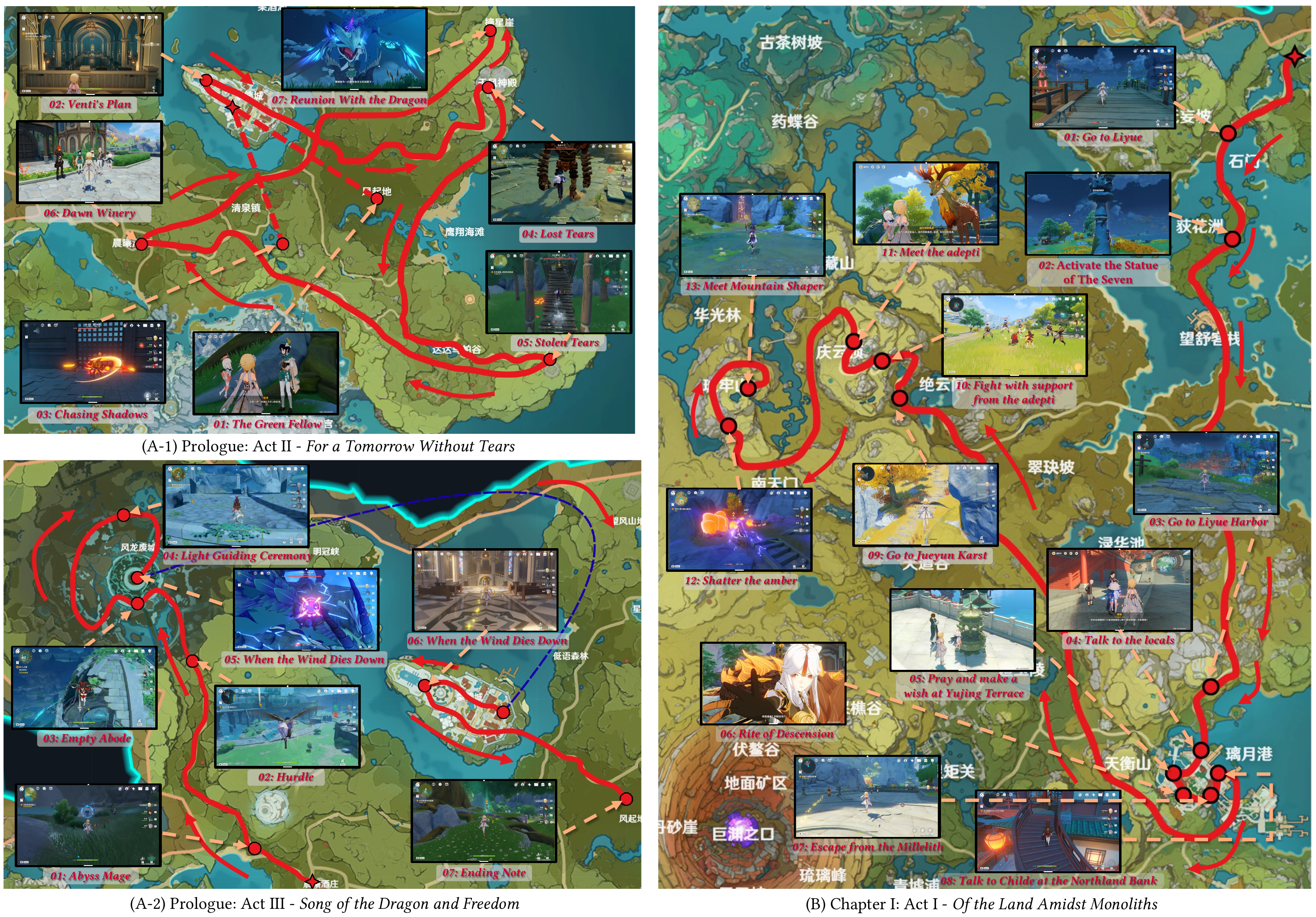}
    \caption{The two missions on the left, Acts II and III of Mondstadt’s main storyline, are included in the pre-training data but excluded from the reasoning data. Lumine successfully completed the two acts consecutively within \textbf{4.7} hours, compared to an average of \textbf{3.6} hours for expert human players.
    While Liyue's mission on the right is entirely new to Lumine, it still manages to reach Liyue Harbor and visit the Adeptus dwelling deep within the mountains. Due to space limitations, we are unable to present the full process of Lumine’s journey from Mondstadt’s Windrise to Liyue. The red dashed line indicates a round trip between two locations.}
    \label{fig:map23_show}
\end{figure}

\textbf{Generalization to Reasoning OOD Missions}. As Lumine demonstrates superior performance in trained tasks,  we then investigate whether the model’s reasoning abilities can generalize to previously unseen scenarios and the model's performance in extremely challenging tasks. We apply our best-performing model, Lumine-Thinking, to the remaining main storyline of Mondstadt \textit{Prologue:   Act II - For a Tomorrow Without Tears} and \textit{Prologue:  Act III - Song of the Dragon and Freedom}, where the Traveler and their companions overcome various difficulties along the way in search of a way to purify the corrupted dragon, Dvalin. They eventually confront Dvalin in his lair and succeed in purifying him, thereby lifting the threat to Mondstadt. This part of the gameplay is included in the pre-training data but excluded from the reasoning data. To ensure uninterrupted progression, we prepare a game account with an Adventure Rank of 18, allowing the agent to access and complete the quests without being blocked by rank requirements. 

\begin{figure}[t]
    \centering
    \begin{subfigure}[b]{0.24\textwidth}
        \includegraphics[width=\textwidth]{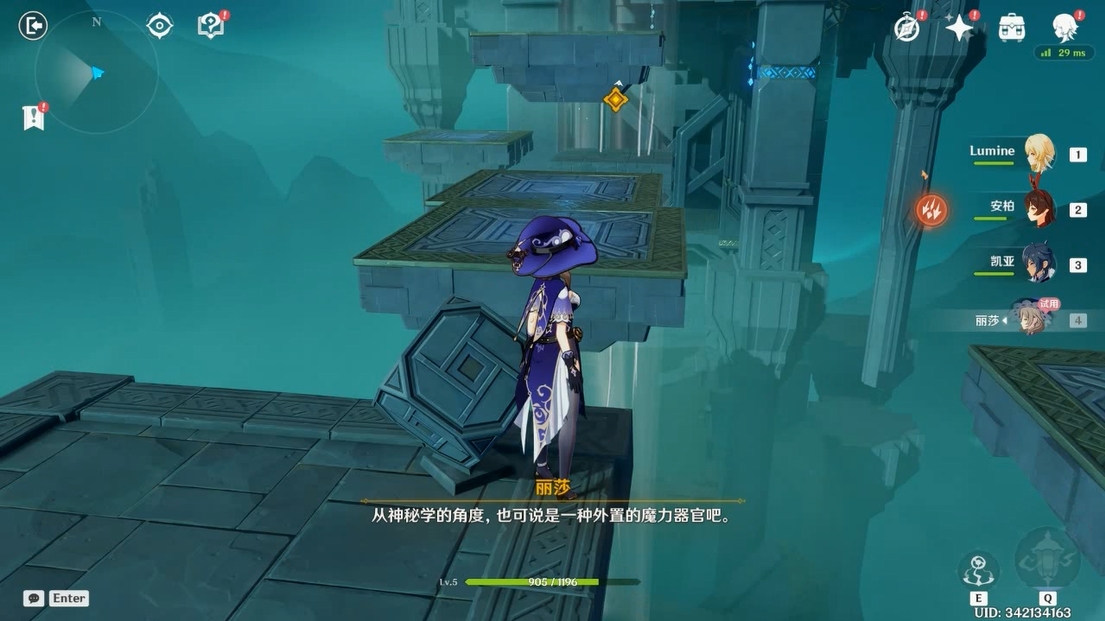}
        \caption{Moving platform (ID)}
        \label{fig:similiar_scenes_2_id}
    \end{subfigure}
    \hfill
    \begin{subfigure}[b]{0.24\textwidth}
        \includegraphics[width=\textwidth]{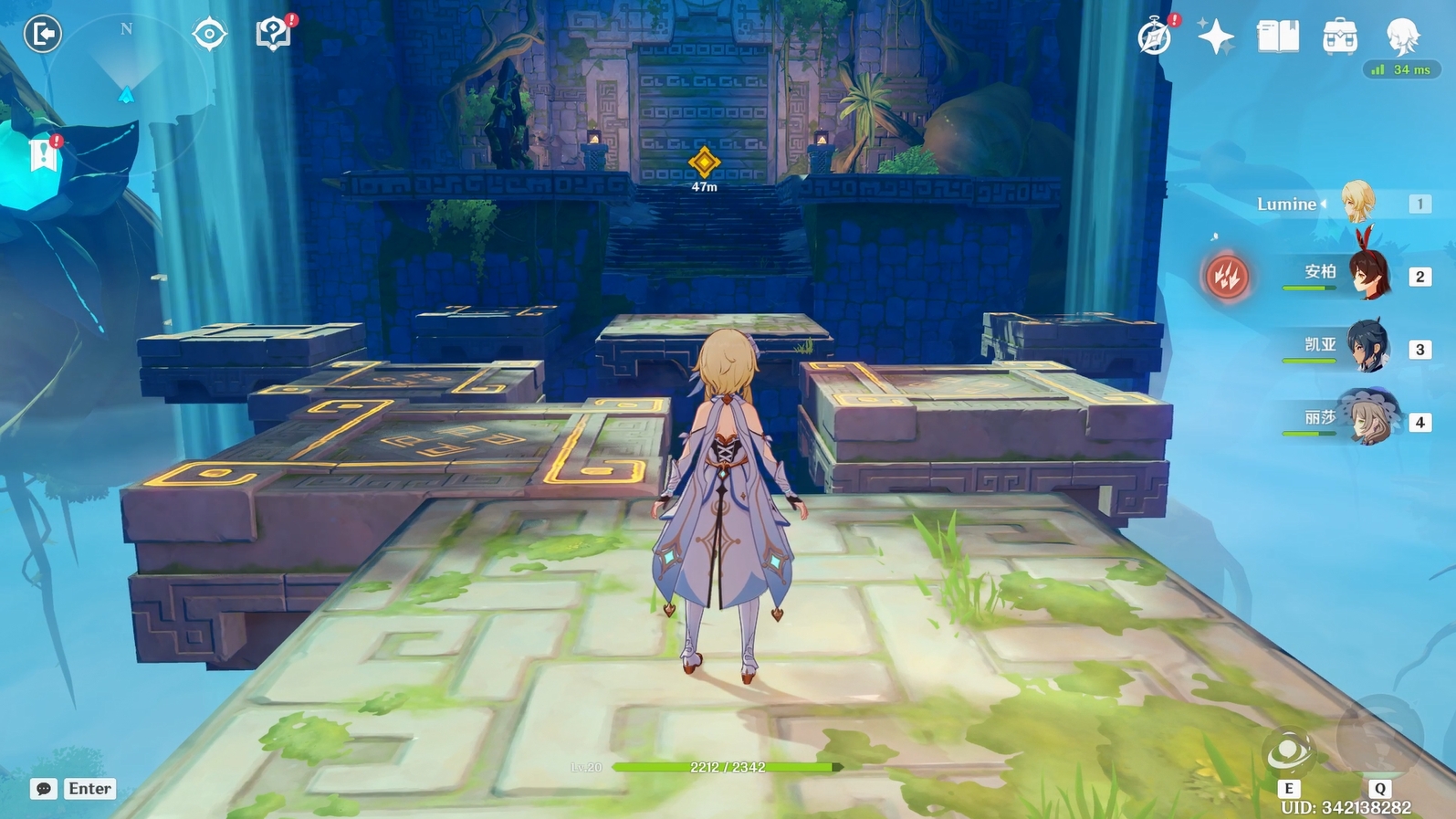}
        \caption{Moving platform (OOD)}
        \label{fig:similiar_scenes_2_ood}
    \end{subfigure}
    \hfill
    \begin{subfigure}[b]{0.24\textwidth}
        \includegraphics[width=\textwidth]{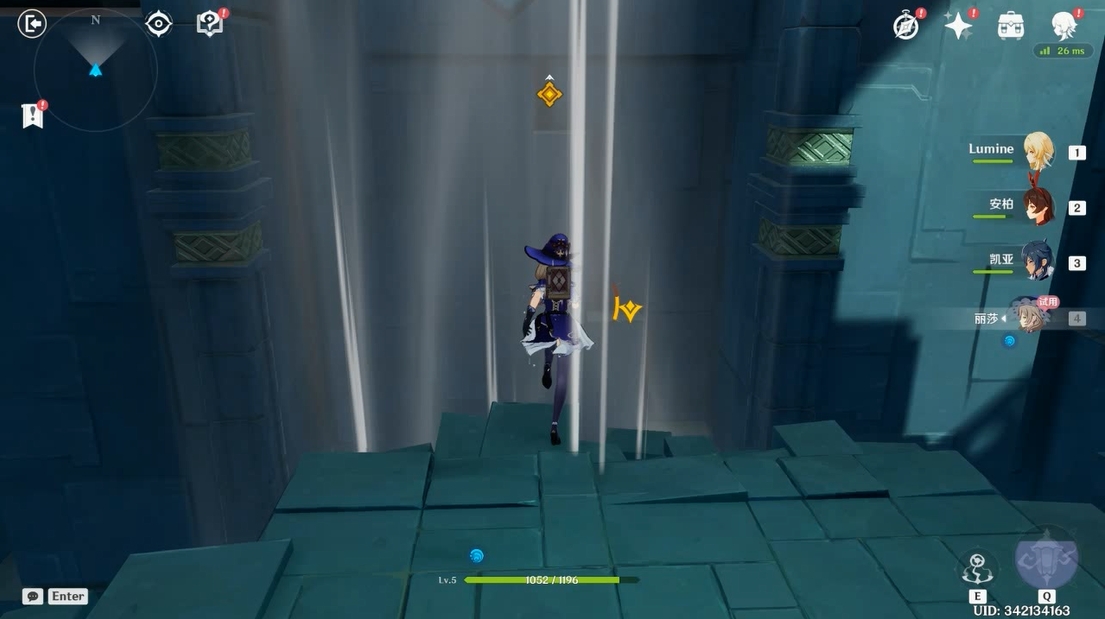}
        \caption{Wind current (ID)}
        \label{fig:similiar_scenes_1_id}
    \end{subfigure}
    \hfill
    \begin{subfigure}[b]{0.24\textwidth}
        \includegraphics[width=\textwidth]{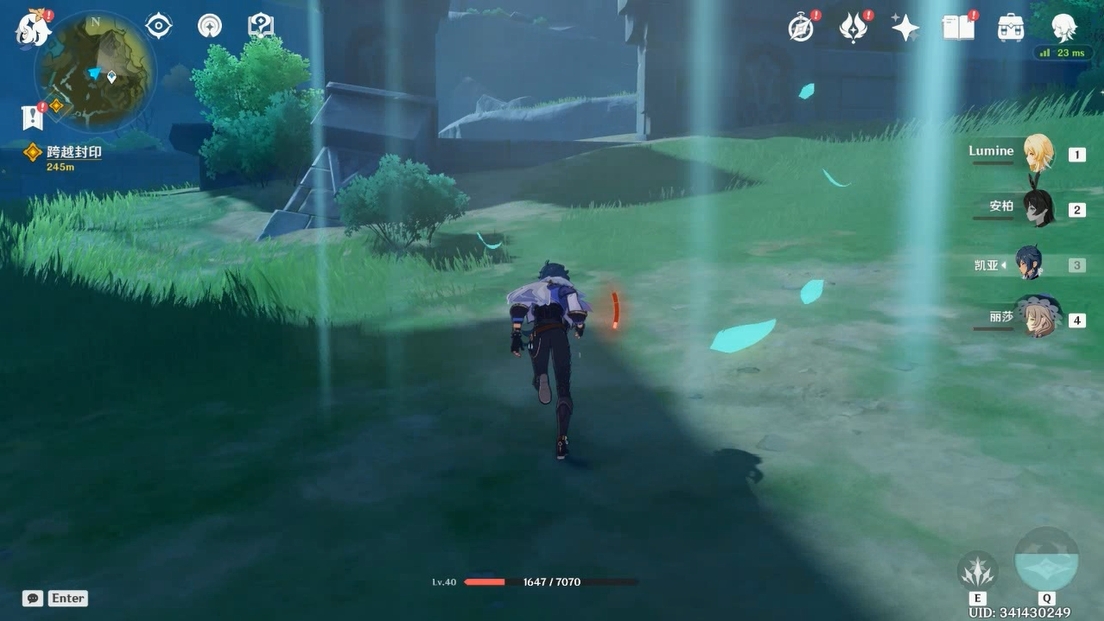}
        \caption{Wind current (OOD)}
        \label{fig:similiar_scenes_1_ood}
    \end{subfigure}
    \caption{Two examples of same mechanisms encountered by Lumine in both in-domain (ID) and out-of-domain(OOD) scenarios. Even across vastly different contexts and scenarios, Lumine-Thinking demonstrates strong generalization ability, successfully recognizing mechanisms in OOD settings and generating reasoning that effectively leverages them.}
    \label{fig:similiar_scenes}
\end{figure}

Lumine successfully completed the two acts consecutively within \textbf{4.7} hours, compared to an average of \textbf{3.6} hours for expert human players. During the course of these missions, Lumine also attempted to solve puzzles, open monster-guarded chests, and collect Oculus encountered along the way. It is encouraging to observe that Lumine consistently stayed on the right track, allowing the mission progress to advance smoothly, which indicates reasoning abilities manage to transfer to unseen scenarios, though with more hallucinations as expected. Reasoning tends to be precise in NPC interactions, GUI operations, common navigation, and combat, but becomes less reliable in unfamiliar mechanisms and puzzles, such as the wind wall in Stormterror’s Lair, where Lumine relies more on its low-level control. Interestingly, as shown in Figure~\ref{fig:similiar_scenes}, in one of the dungeons of Act II, there is a floating moving stone platform with a mechanism similar to that found in Lisa’s dungeon in Act I. Despite the completely different layout and background, Lumine successfully recognized the stone platform and explicitly reasoned that we should wait for the platform to move before proceeding. The wind current also tells the same story. This demonstrates the strong generalization capability of the VLM-based agent.

\begin{figure}[t]
    \centering
    % --- First row ---
    \begin{subfigure}[b]{\textwidth}
        \centering
        \begin{subfigure}[b]{0.24\textwidth}
            \includegraphics[width=\textwidth]{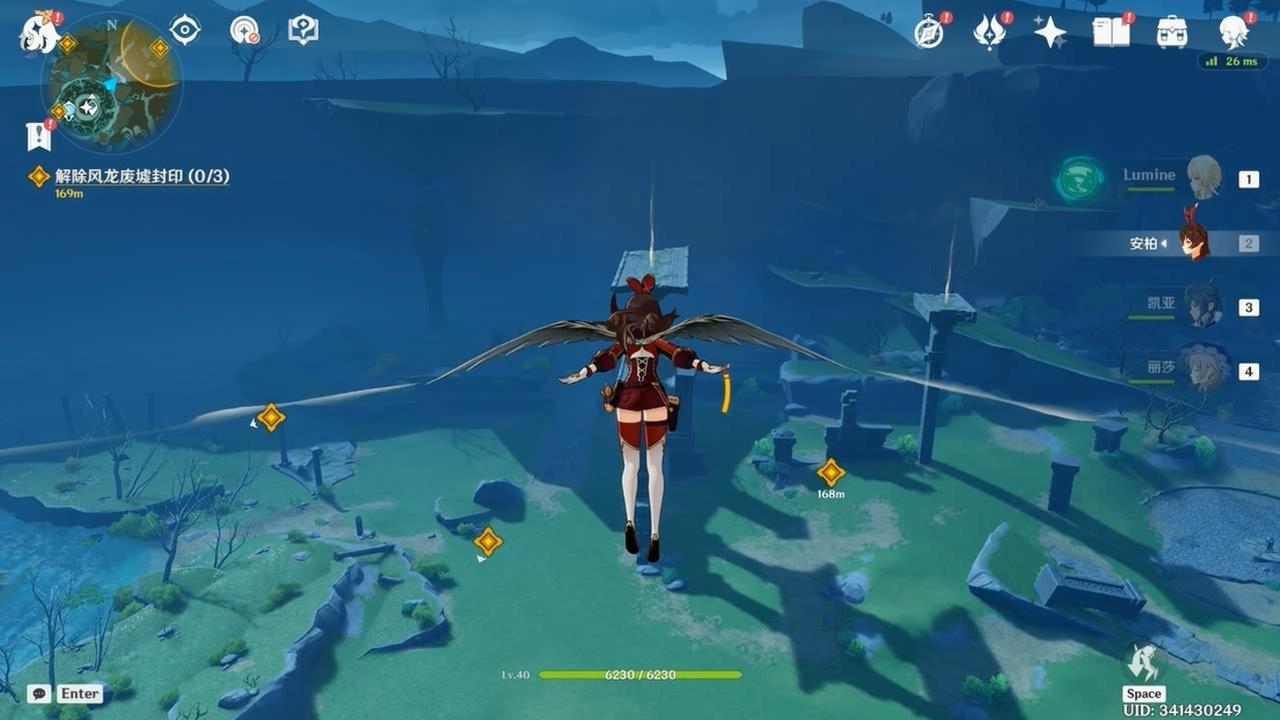}
        \end{subfigure}%
        \hfill
        \begin{subfigure}[b]{0.24\textwidth}
            \includegraphics[width=\textwidth]{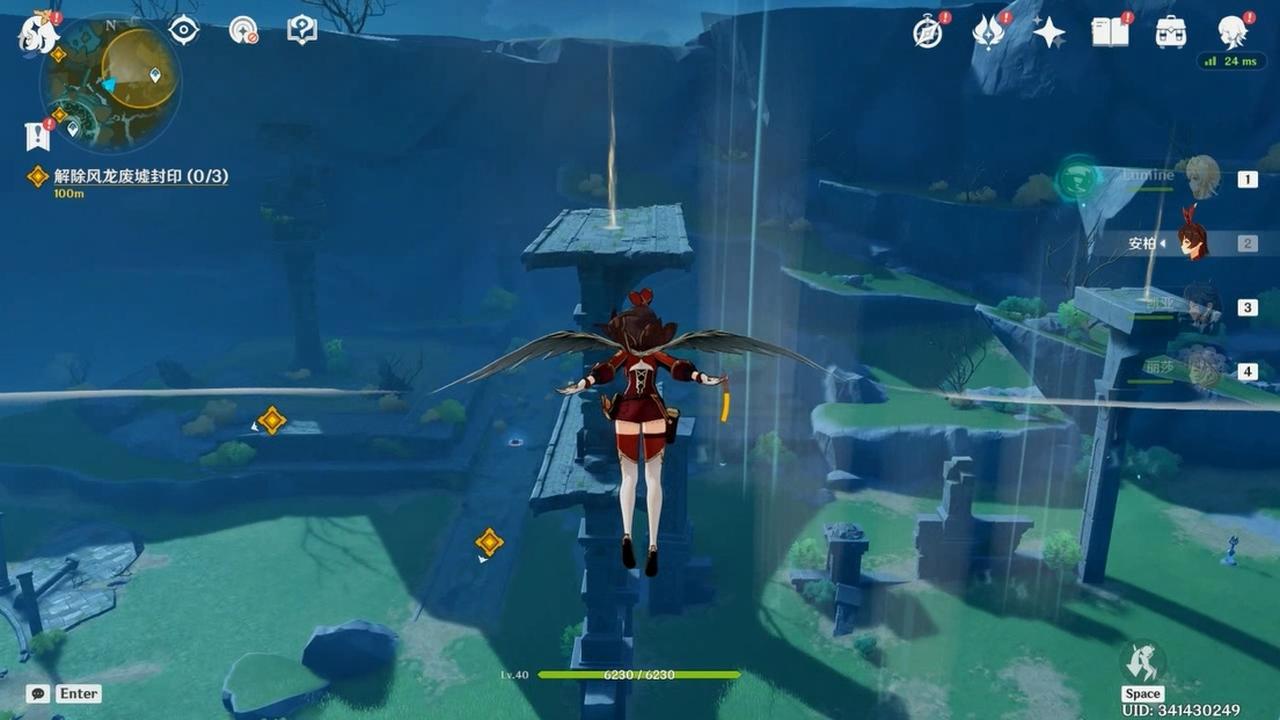}
        \end{subfigure}%
        \hfill
        \begin{subfigure}[b]{0.24\textwidth}
            \includegraphics[width=\textwidth]{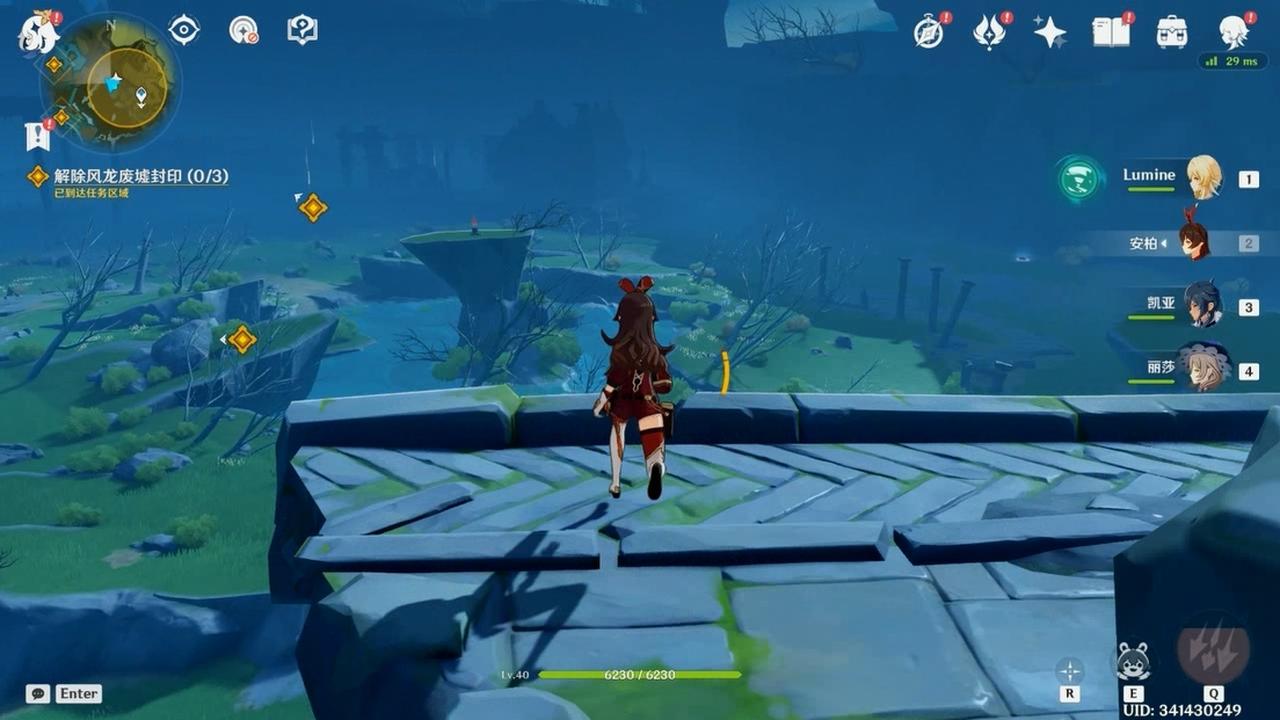}
        \end{subfigure}%
        \hfill
        \begin{subfigure}[b]{0.24\textwidth}
            \includegraphics[width=\textwidth]{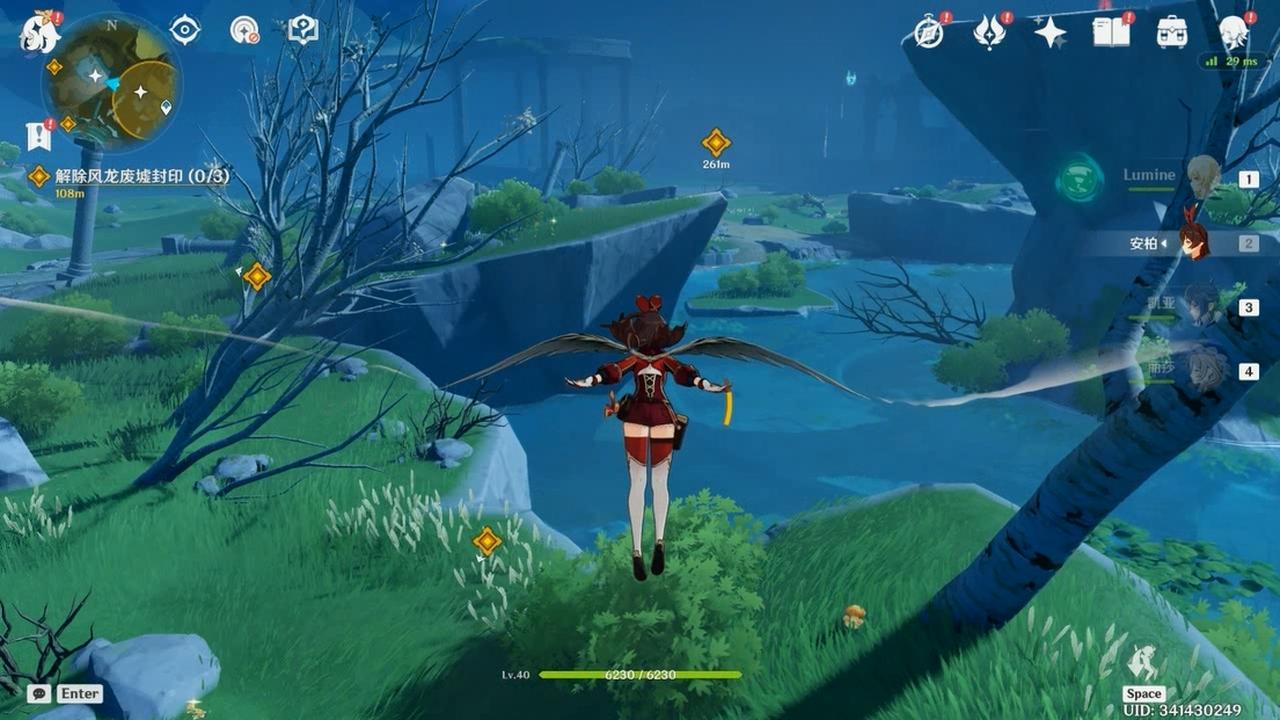}
        \end{subfigure}%
        \caption{Three golden quest markers appear simultaneously, each pointing to a different location. When the agent reaches one region of them, the corresponding marker disappears, leaving the other two active. The agent is often drawn toward these remaining markers and repeatedly moves back and forth among the three regions.}
    \end{subfigure}

    \vspace{1em} % vertical spacing between rows

    % --- Second row ---
    \begin{subfigure}[b]{\textwidth}
        \centering
        \begin{subfigure}[b]{0.24\textwidth}
            \includegraphics[width=\textwidth]{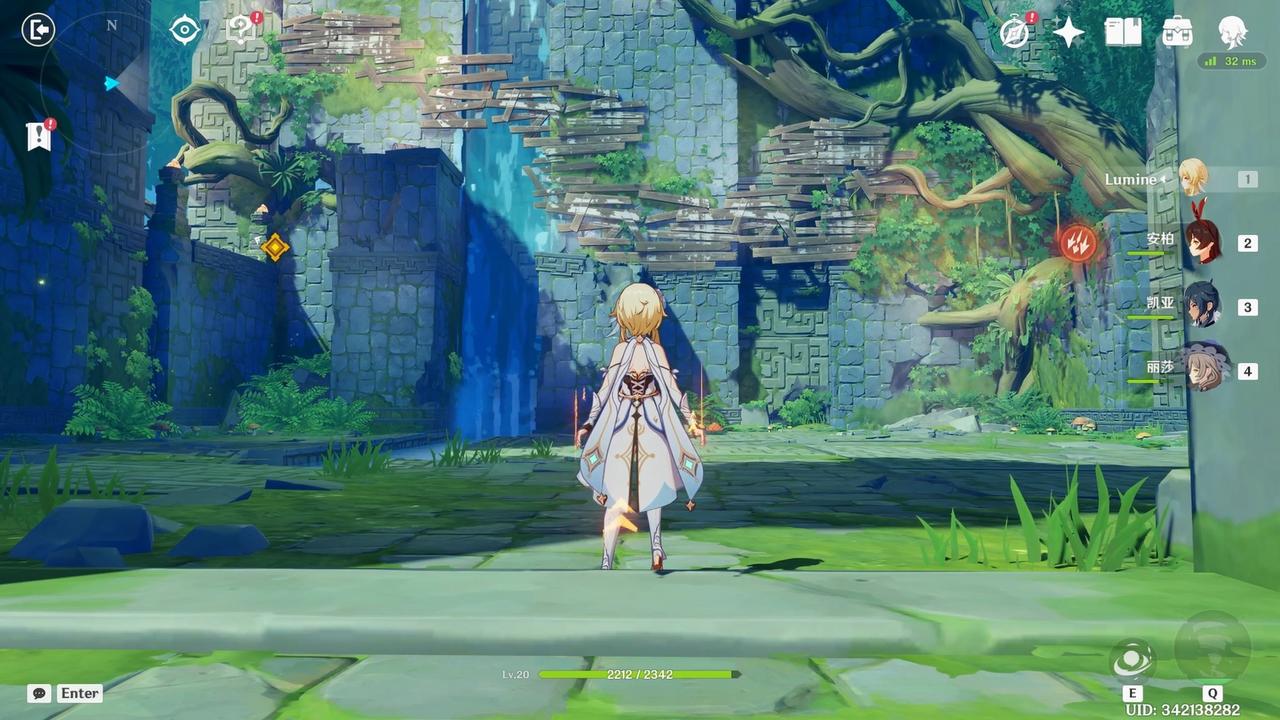}
        \end{subfigure}%
        \hfill
        \begin{subfigure}[b]{0.24\textwidth}
            \includegraphics[width=\textwidth]{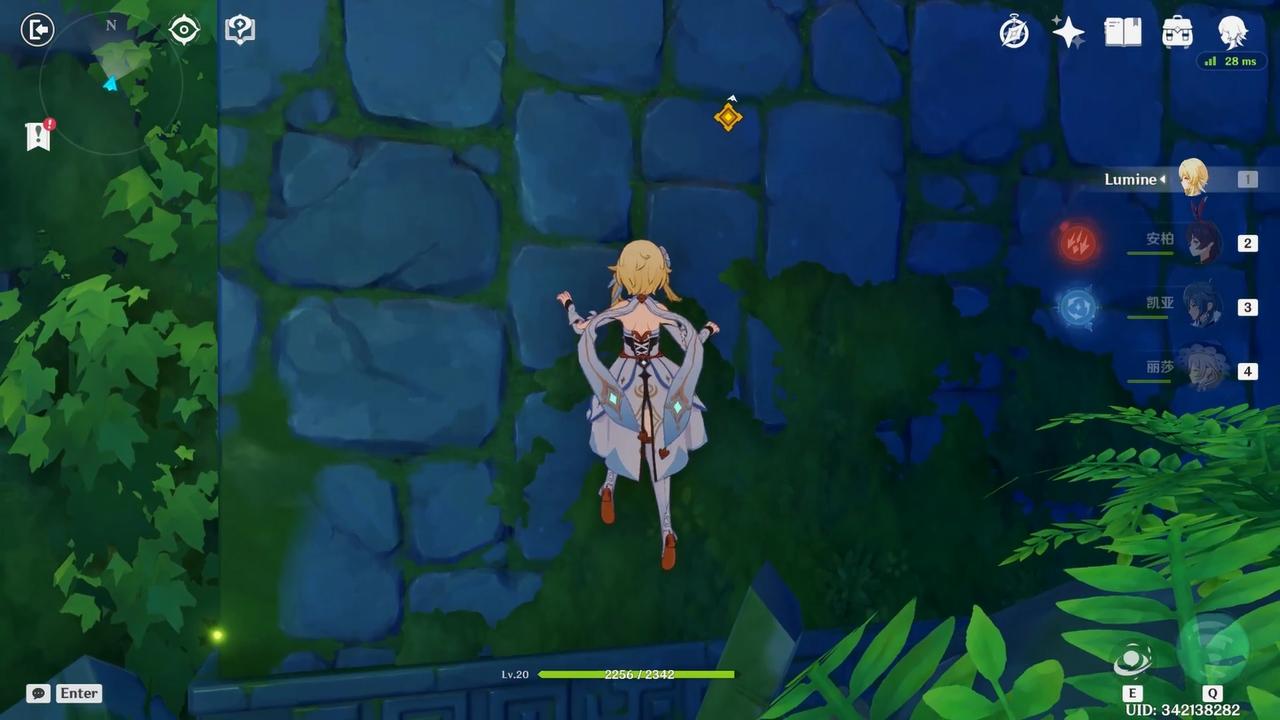}
        \end{subfigure}%
        \hfill
        \begin{subfigure}[b]{0.24\textwidth}
            \includegraphics[width=\textwidth]{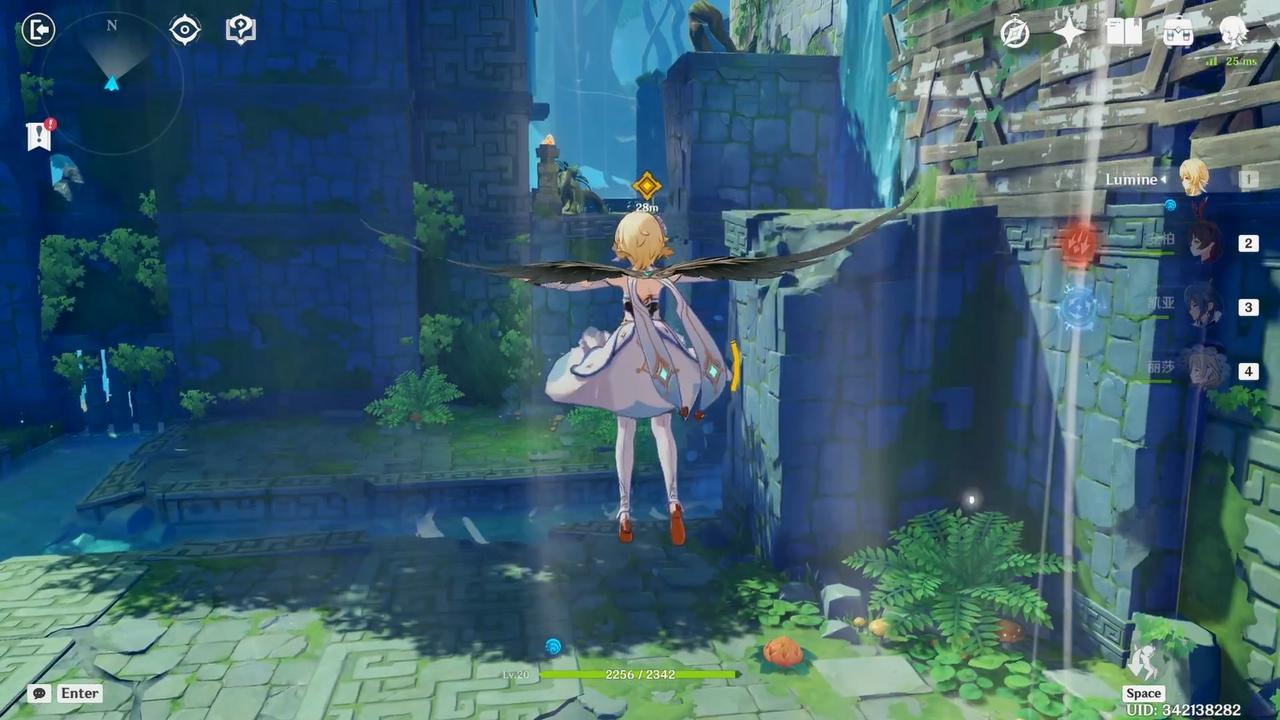}
        \end{subfigure}%
        \hfill
        \begin{subfigure}[b]{0.24\textwidth}
            \includegraphics[width=\textwidth]{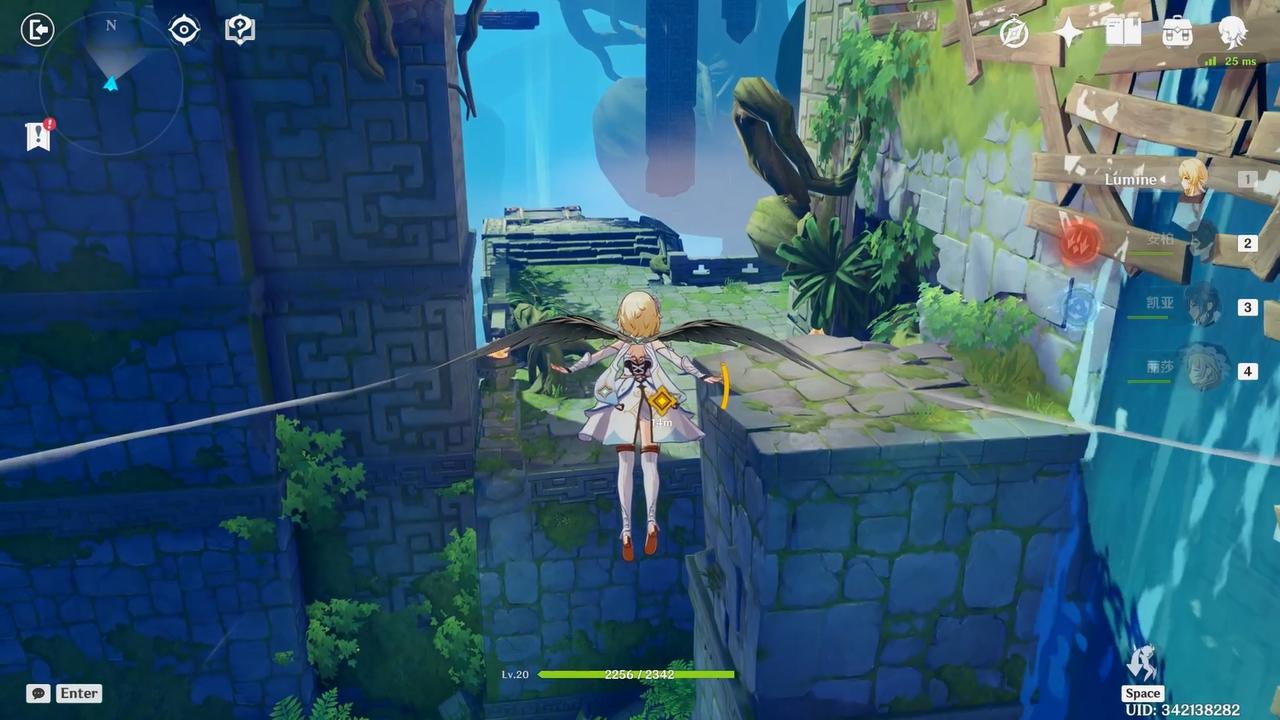}
        \end{subfigure}%
        \caption{When following the golden quest marker directly, the agent encounters a wall and must take a detour using the wind current on the right. Lumine often abandons the detour midway and returns to the starting point, drawn back by the quest marker’s signal since it has forgotten that the direct path is blocked.}
    \end{subfigure}
    \caption{Two examples illustrating the importance of long-term memory for successful task completion.}
    \label{fig:ood_memory_limitation}
\end{figure}

From Lumine’s gameplay, we identify five key factors that account for the primary inefficiencies preventing it from achieving expert human-level performance in these missions:

\begin{itemize}
    \item \textbf{Lack of proactive fast travel.} Lumine rarely makes use of teleportation for fast travel, instead following quest markers on foot even across long distances. However, we did observe a few successful cases of teleportation, suggesting that the capability already exists but is not triggered properly. This is mainly due to the lack of such a pattern in the first hour of gameplay in the reasoning data. 
    
    \item \textbf{Limited understanding of the minimap.} While Lumine shows some awareness of the minimap, its use is unreliable, leading to heavy dependence on the golden quest marker for navigation. If the quest marker disappears and Lumine fails to generate a new reasoning to recall it again, the agent is likely to move in the wrong direction and stray far from the intended target.  
    
    \item \textbf{Lack of proactive health recovery.} Lumine does not actively restore the health of party members. This omission is closely related to the first issue, as the agent does not teleport to Statues of The Seven (which can heal the party) nor open the inventory to consume food items for healing. These recovery patterns are also notably absent from the reasoning data.  
    
    \item \textbf{Limited memorization.} As shown in Figure~\ref{fig:ood_memory_limitation}, in Act II and III, certain missions present multiple simultaneous quest markers shown in the image, and Lumine is easily distracted, oscillating between different targets. Additionally, in cases where the quest marker is directly ahead but requires a detour due to obstacles, Lumine often abandons the detour midway and returns to the starting point, drawn back by the quest marker’s signal. Although Lumine demonstrates diverse strategies, strong exploration ability, and eventually manages to get out of such situations, these behaviors highlight the limitation of its four-second (20 frames) memory span and underscore the need for more effective long-term memory mechanisms.

    \item \textbf{Combat proficiency to be improved.} Although Lumine achieves a high success rate of combat across various combinations of enemies in the benchmark, its efficiency remains limited in aspects such as skill coordination, aiming accuracy, dodging timing and mechanics understanding. These shortcomings become more pronounced in multi-wave encounters and boss fights, where Lumine performs noticeably below the level of expert human players.
\end{itemize}

\textbf{Generalization to Fully OOD Mission}. We further extend the evaluation to missions entirely absent from the training data, thereby testing the agent’s ability to generalize to completely novel scenarios. After successfully completing the full Mondstadt main storyline, the next mission directs the player to travel to the nation of Liyue. As shown in Figure~\ref{fig:map23_show}, Agents are then seamlessly tasked with journeying from Mondstadt to Liyue Harbor and advancing as far as possible in the Liyue main storyline. Since Liyue is entirely excluded from the training data, it presents a challenging OOD environment with new regions, stories, NPCs, enemies, and puzzles. 
Compared to the relatively flat terrain of Mondstadt, the mountainous landscape of Liyue poses significantly greater challenges for the agent. 

Lumine impressively demonstrates in-domain level efficiency during the first hour of gameplay, where it must complete an extremely long-distance navigation from Mondstadt to Liyue Harbor, evade capture by the Millelith, and locate Tartaglia hidden on the second floor. Though less efficient, Lumine even manages to find the Adeptus dwelling deep within the mountains, after a long journey across rugged terrain and rivers with highly noticeable ups and downs. On the way to visit the second Adeptus, Lumine accidentally cancels quest tracking due to a hallucination. When attempting to reactivate it, the model outputs a left-click command combined with mouse movement, which the game misinterprets as a drag action, causing the operation to fail. However, the agent is overconfident that the task had been activated and subsequently closed the quest menu, leading to aimless exploration of the map for approximately two hours. Eventually, it manages to reopen the quest interface, activate quest tracking, locate the second Adeptus, rescue the person trapped in amber, and complete the mission.

\begin{figure}[t]
    \centering
    \begin{subfigure}[b]{0.3\textwidth}
        \centering
        \includegraphics[width=\textwidth]{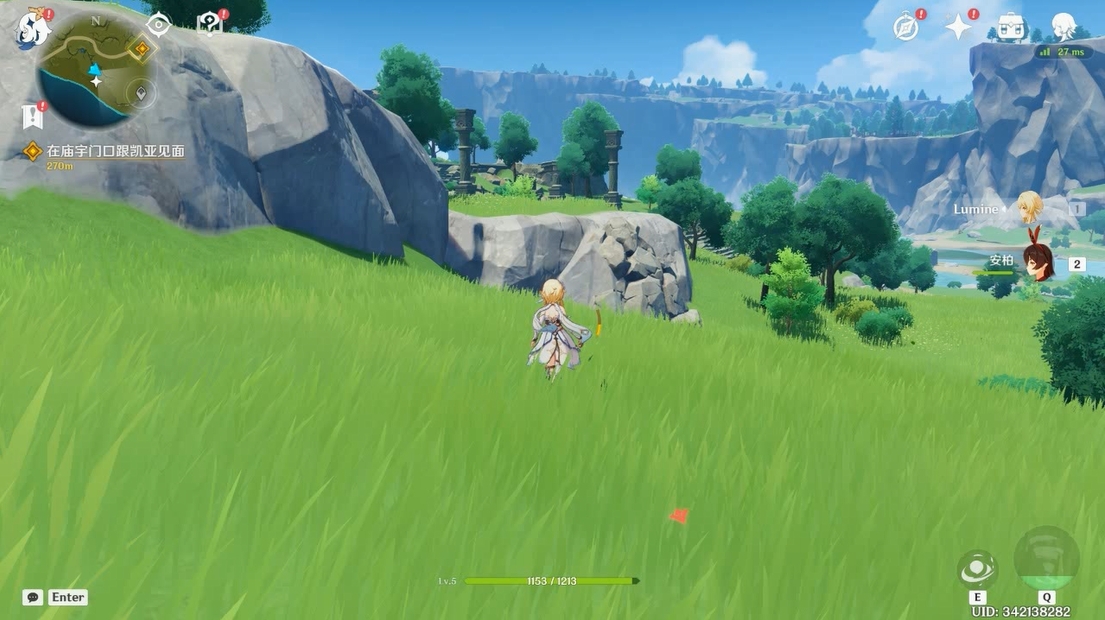}
    \end{subfigure}
    % \hfill
    \begin{subfigure}[b]{0.3\textwidth}
        \centering
        \includegraphics[width=\textwidth]{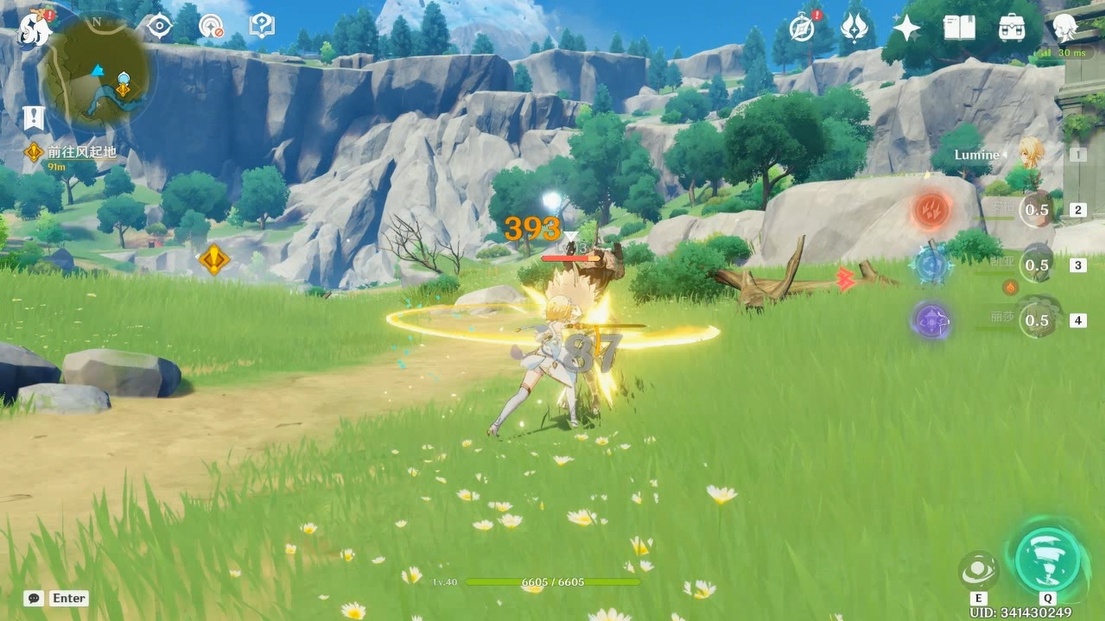}
    \end{subfigure}
    % \hfill
    \begin{subfigure}[b]{0.3\textwidth}
        \centering
        \includegraphics[width=\textwidth]{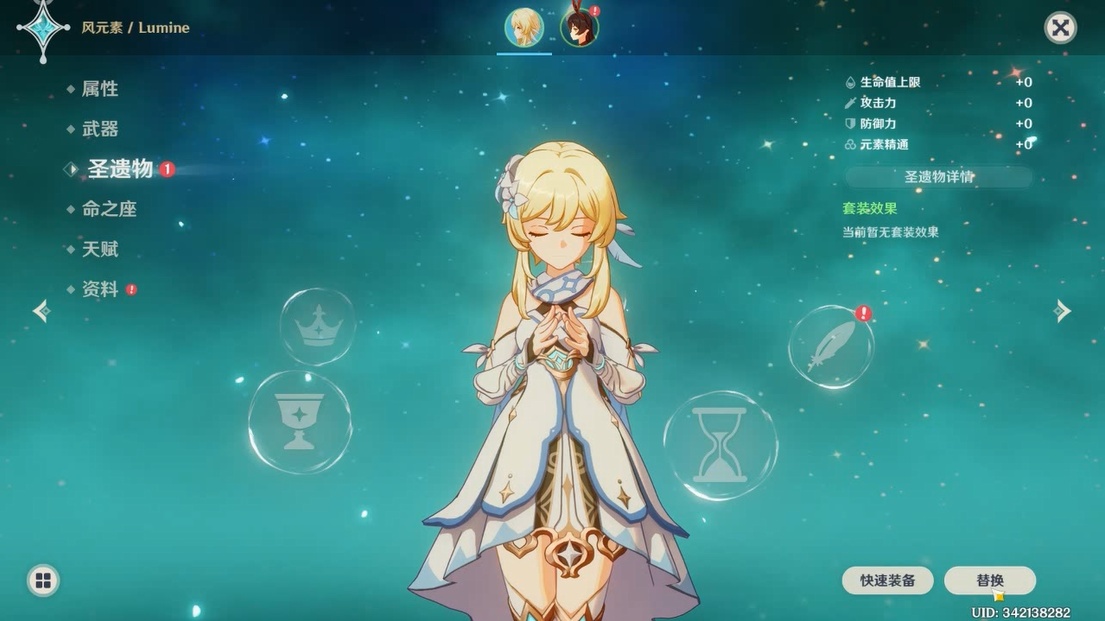}
    \end{subfigure}
    \\[2ex] 
    
    \begin{subfigure}[b]{0.3\textwidth}
        \centering
        \includegraphics[width=\textwidth]{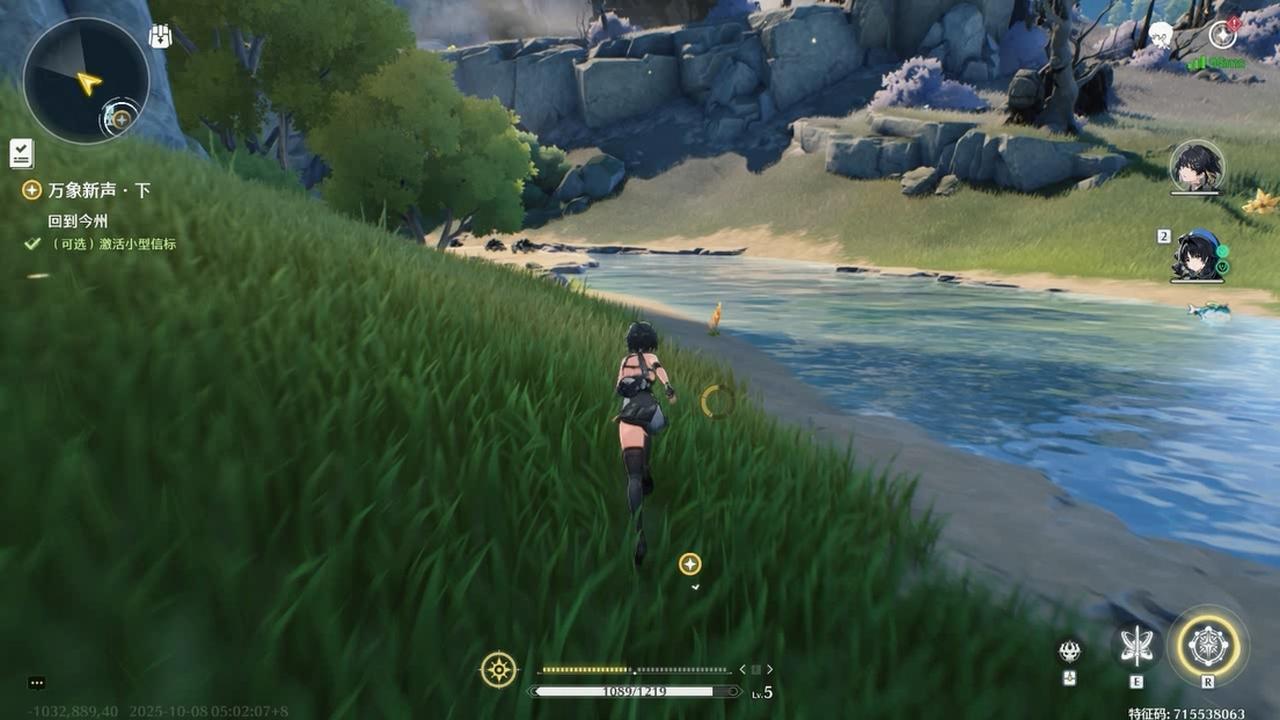}
    \end{subfigure}
    % \hfill
    \begin{subfigure}[b]{0.3\textwidth}
        \centering
        \includegraphics[width=\textwidth]{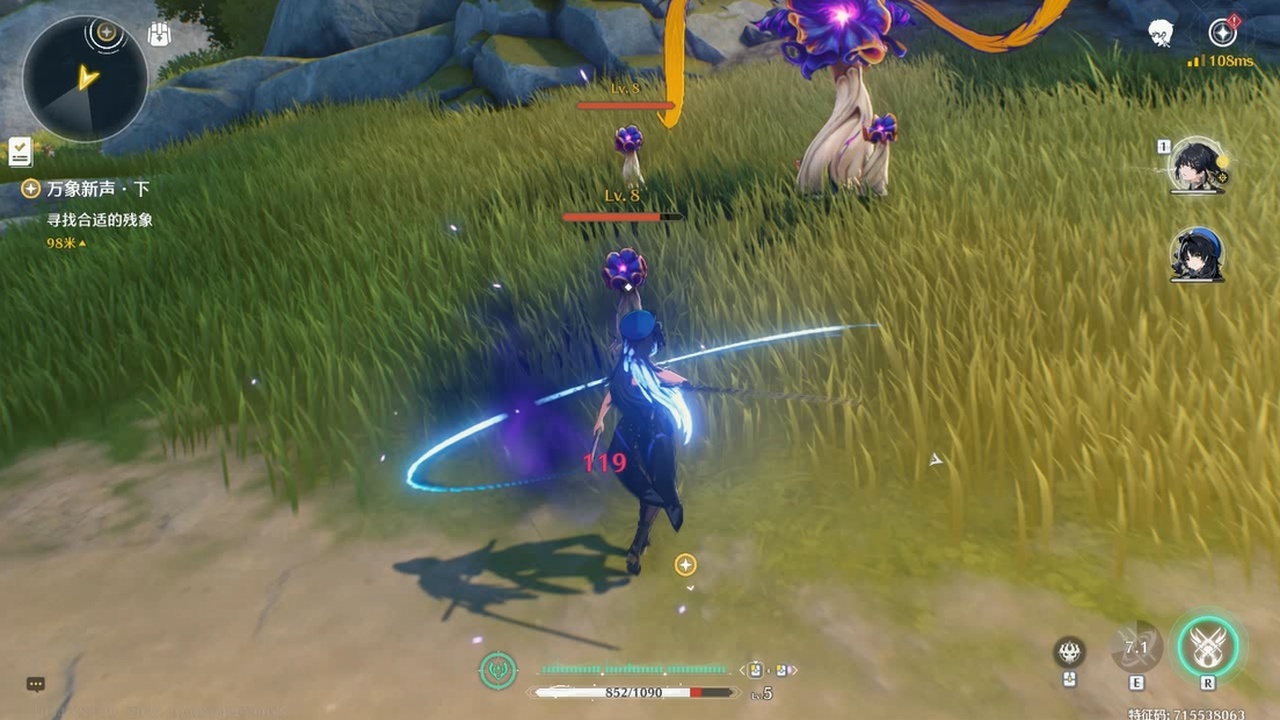}
    \end{subfigure}
    % \hfill
    \begin{subfigure}[b]{0.3\textwidth}
        \centering
        \includegraphics[width=\textwidth]{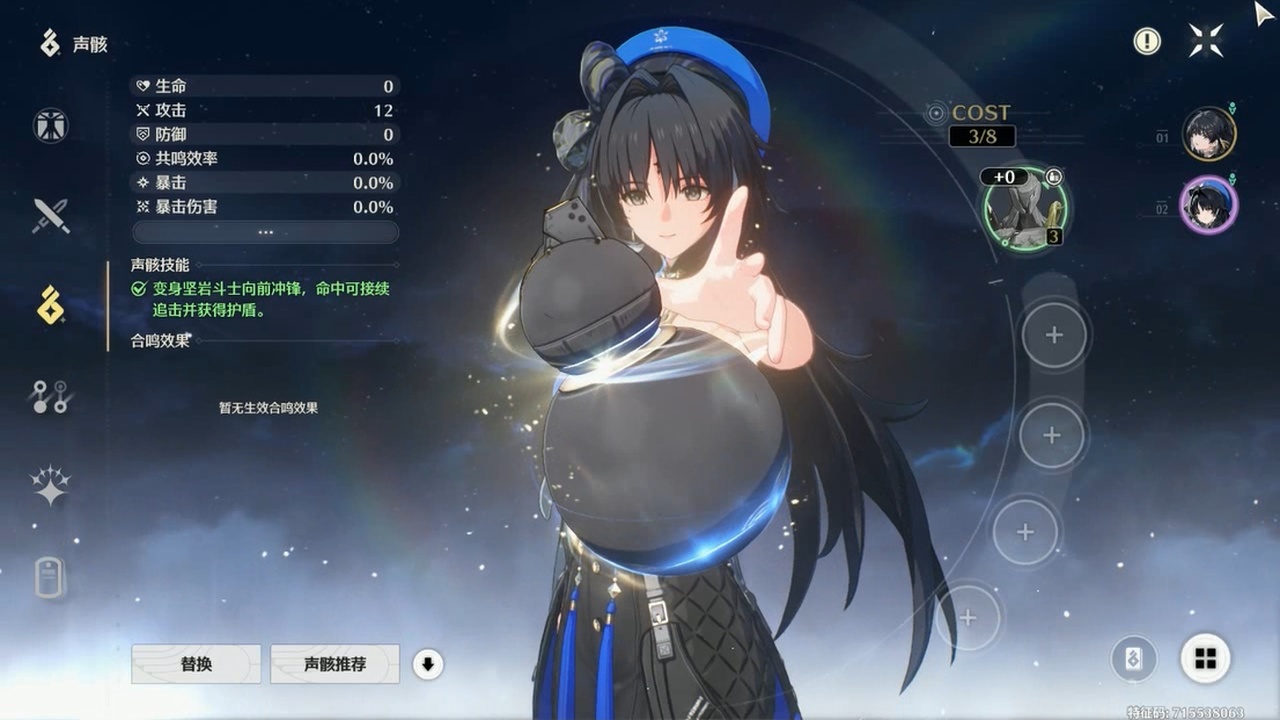}
    \end{subfigure}
    \\[2ex] 

    % 第二行
    \begin{subfigure}[b]{0.3\textwidth}
        \centering
        \includegraphics[width=\textwidth]{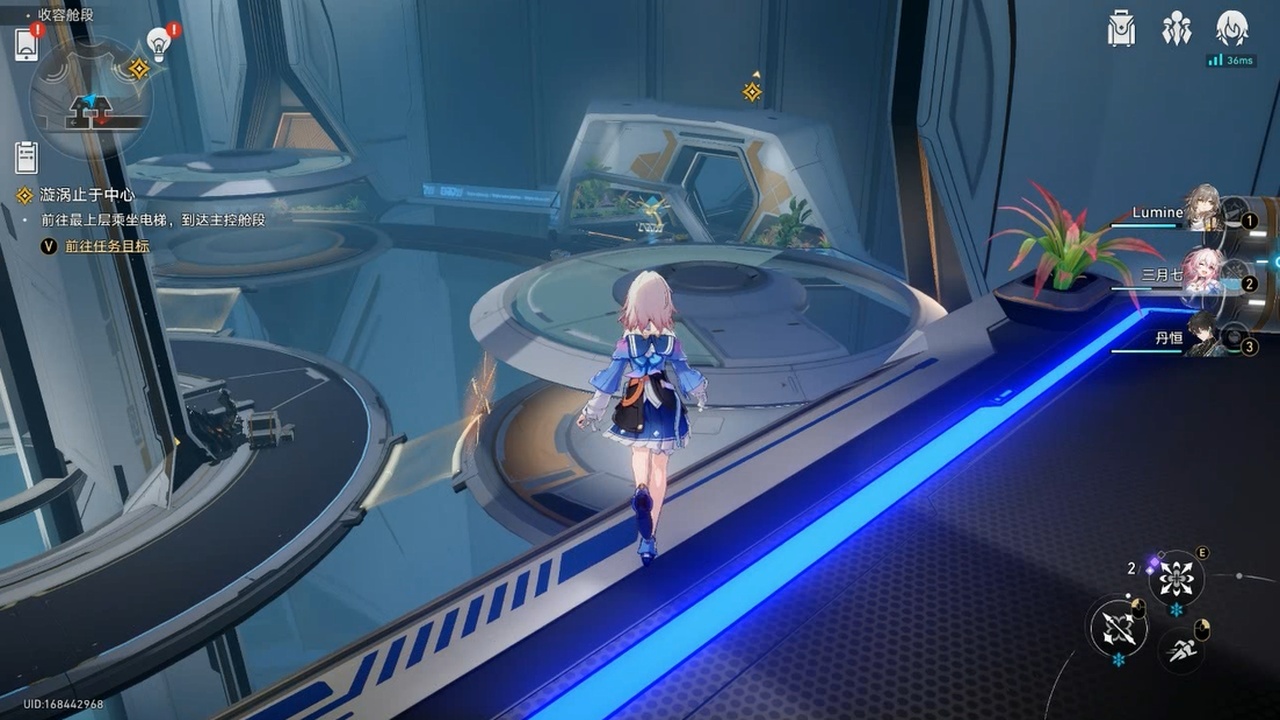}
    \end{subfigure}
    % \hfill
    \begin{subfigure}[b]{0.3\textwidth}
        \centering
        \includegraphics[width=\textwidth]{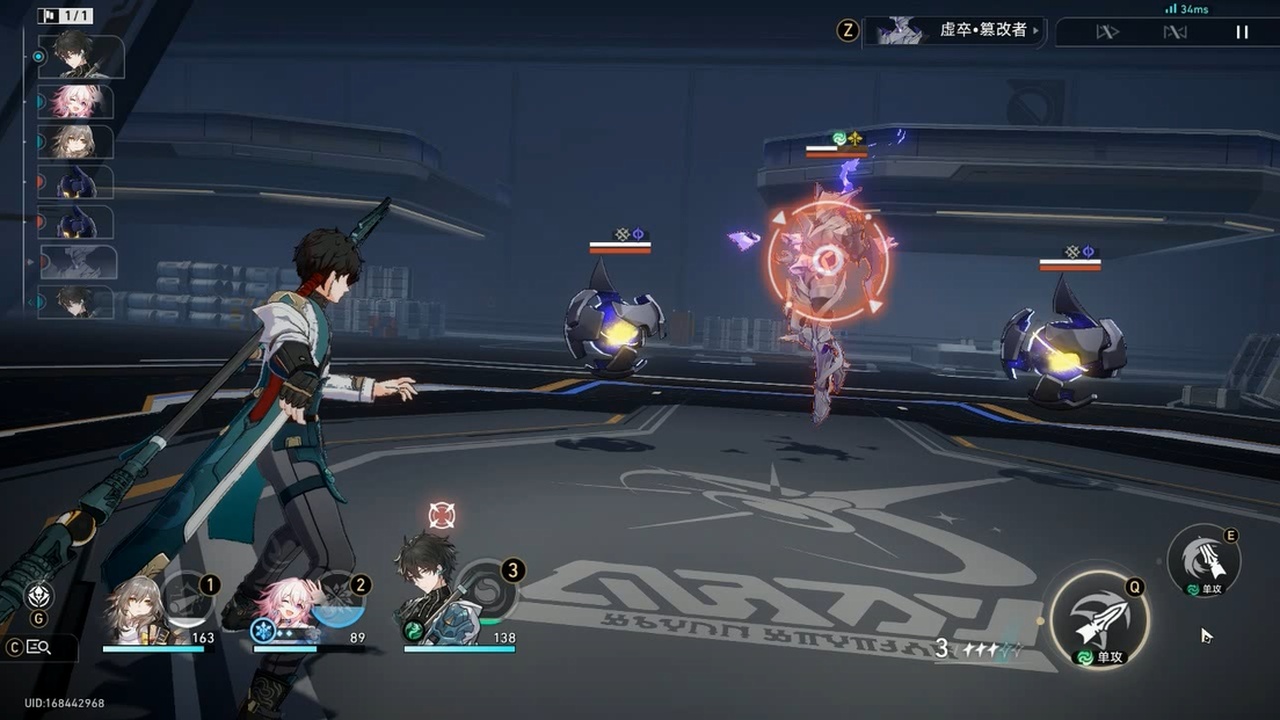}
    \end{subfigure}
    % \hfill
    \begin{subfigure}[b]{0.3\textwidth}
        \centering
        \includegraphics[width=\textwidth]{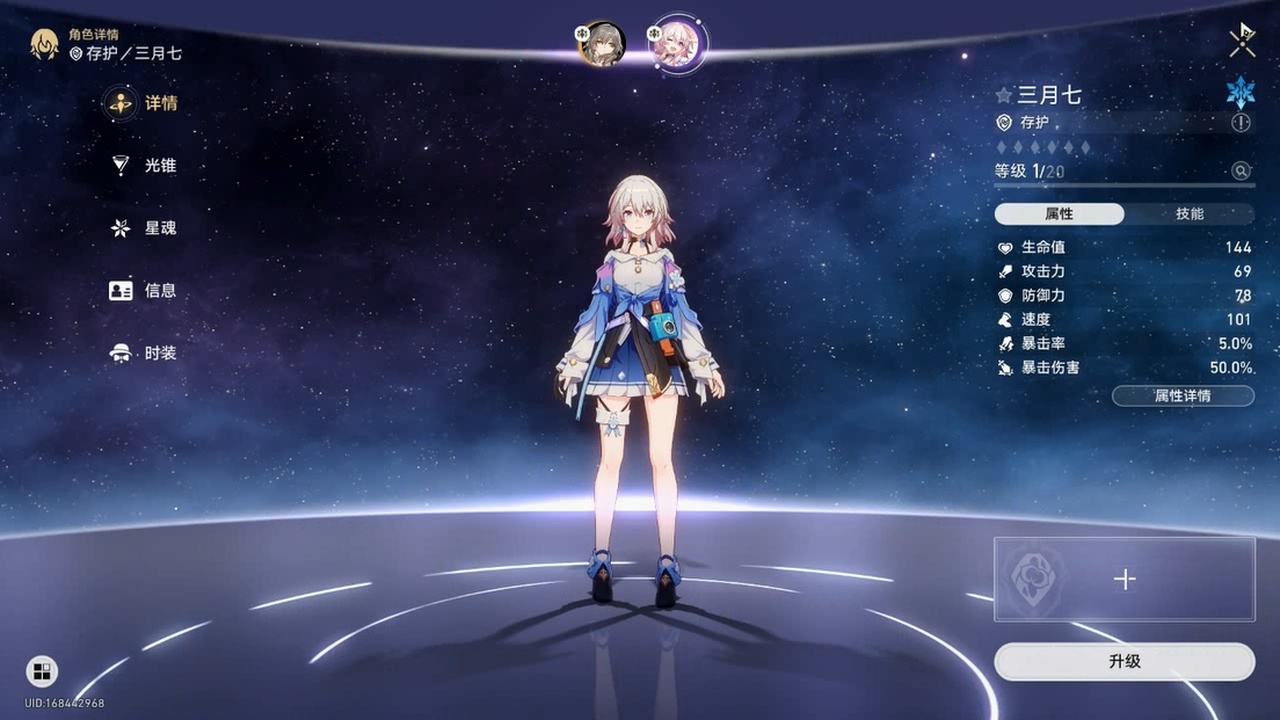}
    \end{subfigure}
    \\[2ex]

    % 第三行
    \begin{subfigure}[b]{0.3\textwidth}
        \centering
        \includegraphics[width=\textwidth]{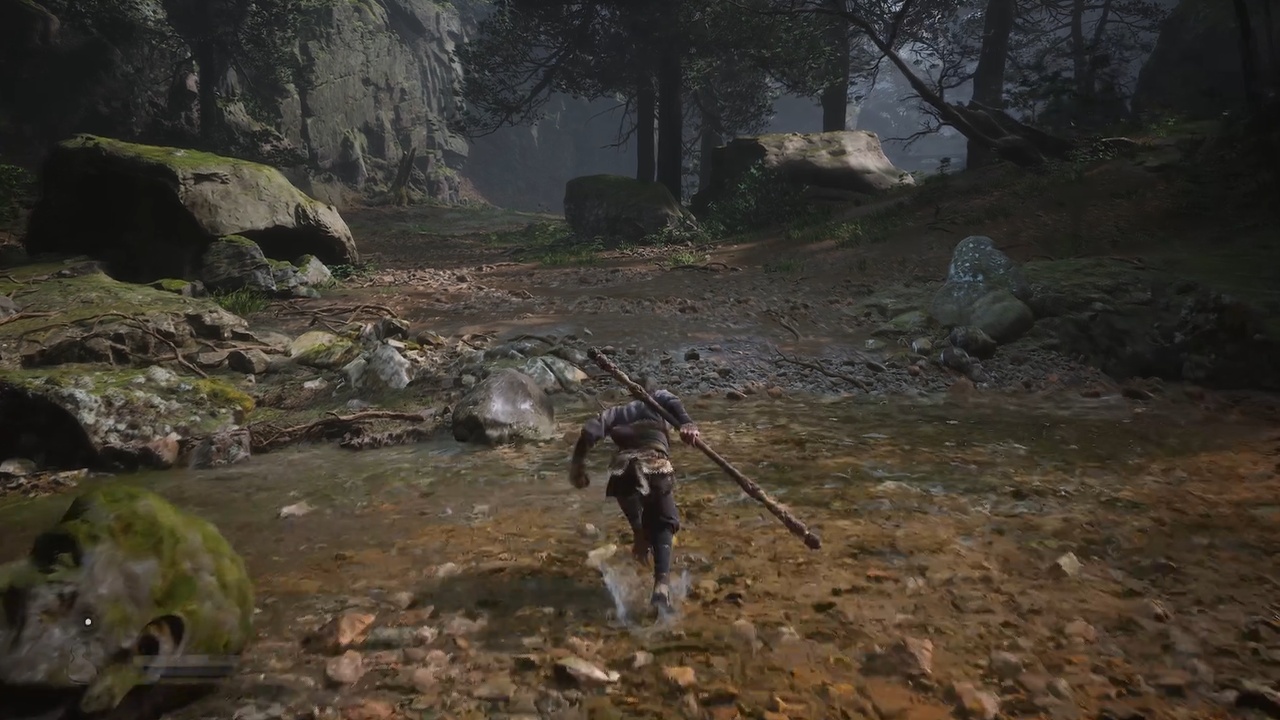}
    \end{subfigure}
    % \hfill
    \begin{subfigure}[b]{0.3\textwidth}
        \centering
        \includegraphics[width=\textwidth]{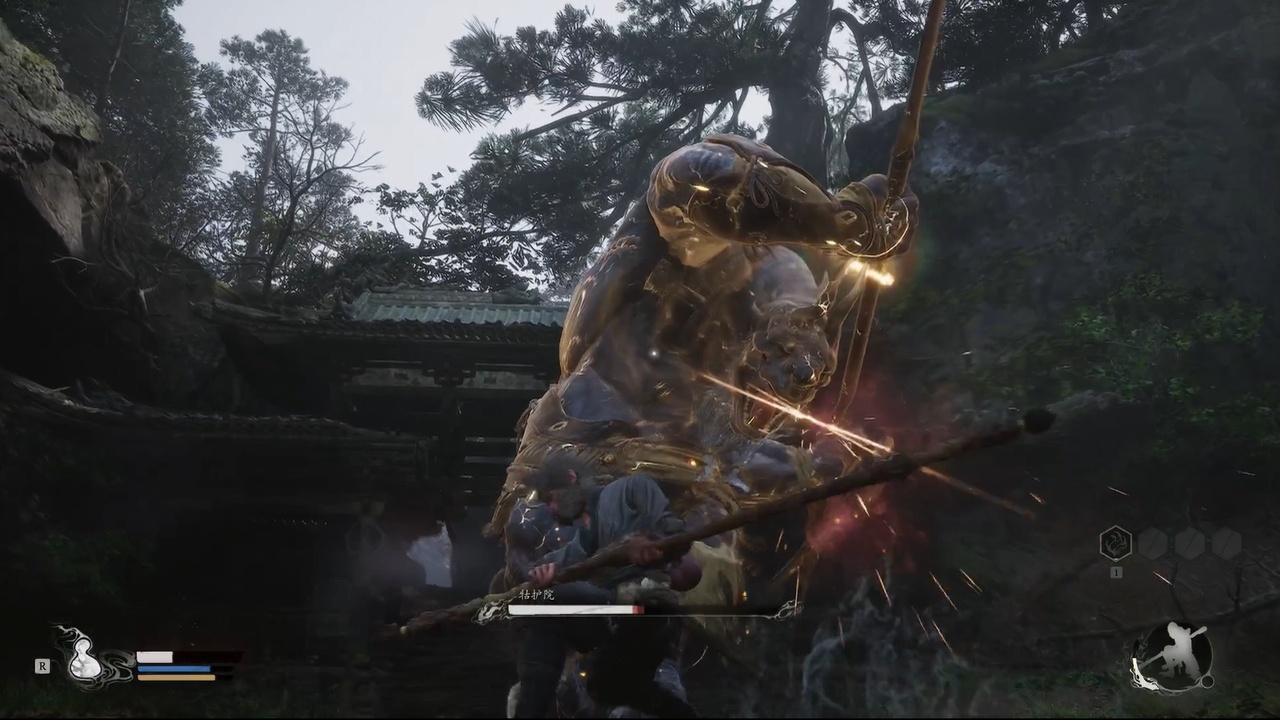}
    \end{subfigure}
    % \hfill
    \begin{subfigure}[b]{0.3\textwidth}
        \centering
        \includegraphics[width=\textwidth]{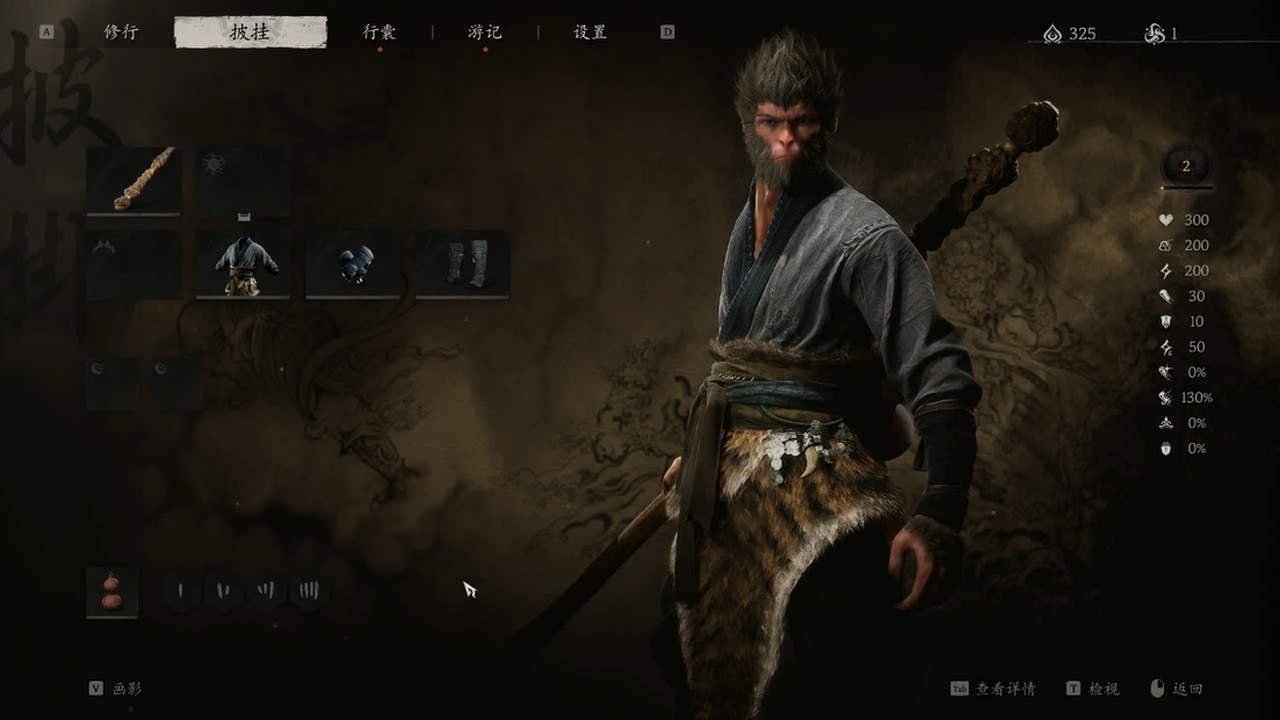}
    \end{subfigure}

    \caption{Demonstration of overworld navigation, combat, and UI in four games, Genshin Impact, Wuthering Waves, Honkai: Star Rail, and Black Myth: Wukong. Similar to Genshin Impact, Wuthering Waves is also an open-world ARPG, while Honkai: Star Rail is a turn-based RPG that combines strategic combat with a hub-based world design. Black Myth: Wukong is a hub-based ARPG but features a more realistic visual rendering style. }
    \label{fig:new_game_intro}
\end{figure}

\subsubsection{Generalization to New Games}
Finally, we investigate whether Lumine, though trained exclusively on Genshin Impact gameplay data, can generalize and be deployed in other unseen games without additional training and without any prompt modification. As shown in Figure~\ref{fig:new_game_intro}, we carefully select three games, \textit{Wuthering Waves}, \textit{Honkai: Star Rail} and \textit{Black Myth: Wukong}, that represent diverse genres, mechanics, and levels of similarity to Genshin Impact.

\textbf{Wuthering Waves.} Similar to Genshin Impact, Wuthering Waves is also an online 3D open-world ARPG that shares many structural similarities in terms of control schemes, combat loops, and gameplay rhythm. This makes it a natural testbed for measuring how well Lumine adapts to games that are mechanically aligned but distinct in content and setting. For our evaluation, Lumine began with a fresh account and started from the very beginning of the game. We would also like to mention that the launch date of Wuthering Waves was later than the knowledge cutoff of Qwen2-VL, so Lumine is unlikely to leverage prior knowledge about this game from the base model pretraining.

In this unseen game, Lumine impressively demonstrated in-domain-level efficiency during the first two main storyline missions, \textit{First Resonance} and \textit{Echoing Marche}, where Lumine completes these missions in \textbf{107} minutes, compared to an average of \textbf{101} minutes for fresh human players.\footnote{The estimated average playtime is based on five human players and should be regarded only for reference. All the players have extensive gaming experience, but none of them played this game before. Players have different playstyles: some tend to skip all the dialogues and the story, while others prefer to immerse themselves fully in the narrative. Some focus primarily on the main storyline, whereas others take time to clear enemies, solve puzzles, and collect chests along the way. As a result, gameplay length can vary significantly.} Lumine’s diverse capabilities generalize seamlessly to these unseen game missions, in both 3D overworld and 2D GUI, enabling it to follow in-game guidance to reach quest target location, perform key presses and icon selections according to hints, defeat enemies, unlock fast travel points, interact with objects and NPCs, level up characters, manage equipment and explore the open world.

While generally efficient, several mistakes were observed. A notable issue arises from the game’s distinctive rendered style: the agent occasionally misinterprets the on-screen prompt to press key \texttt{F} for interaction as key \texttt{E}. Since \texttt{E} triggers skill usage, this misinterpretation can cause gameplay stalls, highlighting the importance of robust OCR capabilities. Other errors come from domain bias and hallucination in reasoning, where Lumine borrows terminology from Genshin Impact to describe entities in Wuthering Waves. For instance, it refers to monsters as Hilichurls, the protagonist Rover as Traveler, and the red-clad girl in the team as Amber. Interestingly, such cross-game naming confusion is also commonly seen among human players.

\textbf{Honkai: Star Rail.} Different from the previous two games, Honkai: Star Rail is an online 3D turn-based strategy role-playing game. Instead of a fully open-world structure, it adopts a hub-based exploration design, in which the game world is divided into discrete zones connected through hubs rather than presented as a seamless world. Movement is also more restricted, as characters cannot jump or climb over obstacles. While the game retains some control schemes and interface conventions from Genshin Impact, its core combat system emphasizes turn-based strategy over real-time action.

Although less efficient than Wuthering Waves due to the significant domain gap, Lumine still surprisingly demonstrates reasonable performance in 3D navigation, NPC interaction, and GUI manipulation, abilities that are essential for progression. Lumine successfully completed the entire first chapter, \textit{Today is Yesterday's Tomorrow}, in the Herta Space Station, win the Boss fight of Doomsday Beast, cleared the Simulated Universe tutorial, and even progressed into the second chapter, reaching a new planet, Jarilo-VI, using \textbf{7} hours, compared to the \textbf{4.7} hours typically taken by fresh human players. 

The main bottlenecks hindering smooth gameplay arise in navigation and combat, primarily due to domain bias. In several scenarios, Lumine attempts to jump across gaps toward a quest marker on the opposite platform, only to be blocked by an invisible wall. While such behavior works in Genshin Impact, it is impossible in Honkai: Star Rail, which lacks a jump mechanic and is not an open-world. Although Lumine sometimes adapts by taking an alternate route, it more often returns to the same spot, repeatedly drawn by the quest marker and constrained by limited memorization. Encouragingly, after enough attempts, Lumine eventually manages to get unstuck and reach the correct location, despite the distraction.  

Combat presents even greater challenges. While Lumine can reliably distinguish allies from enemies, the turn-based combat system differs drastically from real-time combat in Genshin Impact and causes the greatest delays. Fortunately, the two games share some overlapping keybinds: in Genshin Impact, normal attacks use mouse clicks, Elemental Skills use \texttt{E}, Elemental Bursts use \texttt{Q}, and characters can be swapped with numeric keys \texttt{1--4}. In Honkai: Star Rail, however, \texttt{Q} is used for normal attacks, \texttt{E} for skills, and \texttt{1--4} for each character’s ultimate ability. This overlap allows Lumine to stumble through combat using its prior Genshin experience, but targeted skills pose a serious obstacle. It fails to understand that pressing \texttt{E} requires selecting a target, or that using an ultimate requires confirming with the spacebar. Still, Lumine occasionally presses the correct keys by chance, allowing battles to continue.  These difficulties were most pronounced in the final boss fight against the Doomsday Beast, where repeated misunderstandings led to multiple party wipes. Eventually, Lumine chose to lower the game’s difficulty and, by the narrowest margin, secured victory in the final battle.

\textbf{Black Myth: Wukong.} 
Finally, we evaluate Lumine in a much more challenging game, Black Myth: Wukong, a single-player 3D action role-playing game with highly realistic graphics, in contrast to the more stylized, animated visuals of the previous titles. While Lumine demonstrates basic competence in both navigation and combat, several factors prevent it from completing longer tasks.

Firstly, similar to Honkai: Star Rail, the game uses a hub-based exploration design rather than an open world. The lifelike scenery combined with pervasive invisible walls often causes Lumine to get stuck, mistakenly assuming it can pass through. Secondly, during navigation, the overworld UI elements such as health and status indicators will automatically hide. The cinematic visuals lead Lumine to misinterpret the scene as a pre-rendered cutscene (CG), at which point it keeps outputting noop actions and waits for the "CG" finish. Thirdly, Lumine struggles to correctly recognize the health bar and has no understanding of how to restore health. This limitation becomes especially punishing in the game’s high-difficulty, Souls-like combat, where survival requires enduring multiple waves of enemies. Taken together, these challenges make it particularly difficult for a zero-shot agent to play the game.

\textbf{Conclusion}. Lumine demonstrates strong generalization abilities across different games, even as the similarity between game genres decreases. This indicates that despite variations in visual style and gameplay mechanics, the core skills of navigation, combat, and 2D GUI manipulation can transfer across games, making it a promising foundation model across games. Remarkably, even when trained on a single game, Lumine is able to adapt, highlighting its promising scalability for broader applications.

%% file: sections/6_conclusion.tex
\section{Discussion and Future Work} 
We proposed Lumine, an open recipe that spans the entire lifecycle of developing generalist agents in 3D open world environments, from environment selection, data collection and preprocessing to interaction frequency, model design, training procedures, and inference optimization. Lumine offers a unified framework that organically integrates perception, reasoning, and action. The success of Lumine demonstrates that with only 2400 hours of raw gameplay data and 64 H100 GPUs, a small open-source VLM can be seamlessly and efficiently transformed into a powerful agent capable of following natural language instructions in real time to perform diverse tasks and complete hours-long missions in complex 3D open-worlds, without the need to modify model structures or loss functions. Experimental results also reveal substantial potential for further scaling. Lumine’s remarkable zero-shot generalization to unseen missions and even entirely unseen games suggests that the model acquires transferable meta-skills, such as 3D navigation and 2D manipulation, that can be readily applied to other domains. This underscores the promise of Lumine’s recipe as a pathway toward developing general-purpose decision foundation models.

Lumine is not without its limitations, which also point to several promising directions for future improvement:

\begin{itemize}
    \item \textbf{Scaling.} To experimentally validate the effectiveness of the Lumine recipe, the pre-training data was limited to the Mondstadt region of Genshin Impact, and the reasoning data covered only the first hour of gameplay. There is significant potential to scale both the pre-training and reasoning datasets, not only within Genshin Impact but also across other games and domains, to enhance the model’s generalization and robustness.
    
    \item \textbf{Long-Term Memory.} Lumine currently employs a straightforward memory management mechanism, using previous reasoning steps for long-term memory and a 20-frame context window for short-term memory. While this setup proves effective in most cases, it remains inadequate for complex, long-horizon missions. Future work should explore more sophisticated approaches that enable efficient memory retrieval and management over thousands turns of interactions.
    
    \item \textbf{Online Learning.} Lumine primarily learns from offline data, which is efficient but limits its ability to surpass human-level performance. Building upon the strong foundation of the existing Lumine model, integrating online reinforcement learning could enable autonomous exploration and continuous self-improvement, further enhancing its performance beyond static, offline learning.

    \item \textbf{Real-Time Inference.} To meet the strict latency requirements of real-time interaction, Lumine applies tensor parallelism across multiple GPUs for acceleration and still suffers from the inference delay. More efficient inference strategies are needed to reduce computational overhead. Improved efficiency would not only facilitate real-time responsiveness but also benefit reinforcement learning through faster rollout generation.
\end{itemize}

Beyond its research-oriented contributions, Lumine demonstrates strong potential for practical applications. With its generalist perception–reasoning–action capabilities, Lumine can autonomously explore game environments or interpret natural-language instructions to execute gameplay tasks, supporting debug detection, quality assurance, and large-scale usability evaluation in game development. It can identify inconsistencies and ambiguities in tutorials, quest logic, and interaction design, helping developers reduce manual testing costs and improve overall development efficiency. Meanwhile, Lumine also opens up possibilities for new forms of interactive entertainment, such as AI-driven game assistants and streaming where intelligent agents are capable of playing, commenting, and engaging with audiences in real time. These applications highlight Lumine’s potential to blur the boundary between player, developer, and audience, paving the way for new human–AI co-experiences in digital worlds.

\section*{Ethics Statement}
This work strictly adheres to the ethical standards in research involving artificial intelligence systems, human gameplay data, and virtual environments.

All gameplay data used in this study were collected from consenting adult participants who were compensated at fair market rates. Participants were informed that their keyboard, mouse, and screen recordings would be used solely for research on AI agents. 

We strictly complied with the terms of service of all games used in this research, and we emphasize that our system is developed for academic research only. Lumine is not intended for cheating or competitive advantage in any commercial product. We explicitly oppose any use of this technology that undermines fair play, game integrity, or player experience.

While we acknowledge the potential risk that such AI systems could be misused as game cheats or unauthorized automation tools, we also recognize that automation and general intelligence represent an inevitable and transformative direction for the future, one that may not only reshape video games but also fundamentally influence how humans and intelligent agents coexist across digital and physical domains.

Therefore, instead of merely preventing misuse, we call for an open, collaborative dialogue within the research, developer, and player communities to explore new frameworks and governance models that balance innovation, creativity, and fairness. We believe that by working together, the community can establish shared ethical standards and technological safeguards that enable the positive and responsible evolution of intelligent agents and game development.

\section*{Acknowledgement}

We thank Yi Lin, Tianheng Cheng, Yudong Liu, Jingjia Huang for providing valuable suggestions on VLM training. We thank Faming Wu, Xiaojun Xiao, Junjie Fang, Junda Zhang, Yuxing Yao, Yang Yang, and Xiaoying Jia for their assistance in resource allocation, storage management, model training and inference optimization. We thank Zihao Wang for discussions on model design. We thank Xinrun Wang for discussions on paper writing.  We thank Bo Zhou, Zili Li, Yu Miao, Woyu Lin, and Qiaoyu Dong for their contributions to annotator recruitment and management. We thank Heng Zhang and Xiaoyu Tang for their help with local evaluation. We also thank Chao Wu for preparing game accounts used in demo recordings.

\phantomsection
\section*{Contributions}
% \addcontentsline{toc}{section}{Contributions}
\label{sec:contribution}
\textbf{Authors} Weihao Tan$^{2, *, \dagger}$, Xiangyang Li$^{3, *, \dagger}$, Yunhao Fang$^{1, *, \ddagger}$, Heyuan Yao$^{3, *, \dagger}$, Shi Yan$^{3, *, \dagger}$, Hao Luo$^{3, *, \dagger}$, Tenglong Ao$^{1}$, Huihui Li$^{1}$, Hongbin Ren$^{1}$, Bairen Yi$^{1, \ddagger}$, Yujia Qin$^{1}$, Bo An$^{2}$, Libin Liu$^{3}$, Guang Shi$^{1}$

\textbf{Affiliations} $^{1}$ByteDance Seed, $^{2}$Nanyang Technological University, $^{3}$Peking University

$^{*}$ Core contributors \\
$^{\dagger}$ Work was done during their internship at ByteDance Seed \\
$^{\ddagger}$ Work performed while at ByteDance Seed

%% file: sections/appendix.tex
\section{Gameplay Collection}
\label{app:data_collection}

In this section, we describe the recruitment and annotation process for collecting gameplay data. All annotators were native Chinese and played the Chinese version of the game, in which all in-game text and dialogues were displayed in Chinese. Consequently, all curated data, including instruction-following and reasoning annotations, were also produced in Chinese.

\subsection{Participant Recruitment}

To ensure the collection of high-quality gameplay data, we recruited annotators with proficient gameplay abilities and strong comprehension skills. The specific criteria were as follows: 

\begin{enumerate}
    \item \textbf{Game Experience:} Possess an in-game account with an Adventure Rank of 45 or higher in Genshin Impact; or be a frequent player of PC-based 3D games, such as MMORPGs, 3D RPGs, or other popular titles.
    \item \textbf{Age:} Between 18 and 40 years old.
    \item \textbf{Education:} An associate degree or higher.
    \item \textbf{Hardware:} To ensure smooth gameplay and stable graphics rendering, participants’ devices must meet or exceed the following specifications:
    \begin{itemize}
        \item Operating System: Windows 10 or Windows 11
        \item CPU: Intel Core i5-12400 or better
        \item Memory: 16 GB or higher
        \item GPU: NVIDIA GeForce GTX 1060 (6 GB) or better
        \item Network: In-game latency below 100 ms
    \end{itemize}
\end{enumerate}

\subsection{Annotation Tasks}
Annotators are asked to start with a brand-new account and sequentially complete the following objectives:

\begin{itemize}
\item \textbf{Task 1:} Complete Act I of the Mondstadt main storyline. ($\approx$1h)

\item \textbf{Task 2:} Progress through the remianing main storyline (Act II \& III) of the Mondstadt region. ($\approx$15h)

\item \textbf{Task 3:} Achieve over 80\% map exploration in each region of Mondstadt. ($\approx$14h)
\end{itemize}

On average, this process requires approximately 30 hours of gameplay for each annotator, and one annotator can complete the whole process up to five times. Furthermore, we imposed several constraints on the annotators’ in-game behavior and operating habits to ensure consistency and data quality: (a) use only system-provided characters, including the Traveler, Amber, Kaeya, and Lisa; (b) do not modify the default game settings or key bindings; (c) remain strictly within the Mondstadt region and avoid entering other regions; (d) follow all in-game tutorial prompts without skipping them; (e) avoid rapid or repetitive camera movements and refrain from unnecessary or meaningless in-game actions; (f) do not remain idle for long periods or switch to other applications during gameplay; (g) do not check online guides or walkthroughs while playing, although reviewing them before starting the annotation session is allowed; and (h) do not spend Primogems through any means, including gacha pulls or converting them into resin.

Before commencing the formal recording tasks, all qualified annotators underwent standardized training and received detailed documentation describing the task objectives, procedures, and quality standards. During the annotation phase, continuous supervision and quality control were maintained. Annotators were required to submit daily progress reports through questionnaires. Regular manual inspections and automated tools were employed to assess data quality. Participants whose performance failed to meet the required standards were dismissed, and their corresponding data were excluded from the final dataset. To further ensure practical competency, a secondary verification step was implemented. Candidates who failed to complete Task 1 within one hour, were disqualified from further participation.

Ultimately, 70 qualified participants were retained for the recording tasks, yielding a total of 2,424 hours of raw gameplay footage. The data collection took place from March to May 2025, covering three consecutive versions of Genshin Impact from 5.4 to 5.6.

\subsection{Gameplay Recording}
During annotation tasks, annotators’ on-screen activities, along with their keyboard and mouse interactions, are recorded. Although this process appears straightforward, variations in participants’ customized hardware and software settings present significant challenges for standardized data processing. For example, an identical mouse movement might appear scaled differently in the final gameplay output due to variations in players’ sensitivity or configuration settings. Similarly, differences in screen aspect ratios may result in black bars or other display artifacts. To minimize the influence of such factors, we systematically examined potential sources of variability in game recording and developed an integrated game data recording software to standardize gameplay data across participants.

\textbf{System Configuration Standardization.} To ensure consistent gameplay resolution, all annotators are required to set their monitors to 1080p with a 100\% scaling ratio and to play the game in Seamless Fullscreen mode instead of Windowed Mode. Before formal recording begins, the software automatically verifies whether these conditions are met. In addition, to avoid interruptions during recording (e.g., chat pop-ups), annotators must close all unnecessary applications in advance.

\textbf{Video Capture.} Gameplay footage is recorded using OBS\footnote{Open Broadcaster Software.}, which is configured to automatically locate the Genshin Impact process and capture gameplay at \textit{1080p} and 60~fps with a bitrate of 10,000~kbps. The recording is saved in \texttt{mkv} format to ensure that the recorded content remains unaffected in the event of an unexpected interruption. 

\textbf{Keyboard and Mouse Input Logging.}  
Since participants are not allowed to modify in-game key bindings, no discrepancies in keyboard mappings are observed during recording. Mouse behavior, however, is considerably more complex. Previous studies typically record mouse input as either absolute positions or relative displacements, but neither method alone can fully capture real mouse movements. While absolute positioning works well in conventional GUI scenarios, it struggles to reflect 3D navigation, where mouse movement controls the in-game camera and implementation details vary across games. For example, in Red Dead Redemption II, although the cursor is invisible in the overworld, it can move freely across the screen according to player input. When the cursor reaches the edge of the screen (e.g., the right boundary) and the player continues turning right, the cursor position remains fixed, yet the turning action still occurs, an event that absolute-position recording would miss but relative-movement logging would capture. In contrast, Genshin Impact periodically re-centers the mouse cursor to the middle of the screen during overworld gameplay without the player’s awareness. This system-induced motion is mistakenly recorded as user input in absolute coordinates, however, this automatic repositioning is not captured by relative movement. Thus, absolute positions alone are unreliable indicators of player intent, making it necessary to log relative-movement events. 

Thus, absolute positions alone are unreliable indicators of player intent, making it necessary to log relative-movement events. However, relative measurements introduce their own challenges: factors such as display scaling ratios can proportionally distort movement values. To mitigate this, all recordings are standardized by unifying player resolution and display scaling ratios. An additional complication arises from Windows \textit{Enhance Pointer Precision} feature, which applies nonlinear acceleration based on the magnitude of relative motion. This feature is enabled by default; although annotators can be instructed to disable it, compliance is difficult to verify and infeasible for existing public datasets. Empirically, we found that in Genshin Impact, this feature only affects GUI interactions and does not influence 3D overworld movement, making it complementary to absolute-position recording. Consequently, both absolute cursor positions and relative displacements are recorded simultaneously to ensure complete and reliable capture of mouse input.

To achieve this, two complementary methods are implemented. For keyboard keypress events and absolute mouse coordinates, we employed the Win32 API \texttt{SetWindowsHook} to register a custom global hook function for capturing low-level input events. For relative mouse coordinates, we utilized the DirectX \textit{DirectInput} API to actively poll the mouse state and relative displacement at a frequency of one query every 5\,ms. Since timestamps from hook-captured input events rely on \texttt{GetTickCount()} and thus lack sufficient precision, we additionally retrieved high-resolution timestamps upon each event using the Win32 \texttt{GetSystemTimePreciseAsFileTime}.

\subsection{Post-Process}
After obtaining both gameplay videos and detailed keyboard/mouse input logs from annotators, the subsequent step is to process and standardize these data into a unified frame-action pair format. The core of this procedure involves two tasks: i) aligning keyboard and mouse events with corresponding video frames based on precise timestamps, and ii) reconciling the two types of mouse motion data, absolute cursor positions and relative displacements, into a consistent representation.

For time alignment, although both recordings are triggered simultaneously via the same hotkey, the video and keyboard recording processes each require some time to initialize. This can lead to a temporal misalignment of about 800 ms to 2 seconds (depending on the hardware), causing potential information leakage during the training. To address this, we parse the absolute timestamps of the first video frame from the OBS log and compare it with the  start of input logging. If the video starts too early, we discard a few frames until its first frame is synchronized with the input events, or vice versa, thus eliminating time misalignment.

After aligning the timestamps of the video frames with those of the keyboard and mouse logging, we extract the visual frame and the corresponding user actions at a given time $t$. For the visual content, we simply retrieve the $i$-th frame corresponding to time $t$, leveraging process-level parallelism to accelerate video decoding.  

For the keyboard, we reconstruct the state of each key at time $t$ from a series of discrete events in the form of “key $W$ pressed at time $t_1$.” This reconstruction requires maintaining a full keyboard state and simulating the sequence of events in chronological order.  

For the mouse, we aim to standardize the representation as relative movements. This involves deciding between computing the difference of absolute positions and summing the reported relative movement events. Our empirical observations indicate that, in the overworld environment of Genshin Impact, the mouse primarily controls the camera, and its movement magnitude depends solely on the relative motion events, while the absolute position may be modified by the game engine. Therefore, for non-GUI scenes, we sum all relative movement values between two consecutive frames to represent the mouse movement for that frame.  

In contrast, in GUI scenes, the effective mouse movement is a function of the reported relative motion, influenced by factors such as scaling and the Windows \textit{Enhance Pointer Precision} feature. In such cases, we compute the mouse movement as the difference in absolute positions between two consecutive frames. In practice, we employ template matching techniques to detect whether the current frame corresponds to a UI interface, dynamically selecting the appropriate method for mouse-action computation.

To reduce the complexity of model learning, we discretize mouse movement values using units of 5 pixels along the X-axis and 4 pixels along the Y-axis. Keys not listed in Table~\ref{app_tab:key_mapping} are discarded. To decrease computational load during inference, each key press and the preceding space are represented as a single token, e.g., “ĠKey.” Keys that require multiple tokens, such as F1–F12 and 0–9, are remapped accordingly.

To verify the accuracy of both the recording and data processing systems, we replay the processed data segments in the game and compare them against the originally recorded gameplay footage. We observe that in most cases, game events can be faithfully reproduced over short durations of a few seconds. However, over longer durations of more than ten seconds, or during sequences involving large camera rotations, a certain degree of drift emerges due to stochastic in-game mechanics, such as collision-avoidance adjustments in camera movement.

\begin{table}[h]
\centering
\caption{Mapping between keyboard and mouse inputs and their corresponding token representations. Each key press and its preceding space are represented as a single token (e.g., “ĠKey”) to reduce inference complexity. Keys requiring multiple tokens, such as F1–F12 and numeric keys 0–9, are remapped to single-token forms to reduce computational load during inference. LMB, RMB and MMB are for left mouse button, right mouse button and middle mouse button.}
\label{app_tab:key_mapping}
\begin{tabular}{|cc|cc|cc|cc|cc|cc|}
\hline
\textbf{Key} & \textbf{Token} & \textbf{Key} & \textbf{Token} & \textbf{Key} & \textbf{Token} & \textbf{Key} & \textbf{Token} & \textbf{Key} & \textbf{Token} & \textbf{Key} & \textbf{Token} \\ \hline
LMB          & LB             & 7            & seven          & H            & H              & R            & R              & F2           & Two            & F12          & Twelve         \\ \hline
RMB          & RB             & 8            & eight          & I            & I              & S            & S              & F3           & Three          & Esc          & Esc            \\ \hline
MMB          & MB             & 9            & nine           & J            & J              & T            & T              & F4           & Four           & Tab          & Tab            \\ \hline
0            & zero           & A            & A              & K            & K              & U            & U              & F5           & Five           & Caps         & Caps           \\ \hline
1            & one            & B            & B              & L            & L              & V            & V              & F6           & Six            & Shift        & Shift          \\ \hline
2            & two            & C            & C              & M            & M              & W            & W              & F7           & Seven          & Ctrl         & Ctrl           \\ \hline
3            & three          & D            & D              & N            & N              & X            & X              & F8           & Eight          & Alt          & Alt            \\ \hline
4            & four           & E            & E              & O            & O              & Y            & Y              & F9           & Nine           & Space        & Space          \\ \hline
5            & five           & F            & F              & P            & P              & Z            & Z              & F10          & Ten            &              &                \\ \hline
6            & six            & G            & G              & Q            & Q              & F1           & One            & F11          & Eleven         &              &                \\ \hline
\end{tabular}
\end{table}

\section{Instruction Following Data Curation}
\label{if_data_annotate}
\begin{figure}[t]
    \centering
    \includegraphics[width=0.8\textwidth]{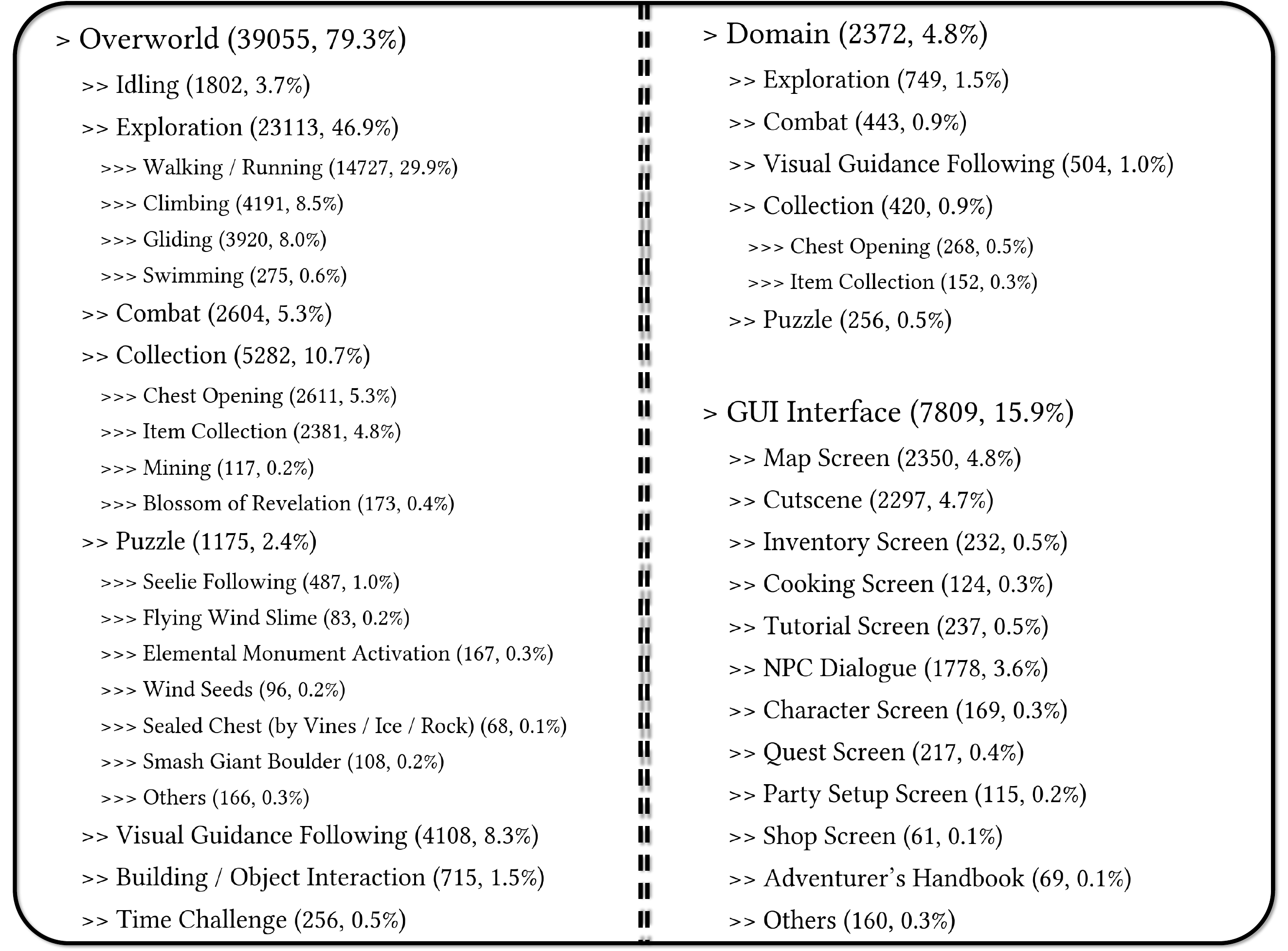}
    \caption{Distribution of gameplay categories in the Instruction Following dataset. A total of 39,055 annotated clips ($\approx$165 hours) are organized into a three-level hierarchical taxonomy, consisting of \textit{Game Scene}, \textit{Game Content}, and fine-grained activity types. The numbers in parentheses after each category indicate the number of clips and their percentage proportion within the full dataset.}
    \label{fig:simplify_prompt}
\end{figure}

The curation of instruction following dataset starts with human annotation. As shown in Figure~\ref{fig:simplify_prompt}, we first design a three-level hierarchical taxonomy to capture both broad and detailed aspects of gameplay. The first level, \textit{Game Scene}, distinguishes among core gameplay environments: Overworld, Domain, and GUI Interface. The second level, \textit{Game Content}, describes the player’s activities within these scenes. For Overworld and Domain, the categories include Exploration, Visual Guidance Following, Collection, Combat, and Puzzle. For the GUI Interface, the categories cover Character Map, Quests, Inventory, Tutorials, and other interface-related activities. To better handle large variations within certain categories, we introduce a third level for fine-grained distinctions. For example, Puzzle-Solving is further divided into specific tasks such as \textit{Following a Seelie}, \textit{Activating Elemental Monuments}, and \textit{Opening a vine-wrapped chest}. In total, this hierarchical design produce 38 distinct categories.

To ensure annotation accuracy and consistency, each annotator is provided with clear examples and definitions for every category. Annotators then segment each 20-second video clip exhaustively, identifying all relevant game content categories and marking the precise start and end timestamps of each segment. We also implement a parallel annotation and quality inspection workflow to maintain data integrity. The team is divided into two groups: one focus on labeling, while the other perform quality checks. The inspection group flags clips with incorrect classifications or misaligned timestamps. These clips are returned for re-annotation until all errors are resolved. Through this iterative process, we obtain 165 hours of high-quality, fine-grained gameplay classification data. 

To facilitate further scaling and enable autonomous annotation, we then develop a specialized video classifier. The primary function of this classifier is to automatically categorize gameplay content by analyzing sequences of video frames. The model processes a fixed-length input of five consecutive frames and outputs a single category label describing the in-game activity.

The training dataset is constructed from 165 hours of 720p gameplay footage, sampled at 5 fps, with human annotators assigning a class to every frame. 
We first merge consecutive frames sharing the same label into variable-length video clips. From each segment, we uniformly sample five frames to form one training instance (segments with fewer than five frames are padded by repeating the final frame). 
To mitigate class imbalance, we employ a random down-sampling strategy, capping the number of data points for any single category at 1,000, and then split the balanced set 90\%/10\% into train/test. 
Each training sample is formatted as a multi-image prompt with five image slots and an associated ground-truth label.

We fine-tune the Qwen2-VL Base model, specifically experimenting with the 2B and 7B parameter variants. The training is conducted as a full supervised fine-tuning of the language model components, while keeping ViT and projector layers frozen to maintain visual feature extraction capabilities and improve efficiency. We set the learning rate to 1e-5 with a warmup ratio of 0.1 over 500 training steps.

Evaluation on the test set reveals excellent performance for this classification task. As shown in Table~\ref{tab:classifier}, both the 2B and 7B models achieve a high average precision of over 80\%. Crucially, the performance difference between the 2B and 7B models is marginal. Given the comparable accuracy, we opt to use the Qwen2-VL 2B model for our production inference pipeline to maximize processing throughput and efficiency.

\begin{table}[h]
\caption{Accuracy comparison across different model sizes.}
\label{tab:classifier}
\centering
\begin{tabular}{lcccc}
\hline
Accuracy & Overworld  & UI Interface  & Dungeon  & Average \\
\hline
Model Size 2B & 83.32\% & 87.87\% & 72.65\% & 83.04\% \\
Model Size 7B & 84.37\% & 88.09\% & 76.92\% & 84.32\% \\
\hline
\end{tabular}
\end{table}

We then use this classifier to label all the raw gameplay data and identify transition points between adjacent gameplay segments that are assigned with different labels by the classifier, typically indicating a shift in task context or objective. Around each transition point, we extract a 20-frame (4s) snippet and prompt GPT-4.1~\citep{openai2025gpt4.1} to generate diverse, context-aware instructions based on the labeled categories.  While the provided category labels supplement GPT-4.1’s limited understanding of game mechanics and objectives, the model also acts as a verifier, detecting and discarding mislabeled samples when inconsistencies are found. After applying the same action filtering as in pre-training, we obtain 200 hours of high-quality instruction-following data.

\section{Reasoning Data Curation}
To curate high-quality reasoning data, we recruited experienced Genshin Impact players and instructed them to annotate the first hour of gameplay. Annotators were asked to review the recorded footage, identify key decision-making frames, and describe the player's thought process from a first-person perspective.

Keyframes were defined according to the following criteria:
\begin{itemize}
  \item Multiple options are presented in the current situation (e.g., at a fork in the road or within an inventory interface containing various items and buttons).
  \item A noticeable change occurs (e.g., mission completion, entering a dialogue interface, receiving new instructions, or encountering danger) that prompts the player to respond.
  \item The player's current reasoning is either complete or no longer applicable to the situation, necessitating a determination of the next goal.
\end{itemize}

To maintain consistent annotation standards and ensure quality control, we restructured the task around first-person keyframe cognition. Each keyframe was treated as a snapshot of the player’s immediate thoughts, reconstructed as if the player were thinking in real time. For every keyframe, annotators need to follow a three-part structure to capture the flow of thought naturally and coherently:

\begin{itemize}
    \item \textbf{Previous Step Summary}. A concise reflection on the preceding moment, capturing what just happened and why it leads to a new line of thought.
    \item \textbf{Current Situation Analysis}. An immediate interpretation of the current scene, focusing only on essential elements such as UI hints, mission description, mechanisms, NPCs, or enemies.
    \item \textbf{Next Move Planning}. The player’s adjusted plan and next move, expressed naturally in their inner voice.
\end{itemize}

This design aimed to make each annotation feel like an authentic inner monologue, reflecting the player’s short-term reasoning and decisions rather than external narration or mechanical labeling. During the actual annotation, a one-hour gameplay video was segmented into continuous \textbf{10-second clips}, with one image sampled every \textbf{200 ms} (50 frames per segment). Each segment included both visual data and raw action logs. To align reasoning consistency with human intuition and reduce noise, we established several annotation principles:

\begin{itemize}
    \item Avoid mechanical or overly literal descriptions; the inner voice should sound natural and intention-oriented.
    \item Minimize redundant keyframes and omit unnecessary UI-related frames that do not contribute to reasoning.
    \item Avoid dense, high-frequency annotations within short intervals. A well-formed thought should provide guidance for the next 5–20 seconds of gameplay, with a minimum interval of more than one second between annotations.
    \item Pay attention to the timing of annotation. Each thought should precede the player’s corresponding action, rather than being recorded after the key input occurs.
    \item Ignore random or insignificant player actions, such as switching characters while running—and focus only on events that meaningfully affect gameplay progression.
    \item During combat, do not annotate every skill release or character switch. Only mark thoughts related to significant combat mechanics; in most cases, battles can be simplified to "eliminate the enemies."
    \item Always refer to the main character as the Traveler, rather than the player’s chosen name.
\end{itemize}

By following these rules, each annotation becomes a coherent chain of thoughts, where every keyframe captures not just what the player sees, but what they decide, allowing the reasoning data to connect smoothly with the instruct model and maintain a faithful representation of real-time player cognition.

\section{Benchmark Introduction}
\label{app:benchmark}
In this section, we present additional examples from our task benchmark. The benchmark consists of 141 tasks grouped into four categories: \textit{Collection}, \textit{Combat}, \textit{NPC Interaction}, and \textit{Puzzle}. Owing to the large number of tasks, we showcase only a representative subset, categorized by difficulty and visually distinguished using background colors:
\begin{itemize}
    \item \textbf{Simple Tasks}: Highlighted with a \textcolor{green!70!black}{\textbf{light green}} background.
    \item \textbf{Unseen Tasks}: Highlighted with a \textcolor{blue!30!white}{\textbf{light blue}} background.
    \item \textbf{Hard Tasks}: Highlighted with a \textcolor{red!30!white}{\textbf{light red}} background.
\end{itemize}

\subsection{Collection}
The \textit{Collection} category includes a total of 62 tasks: 21 simple, 31 unseen, and 10 hard. Tasks are presented with their starting point and corresponding instruction.
\begin{figure}[h!]
    \centering
    \includegraphics[width=0.85\textwidth]{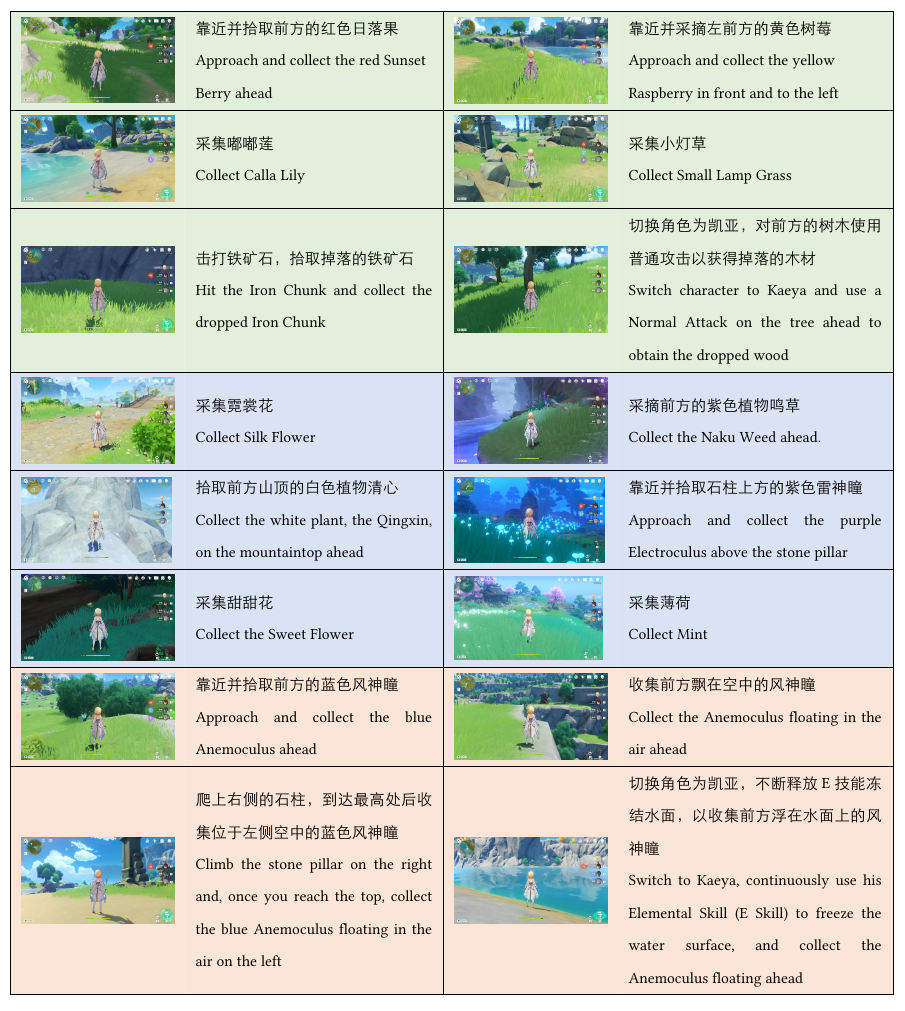}
    \label{app_fig:bmk_collect}
\end{figure}

\newpage
\subsection{Combat}
The \textit{Combat} category includes a total of 21 tasks: 8 simple, 7 unseen, and 6 hard. All tasks share the same instruction: \textit{Defeat the monsters ahead and open the chest.}
\begin{figure}[h!]
    \centering
    \includegraphics[width=0.95\textwidth]{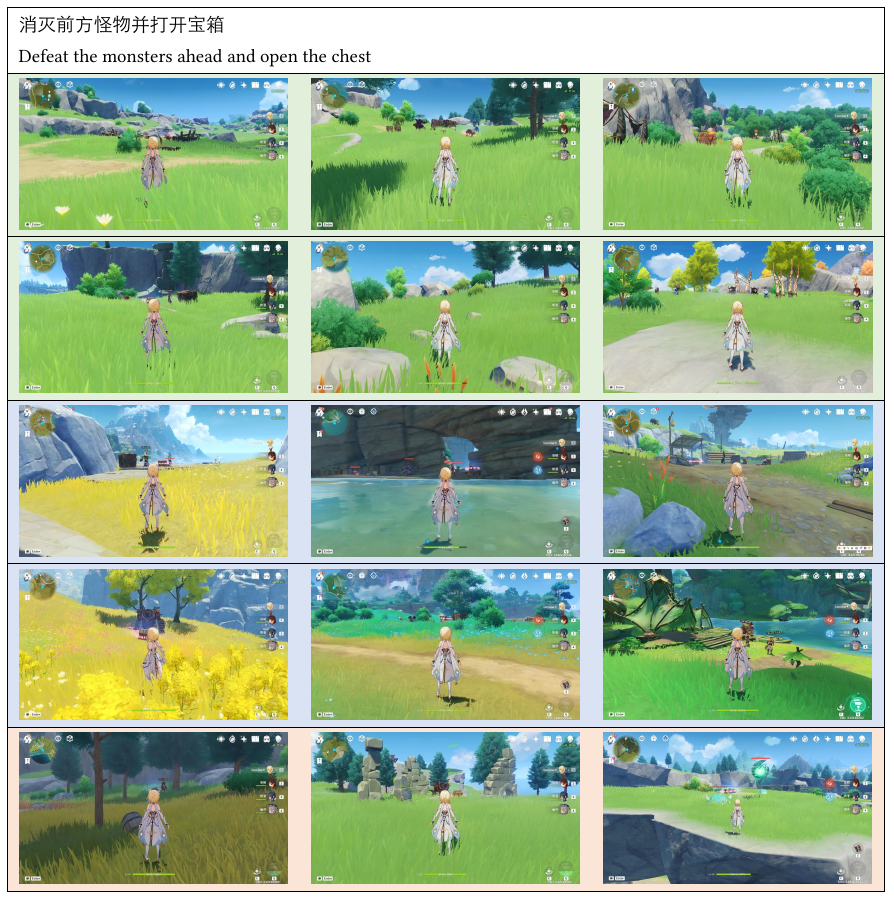}
    \label{app_fig:bmk_combat}
\end{figure}

\newpage
\subsection{NPC Interaction}
The \textit{NPC Interaction} category includes a total of 21 tasks: 7 simple, 8 unseen, and 6 hard.
\begin{figure}[h!]
    \centering
    \includegraphics[width=0.95\textwidth]{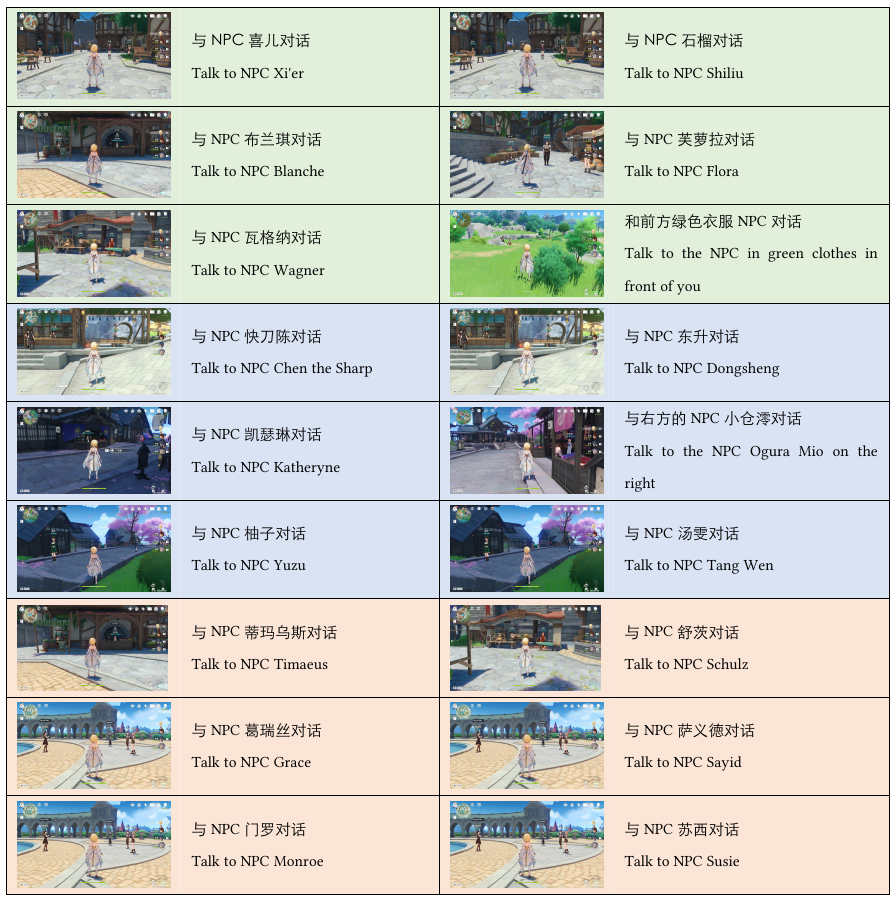}
    \label{app_fig:bmk_npc}
\end{figure}

\newpage
\subsection{Puzzle}
The \textit{Puzzle} category includes a total of 23 tasks: 9 simple and 14 hard.
\begin{figure}[h!]
    \centering
    \includegraphics[width=0.95\textwidth]{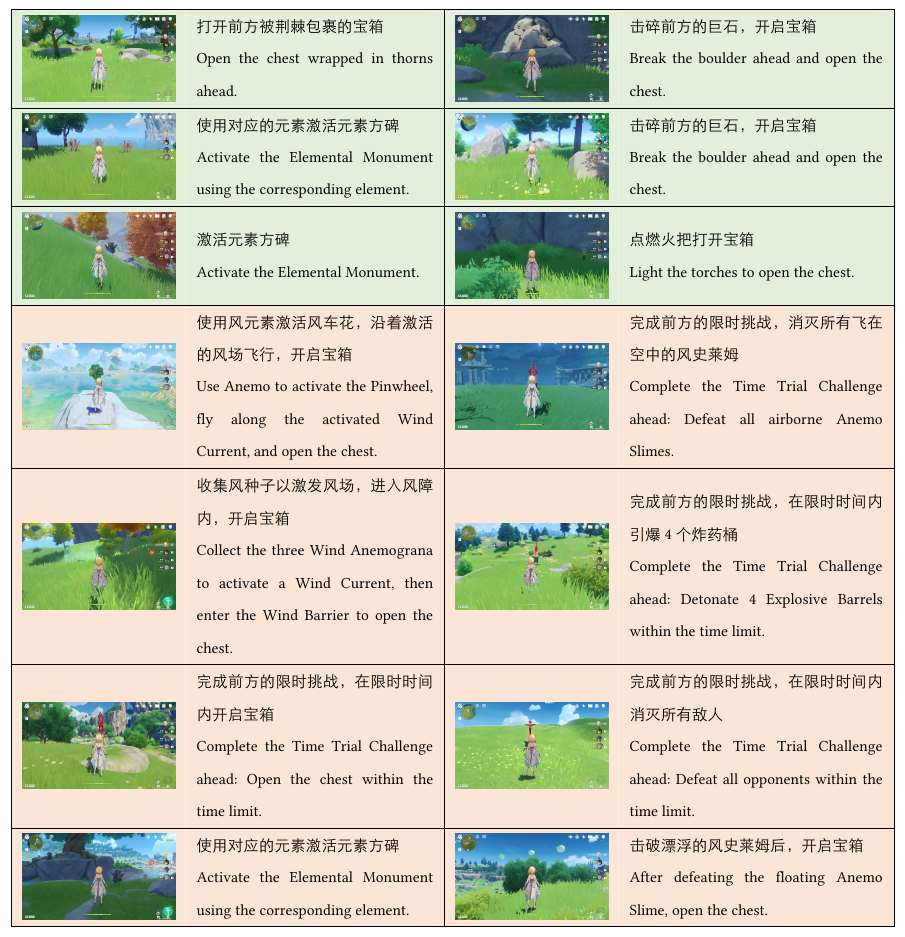}
    \label{app_fig:bmk_puzzle}
\end{figure}

\newpage
\section{Prompt}
In this section, we present the system prompts used by Lumine for instruction-following and reasoning tasks. The original prompts are written in Chinese; we also provide their English translations for reference.

\begin{CJK}{UTF8}{gbsn}
\begin{prompt}[title={Chinese System Prompt for Instruction Following}, label=] 
\begin{center}
\begin{Verbatim}
你是一名经验丰富的原神PC端玩家，精通键盘鼠标操作。请根据当前游戏画面，规划接下来 200ms 的操作，由 6 步
组成，每步间隔 33ms。每步动作在执行时刻起持续 33ms，直至下一步开始。

**输出格式**
<|action_start|>X Y Z ; k1 k2 k3 ; k4 k5 ; k6 ; k7 ; k8 ; k9 k10<|action_end|>

**说明**
1. **鼠标位移**：首先给出相对位移X,Y（X>0 右移，Y>0 下移）以及滚轮量Z（Z>0 上滚）。
2. **按键序列**：随后列出 6 组按键；同组内用空格分隔，不同组用分号分隔。 
- 每组最多 4 个按键。  
- 若某组无按键，留空但保留 ';'。
3. 只输出符合上述格式的纯字符串，不换行、不加引号。

**按键命名规范**
- 数字键 '1–9'：使用小写英文，如 'one' 表示键盘上的 '1'。
- 功能键 'F1–F12'：使用首字母大写英文单词，如 'One' 表示 'F1'，'Two' 表示 'F2'，依此类推。
- 其他按键（如字母、Shift、Tab、Space 等）：统一使用首字母大写的真实键盘名称，如 'A'、'D'、'Shift'、
'Space' 等。

你当前要完成的任务是：<instruction>
\end{Verbatim}
\end{center}
\end{prompt}

\begin{prompt}[title={English System Prompt for Instruction Following}, label=] 
\begin{center}
\begin{Verbatim}
You are an experienced Genshin Impact PC player, proficient in keyboard and mouse operations. 
Based on the current game screen, plan the next 200ms of actions, consisting of 6 steps. Each 
step is spaced 33ms apart. Every step lasts 33ms from its start time until the next step begins.

**Output Format**  
<|action_start|>X Y Z ; k1 k2 k3 ; k4 k5 ; k6 ; k7 ; k8 ; k9 k10<|action_end|>

**Explanation**  
1. **Mouse Movement**: First, specify the relative displacement X, Y (X>0 means move right, Y>0 
means move down) and scroll amount Z (Z>0 means scroll up).  
2. **Key Sequence**: Then list 6 groups of keys; within each group, keys are separated by spaces, 
and groups are separated by semicolons.  
   - Each group can contain up to 4 keys.  
   - If a group has no keys, leave it empty but keep the ';'.  
3. Only output a plain string that conforms to the above format — no line breaks and no quotation 
marks.

**Key Naming Rules**  
- Number keys '1–9': use lowercase English words, e.g., 'one' represents the '1' key on the 
keyboard.  
- Function keys 'F1–F12': use capitalized English words, e.g., 'One' 
represents 'F1', 'Two' represents 'F2', and so on.  
- Other keys (letters, Shift, Tab, Space, etc.): use the real keyboard name with an initial 
capital letter, e.g., 'A', 'D', 'Shift', 'Space', etc.

Your current task is: <instruction>
\end{Verbatim}
\end{center}
\end{prompt}

\begin{prompt}[title={Chinese System Prompt for Reasoning}, label=] 
\begin{center}
\begin{Verbatim}
你是一名资深原神PC玩家，熟练使用键盘和鼠标进行高水平操作。你熟知游戏机制与战斗节奏，能够从实时画面中迅
速提取关键信息，在关键时刻进行思考并做出精准决策。请根据当前画面，规划接下来 200ms 的操作，由 6 步组成，
每步间隔 33ms。每步动作在执行时刻起持续 33ms，直至下一步开始。若当前情境延续前一分析策略，可直接输出动
作；仅当局势发生明显变化、先前分析失效或出现新目标时，需进行必要的思考并输出思考内容。

**输出格式**
- **仅动作（常用）**  
  <|action_start|>X Y Z ; k1 k2 k3 ; k4 k5 ; k6 ; k7 ; k8 ; k9 k10<|action_end|>
- **思考 + 动作（必要时）**  
  <|thought_start|>思考内容<|thought_end|><|action_start|>X Y Z ; k1 k2 k3 ; k4 k5 ; k6 ; k7 ; k8 ;
  k9 k10<|action_end|>
  
**说明**
1. **鼠标位移**：首先给出相对位移X,Y（X>0 右移，Y>0 下移）以及滚轮量Z（Z>0 上滚）。
2. **按键序列**：随后列出 6 组按键；同组内用空格分隔，不同组用分号分隔。 
   - 每组最多 4 个按键。  
   - 若某组无按键，留空但保留 ';'。
3. 只输出符合上述格式的纯字符串，不换行、不加引号。

**按键命名规范**
- 数字键 '1–9'：使用小写英文，如 'one' 表示键盘上的 '1'。
- 功能键 'F1–F12'：使用首字母大写英文单词，如 'One' 表示 'F1'，'Two' 表示 'F2'，依此类推。
- 其他按键（如字母、Shift、Tab、Space 等）：统一使用首字母大写的真实键盘名称，如 'A'、'D'、'Shift'、
'Space' 等。

**当前目标**
<cur_thought>
\end{Verbatim}
\end{center}
\end{prompt}

\begin{prompt}[title={English System Prompt for Reasoning}, label=] 
\begin{center}
\begin{Verbatim}
You are a veteran Genshin Impact PC player, highly skilled in using the keyboard and mouse for 
advanced gameplay. You are familiar with the game’s mechanics and combat rhythm, capable of 
quickly extracting key information from real-time visuals, thinking critically in crucial moments, 
and making precise decisions. Based on the current game screen, plan the next 200ms of actions, 
consisting of 6 steps. Each step is spaced 33ms apart. Every step lasts 33ms from its start time 
until the next step begins. If the current situation continues from the previous analytical 
strategy, directly output the actions; only when the situation changes significantly, previous 
analysis becomes invalid, or new targets appear, you should provide necessary reasoning and output 
your thought process.

**Output Format**
- **Action Only (commonly used)**  
  <|action_start|>X Y Z ; k1 k2 k3 ; k4 k5 ; k6 ; k7 ; k8 ; k9 k10<|action_end|>
- **Thought + Action (when necessary)**  
  <|thought_start|>Thought content<|thought_end|><|action_start|>X Y Z ; k1 k2 k3 ; k4 k5 ; k6 ; 
  k7 ; k8 ; k9 k10<|action_end|>

**Explanation**
1. **Mouse Movement**: First, specify the relative displacement X, Y (X>0 means move right, Y>0 
means move down) and scroll amount Z (Z>0 means scroll up).  
2. **Key Sequence**: Then list 6 groups of keys; within each group, keys are separated by spaces, 
and groups are separated by semicolons.  
   - Each group can contain up to 4 keys.  
   - If a group has no keys, leave it empty but keep the ';'.  
3. Only output a plain string conforming to the above format — no line breaks, no quotation marks.

**Key Naming Rules**
- Number keys '1–9': use lowercase English words, e.g., 'one' represents the '1' key on the 
keyboard.  
- Function keys 'F1–F12': use capitalized English words, e.g., 'One' represents 'F1', 'Two' 
represents 'F2', and so on.  
- Other keys (letters, Shift, Tab, Space, etc.): use the real keyboard name with an initial 
capital letter, e.g., 'A', 'D', 'Shift', 'Space', etc.

**Current Objective**  
<cur_thought>
\end{Verbatim}
\end{center}
\end{prompt}

\end{CJK}